\documentclass[11pt]{article}

\usepackage{graphicx,amsfonts,latexsym}
\usepackage{eurosym,color}

\usepackage[english]{babel}
\usepackage{amssymb}
\usepackage{amsmath}
\usepackage{latexsym}
\usepackage{graphicx}
\usepackage[lined,boxed,commentsnumbered]{algorithm2e}
\usepackage[multiple]{footmisc}
\usepackage{xspace}
\usepackage{epstopdf}

\newcommand{\toCheck}[1]{#1\xspace}
\newcommand{\toChange}[1]{#1\xspace}

\newcommand{\one}{{\mathbf{1}}}
\newcommand{\tran}{^{\top}}

\newcommand{\qed}{{\hfill $\square$}}

\newcommand{\diag}{\mbox {\rm diag}}
\newcommand{\tr}{\mbox {\rm Tr }}
\newcommand{\lam}{\lambda}

\newcommand{\angleDomain}{(-\pi,+\pi]}

\newcommand{\e}{{\mathrm e}}
\newcommand{\inv}{^{-1}}
\newcommand{\beq}{\begin{equation}}
\newcommand{\eeq}{\end{equation}}
\newcommand{\bea}{\begin{eqnarray}}
\newcommand{\eea}{\end{eqnarray}}
\newcommand{\beas}{\begin{eqnarray*}}
\newcommand{\eeas}{\end{eqnarray*}}
\newcommand{\ba}{\begin{array}}

\newcommand{\ea}{\end{array}}
\newcommand{\bit}{\begin{itemize}}
\newcommand{\eit}{\end{itemize}}
\newcommand{\ben}{\begin{enumerate}}
\newcommand{\een}{\end{enumerate}}

\newcommand{\Real}[1]{ { {\mathbb R}^{#1} } }

\newcommand{\Complex}[1]{ { {\mathbb C}^{#1} } }

\newcommand{\ped}[1]{{_{\mathrm{#1}}}}

\newtheorem{corollary}{Corollary}
\newtheorem{theorem}{Theorem}
\newtheorem{lemma}{Lemma}

\newtheorem{assumption}{Assumption}
\newtheorem{proposition}{Proposition}
\newtheorem{remark}{Remark}
\newtheorem{definition}{Definition}

\newcommand{\setal}{~\emph{et.~al}\xspace}

\newcommand{\calA}{{\mathcal A}}

\newcommand{\calG}{{\mathcal G}}

\newcommand{\calN}{{\mathcal N}}
\newcommand{\calT}{{\mathcal T}}

\newcommand{\calP}{{\mathcal P}}

\newcommand{\calE}{{\mathcal E}}

\newcommand{\calL}{{\mathcal L}}
\newcommand{\calV}{{\mathcal V}}
\newcommand{\calW}{{\mathcal W}}
\newcommand{\calX}{{\mathcal X}}

\newcommand{\subt}{\mbox{\rm s.t.:}}

\newcommand{\red}[1]{{\color{red}#1}}

\newcommand{\eye}{I}
\newcommand{\zeros}{0}
\newcommand{\ones}{1}
\newcommand{\vect}[1]{\left[\begin{array}{c}  #1  \end{array}\right]}
\newcommand{\matTwo}[1]{\left[\begin{array}{cc}  #1  \end{array}\right]}

\newcommand{\nrNodes}{n}
\newcommand{\Rij}{R_{ij}}
\newcommand{\Rji}{R_{ji}}
\renewcommand{\ij}{_{ij}}
\newcommand{\ji}{_{ji}}
\newcommand{\ii}{_{ii}}
\newcommand{\Dij}{D_{ij}}
\newcommand{\Dji}{D_{ji}}

\newcommand{\Deltaij}{\Delta_{ij}}
\newcommand{\todo}[1]{}

\newcommand{\SOtwo}{SO(2)}
\newcommand{\SEtwo}{SE(2)}

\newcommand{\diffp}{p_{j}-p_{i}}
\newcommand{\calNout}{\calN^{\text{out}}}
\newcommand{\calNin}{\calN^{\text{in}}}
\newcommand{\PGO}{PGO\xspace}
\newcommand{\PGOlong}{Pose Graph Optimization\xspace}
\newcommand{\hide}[1]{}
\newcommand{\hideProof}[1]{\red{hidden proof.}}

\newcommand{\aSOtwo}{\alpha\SOtwo}

\newcommand{\Rtemp}{Z}

\newcommand{\Wfull}{\calW}
\newcommand{\Laplacian}{\calL}
\newcommand{\LaplacianRed}{L}
\newcommand{\AugLaplacian}{\bar \Laplacian}
\newcommand{\Incidence}{\calA}
\newcommand{\AugIncidence}{\bar \Incidence}
\newcommand{\IncidenceUGGcpx}{\tilde U}

\newcommand{\barD}{\bar D}
\newcommand{\barU}{\bar U}

\newcommand{\SZEP}{SZEP\xspace}

\newcommand{\panc}{\rho}
\newcommand{\Wanc}{W}
\newcommand{\Sanc}{S}
\newcommand{\Lanc}{L}
\newcommand{\Aanc}{A}

\newcommand{\Qcpx}{\tilde Q}
\newcommand{\Scpx}{\tilde S}
\newcommand{\Lcpx}{\LaplacianRed}

\newcommand{\pcpx}{\tilde \panc}
\newcommand{\rcpx}{\tilde r}
\newcommand{\Wcpx}{\tilde \Wanc}

\newcommand{\lagrangian}{\mathbb{L}}
\newcommand{\xcpx}{\tilde x}
\newcommand{\Xcpx}{\tilde X}
\newcommand{\dimWcpx}{2n-1}

\newcommand{\zcpx}{\tilde z}
\newcommand{\pose}{x}
\newcommand{\eq}{Eq.}

\begin{document}
\author{Giuseppe C. Calafiore\thanks{G.C. Calafiore, Dipartimento di Automatica e Informatica,
Politecnico di Torino, Italy.
E-mail: {\tt giuseppe.calafiore@polito.it}
} 
 \xspace, Luca Carlone\thanks{
L.\,Carlone, School of Interactive Computing,  
College of Computing, Georgia Institute of Technology, Atlanta, GA, USA. 
E-mail: {\sf luca.carlone@gatech.edu}
}
\xspace, and Frank Dellaert\thanks{
F.\,Dellaert, School of Interactive Computing,  
College of Computing, Georgia Institute of Technology, Atlanta, GA, USA. 
E-mail: {\sf dellaert@cc.gatech.edu} 
}
}
\title{Pose Graph Optimization in the Complex Domain:  \\
Lagrangian Duality, Conditions For Zero \\ 
Duality Gap, and Optimal Solutions}

\date{}
\maketitle

\begin{abstract}
\emph{\PGOlong} (\PGO) is the problem of estimating a set of poses from 
pairwise relative measurements. 
\PGO is a nonconvex problem, and currently no known technique can guarantee the computation of a
global 
optimal solution. In this paper, we show that Lagrangian duality allows computing a 
globally optimal solution, under certain conditions that are satisfied in many practical cases.
 Our first contribution is to frame the \PGO problem in the complex domain. This makes analysis 
easier and allows drawing connections with the recent literature on \emph{unit gain graphs}. 
Exploiting this connection we prove nontrival results about the spectrum of the matrix underlying 
the problem. 
%
The second contribution is to formulate and analyze the properties of the Lagrangian
dual problem in the complex domain. The dual problem is a semidefinite program (SDP).
Our analysis shows that the duality gap is 
connected to the number of eigenvalues of the \emph{penalized pose graph matrix}, which 
arises from the solution of the SDP. We prove that if this matrix has a 
\emph{single eigenvalue in zero}, then (i) the duality gap is zero, (ii) the primal \PGO problem has a 
unique solution, and (iii) the primal solution can be computed by \emph{scaling} an eigenvector 
of the penalized pose graph matrix. 
%
The third contribution is algorithmic: we exploit the dual problem and propose
an algorithm that computes a guaranteed optimal solution for 
\PGO when the penalized pose graph matrix satisfies the Single Zero Eigenvalue Property (\SZEP).
We also propose a variant that deals with the case in which the \SZEP is not satisfied.
This variant, while possibly suboptimal, provides a very good estimate for \PGO 
in practice. 
The fourth contribution is a numerical analysis. Empirical evidence shows that 
in the vast majority of cases ($100\%$ of the tests under noise regimes of practical robotics applications) the penalized 
pose graph matrix does satisfy the \SZEP, hence 
 our approach allows computing the global optimal solution.
Finally, we report simple counterexamples in which the duality gap is nonzero, 
 and discuss open problems.
\vspace{.1cm}
\end{abstract}
%


\newcommand{\parStyle}[1]{{\bf #1}}
\section{Introduction}

\emph{Pose graph optimization} (\PGO) consists in the estimation of the poses (positions and orientations)
of a mobile robot, from 
relative pose measurements. The problem can be formulated as the minimization of a nonconvex cost, and 
can be conveniently visualized as a graph, 
in which a (to-be-estimated) pose is attached to each vertex, and a given relative pose measurement 
is associated to each edge. 

\PGO is a key problem in many application endeavours. 
In robotics, it lies at the core of 
state-of-the-art algorithms for localization and mapping in both single 
robot~\cite{Lu97,Duckett02ar,Frese05tro,Olson06icra,Grisetti09its,Carlone11rss,Dubbelman12icra,Johannsson13icra,Carlone14ijrr,Carlone14tro} 
and multi robot~\cite{Kim10icra,Aragues11icra,Knuth12icra,Knuth13icra,Indelman14icra}
systems. In computer vision and control, problems that are 
closely related to \PGO need to be solved for 
structure from motion~\cite{Govindu01cvpr,Martinec07cvpr,Arie123dimpvt,Govindu04cvpr,
Sharp04pami,Hartley13ijcv,Fredriksson12accv}, 
attitude synchronization~\cite{Thunberg11cdc,Hatanaka10cdc,Olfati-Saber06cdc}, 
camera network calibration~\cite{Tron09cdc,Tron12cdc}, sensor network 
localization~\cite{Piovan13automatica,Peters14sicon}, and distributed consensus on 
manifolds~\cite{Sarlette09sicon,Tron12tac}.
Moreover, similar formulations arise in molecule structure determination 
from microscopy imaging~\cite{Singer10achm,Bandeira13siam}.

\begin{figure}[t]
\begin{minipage}{\textwidth}
\begin{tabular}{cc}
\begin{minipage}{7cm}%
\centering
\hspace{-7mm}\includegraphics[scale=0.24]{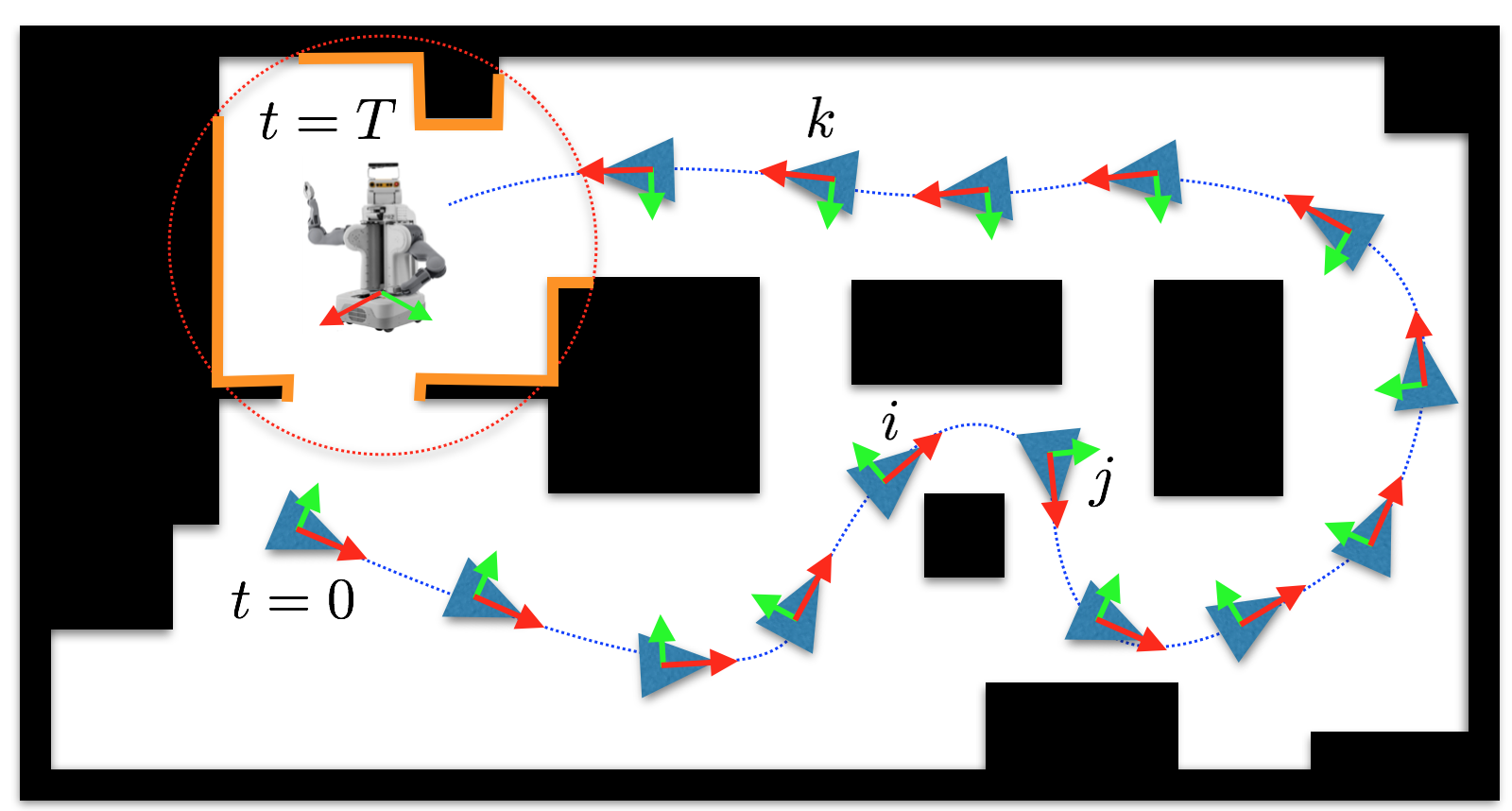} \\ (a)
\end{minipage}%
&
\hspace{-4mm}
\begin{minipage}{5cm}%
\centering
\includegraphics[scale=0.24]{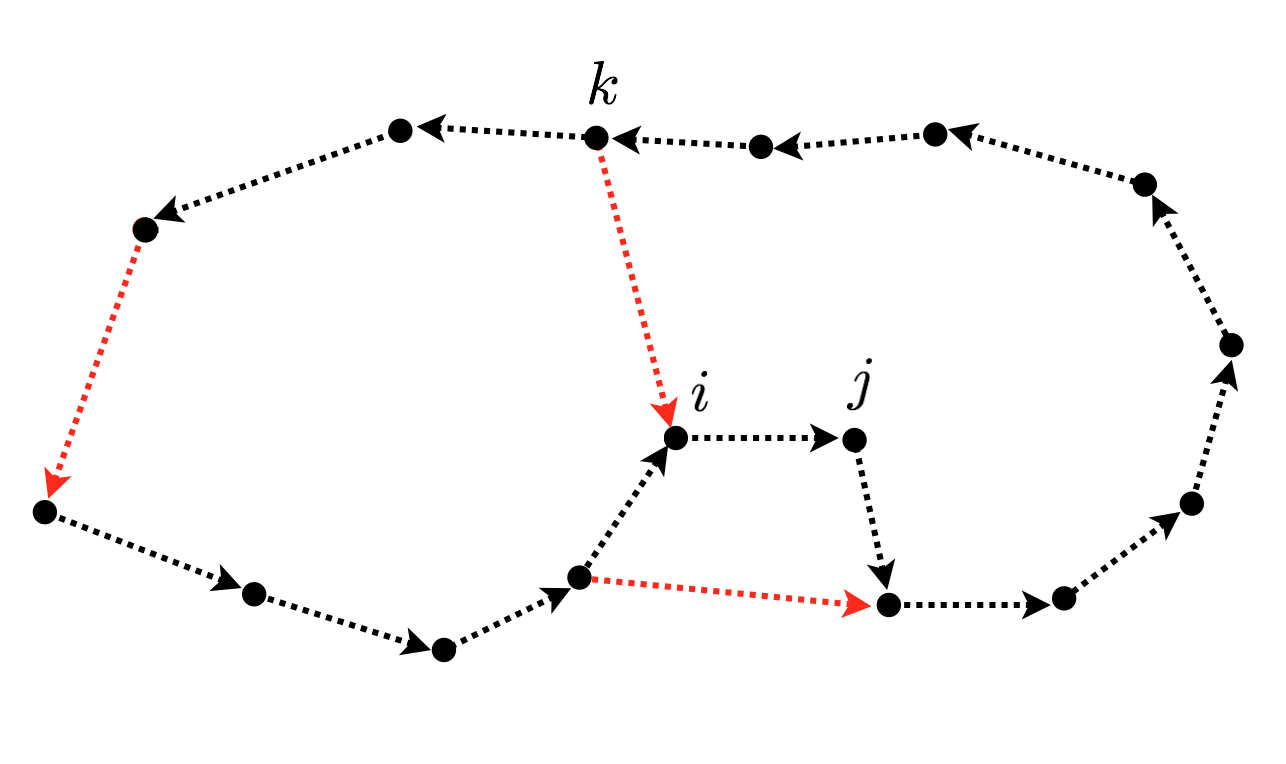}  \\ (b)
\end{minipage}%
\end{tabular}%
\end{minipage}%
\caption{\label{fig:motivation} (a) Pose graph optimization in robotics. A mobile robot is 
deployed in an unknown environment at time $t=0$.
At each time step the robot measures distances from obstacles within the 
sensing radius (red circle). The sensor footprint (i.e., the set of measurements) at time $T$ is visualized 
 in orange. By matching sensor footprints acquired at different time steps, the robot establishes relative 
 measurements between poses along its trajectory. \PGO consists in the estimation of robot poses from these relative 
 measurements.  (b) Directed graph underlying the problem.  
}
\end{figure}

A motivating example in robotics is the one pictured in Fig.~\ref{fig:motivation}(a). 
A mobile robot is deployed in an unknown environment at time $t=0$. The robot 
traverses the environment and at each discrete time step acquires a sensor measurement 
(e.g., distances from obstacles within the sensing radius). 
 From wheel rotation, the robot is capable of measuring the relative motion between 
 two consecutive poses (say, at time $i$ and $j$). Moreover, 
comparing the sensor measurement, acquired at different times, the robot can also 
extrapolate relative measurements between non consecutive poses (e.g., between $i$ and $k$ 
in the figure). \PGO uses these measurements to estimate robot poses.
The graph underlying the problem is shown in Fig.~\ref{fig:motivation}(b), where
we draw in different colors the edges due to relative motion measurements (the 
\emph{odometric edges}, in black) and the edges connecting non-consecutive poses (the \emph{loop closures}, in red).
The importance of estimating the robot poses is two-fold. First, the knowledge of the current robot pose 
is often needed for performing high-level tasks within the environment. Second, from the knowledge of 
all past poses, the robot can \emph{register} all sensor footprints in a common frame, and 
obtain a \emph{map} of the environment, which is needed for model-based navigation and path planning.  

\parStyle{Related work in robotics.} 
Since the seminal paper~\cite{Lu97}, \PGO attracted large attention 
from the robotics community.
Most state-of-the-art techniques currently rely on iterative 
nonlinear optimization, which refines a given initial guess. 
The Gauss-Newton method is a popular choice~\cite{Kuemmerle11icra,Kaess08tro,Kaess12ijrr}, as it converges 
quickly when the initialization is close to a minimum of the cost function. 
Trust region methods (e.g., the Levenberg-Marquart method, or Powell's Dog-Leg method~\cite{Levenberg44qam})
have also been applied successfully to \PGO~\cite{Rosen12icra,Rosen14tro}; 
the gradient method has been shown to have a large convergence basin, 
while suffering from long convergence tails~\cite{Olson06icra,Grisetti09its,Knuth13icra}.
%
A large body of literature focuses on speeding up computation.
This includes exploiting sparsity~\cite{Kaess08tro,Eustice06ijrr},  
using reduction schemes to limit the number of poses~\cite{Konolige04aaai,CarlevarisBianco13icra}, 
faster linear solvers~\cite{Frese05tro,Dellaert10iros}, or  
approximate solutions~\cite{Dubbelman13icra,Carlone14ijrr}.

\PGO is a nonconvex problem and iterative optimization techniques can only guarantee local convergence.
State-of-the-art iterative solvers fail to converge to a global minimum of the cost for 
relatively small noise levels~\cite{Carlone14tro,Carlone15icra-init3D}.
This fact recently triggered efforts towards the design of more robust techniques, together with 
a theoretical analysis of \PGO. Huang\setal\cite{Huang12icra} 
discuss the number of local minima in small \PGO problems. 
Knuth and Barooah~\cite{Knuth13ras} investigate 
the growth of the error in absence of loop closures.  Carlone~\cite{Carlone13icra} 
provides conservative approximations of the basin of convergence for the Gauss-Newton method.
Huang\setal~\cite{Huang10iros} and Wang\setal\cite{Wang12rss} discuss the nonlinearities in \PGO.
In order to improve global convergence, a successful strategy consists in solving for the 
rotations first, and then using the resulting estimate to bootstrap iterative methods 
 for \PGO~\cite{Carlone11rss,Carlone14ijrr,Carlone14tro,Carlone15icra-init3D}. This is convenient 
 because the \emph{rotation subproblem}\footnote{We use the term 
``rotation subproblem'' to denote the problem of associating a rotation to each node in the graph, 
using relative rotation measurements. This corresponds to disregarding the translation measurements in \PGO.} 
\toChange{can be solved in closed-form in 2D~\cite{Carlone14tro}}, 
 and many heuristic algorithms for rotation estimation also perform well in 
 3D~\cite{Martinec07cvpr,Govindu01cvpr,Fredriksson12accv,Carlone15icra-init3D}.
 Despite the empirical success of state-of-the-art techniques, no approach 
 can guarantee global convergence. It is not even known if the global 
 optimizer itself is unique in general instances (while it is known that 
 the minimizer is unique with probability one in the rotation subproblem~\cite{Carlone14tro}).
 The lack of guarantees promoted a recent interest in \emph{verification} techniques for \PGO.
 Carlone and Dellaert~\cite{Carlone15icra-verify}  
 use duality to evaluate the quality of a candidate solution in planar \PGO. 
 The work~\cite{Carlone15icra-verify} 
 also provides empirical evidence that in many problem instances the duality gap, i.e., the 
mismatch between the optimal cost of the primal and the dual problem, is zero. 
 
\parStyle{Related work in other fields.} 
Variations of the \PGO problem appear in different research fields.
In computer vision, a somehow more difficult variant of the problem
is known as \emph{bundle adjustment}~\cite{Govindu01cvpr,Martinec07cvpr,Arie123dimpvt,Govindu04cvpr,
Sharp04pami,Hartley13ijcv,Fredriksson12accv}. Contrarily to \PGO, in bundle adjustment 
the relative measurements between the (camera) poses are only known up to scale. 
While no closed-form solution is known for bundle adjustment, many authors 
focused on the solution of the rotation subproblem~\cite{Govindu01cvpr,Martinec07cvpr,Arie123dimpvt,Govindu04cvpr,
Sharp04pami,Fredriksson12accv,Hartley13ijcv}. The corresponding algorithms have excellent performance in practice, 
but they come with little guarantees, as they are based on relaxation. 
Fredriksson and Olsson~\cite{Fredriksson12accv} use duality theory to design a verification technique for 
quaternion-based rotation estimation.

Related work in multi robot systems and sensor networks also includes 
contributions on rotation estimation 
(also known as \emph{attitude synchronization}~\cite{Thunberg11cdc,Hatanaka10cdc,Olfati-Saber06cdc,Chiuso08cis,Wang13ima}).
Borra\setal\cite{Borra12automatica} propose a distributed algorithm for planar rotation estimation.
Tron and Vidal~\cite{Tron09cdc,Tron12cdc} provide convergence results for distributed attitude consensus 
using gradient descent; distributed consensus on 
manifold~\cite{Sarlette09sicon} is related to 
estimation from relative measurements, as discussed in~\cite{Tron12tac}.
A problem that is formally equivalent to \PGO is discussed in~\cite{Piovan13automatica,Peters14sicon} 
with application to sensor network localization. Piovan\setal\cite{Piovan13automatica} 
provide observability conditions and discuss iterative algorithms that 
reduce the effect of noise. Peters\setal\cite{Peters14sicon} study 
pose estimation in graphs with a single loop (related closed-form solutions 
also appear in other literatures~\cite{Sharp04pami,Dubbelman10iros}), and provide 
an estimation algorithm over general graphs, based on the limit of a set of continuous-time differential equations, 
proving its effectiveness through numerical simulations. 
\toChange{We only mention that a large literature in sensor network localization also deals with 
other types of relative measurements~\cite{Mao07cn}, including relative positions (with known rotations)~\cite{Barooah07csm,Russell11tsp}, 
relative distances~\cite{Doherty01infocom,Eren04infocom,Costa06acm,Biswas06acm,Singer08nas,Calafiore10cdc,Cucuringu12tsn,Cucuringu12ima}
and relative bearing measurements~\cite{Stanfield47iee,Eren06cdc,Tron15acc}.}

A less trivial connection can be established with related work in   
molecular structure determination from \emph{cryo-electron microscopy}~\cite{Singer10achm,Singer11siam}, 
which offers very lucid and mature treatment of rotation estimation.
Singer and Shkolnisky~\cite{Singer10achm,Singer11siam} provide two approaches for rotation estimation, 
based on relaxation and semidefinite programming (SDP). Another merit of~\cite{Singer10achm} is 
to draw connections between planar rotation estimation and the ``\textsc{max-2-lin mod l}'' problem in combinatorial optimization, 
and ``\textsc{max-k-cut}'' problem in graph theory. Bandeira\setal~\cite{Bandeira13siam} provide
 a Cheeger-like inequality that establishes performance bounds for the SDP 
relaxation. Saunderson\setal~\cite{Saunderson14cdc,Saunderson14arxiv} propose a tighter SDP 
relaxation, based on a spectrahedral representation of the convex hull of the rotation group.

%
%
%

\parStyle{Contribution.} This paper shows that the use of Lagrangian duality
 allows computing a \emph{guaranteed} globally optimal solution for \PGO in many practical cases,
and proves that in those cases the solution is unique.

Section~\ref{sec:preliminaries} recalls preliminary concepts, and discusses
 the properties of a particular set of $2 \times 2$ matrices, which are 
scalar multiples of a planar rotation matrix. These 
matrices are omnipresent in planar \PGO and acknowledging this 
fact allows reformulating the problem over complex variables.

Section~\ref{sec:PGO} frames \PGO as a problem in complex variables. This makes analysis 
easier and allows drawing connections with the recent literature on \emph{unit gain graphs}~\cite{Reff:11}.
Exploiting this connection we prove nontrival results about the spectrum of the matrix underlying 
the problem (the \emph{pose graph matrix}), such as the number of zero eigenvalues in particular graphs. 

Section~\ref{sec:duality} formulates the Lagrangian dual problem in the complex domain.
Moreover it presents an SDP relaxation of \PGO, interpreting the relaxation as 
the dual of the dual problem. Our SDP relaxation is related to the one of~\cite{Singer10achm,Fredriksson12accv}, 
but we deal with 2D poses, rather than rotations; moreover, we only use the SDP relaxation to 
complement our discussion on duality and to support some of the proofs. 
Section~\ref{sec:analysis} contains keys results that relate the solution of the dual problem to 
the primal \PGO problem. We show that the \emph{duality gap} is
connected to the zero eigenvalues of the \emph{penalized pose graph matrix}, which 
arises from the solution of the dual problem. We prove that if this matrix has a 
\emph{single eigenvalue in zero}, then (i) the duality gap is zero, (ii) the primal \PGO problem has a 
unique solution (up to an arbitrary roto-translation), 
and (iii) the primal solution can be computed by \emph{scaling} the eigenvector 
of the penalized pose graph matrix corresponding to the zero eigenvalue. To the best 
of our knowledge, this is the first work to discuss the uniqueness of the \PGO solution for 
general graphs and to provide a provably optimal solution. 

\toCheck{
Section~\ref{sec:algorithms} exploits our analysis of the dual problem to devise 
computational approaches for \PGO. We propose an algorithm 
that computes a guaranteed optimal solution for 
\PGO when the penalized pose graph matrix satisfies the Single Zero Eigenvalue Property (\SZEP).
We also propose a variant that deals with the case in which the \SZEP is not satisfied.
This variant, while possibly suboptimal, is shown to perform well in practice, 
 outperforming related approaches.
}

Section~\ref{sec:experiments} elucidates on our theoretical results with numerical tests. 
In practical regimes of operation (rotation noise $<0.3$ rad and translation noise $<0.5$ m), 
 our Monte Carlo runs always produced a penalized pose graph matrix satisfying the \SZEP.
 Hence, in all tests with reasonable noise our approach enables the computation of the optimal solution.
 For larger noise levels (e.g., $1$ rad standard deviation for rotation measurements), we observed 
 cases in which the penalized pose graph matrix has multiple eigenvalues in zero.
%
To stimulate further investigation towards 
structural results on duality (e.g., maximum level of noise for which the duality gap is provably zero)
 we report simple examples in which the duality gap is nonzero.

\section{Notation and preliminary concepts}
\label{sec:preliminaries}

Section~\ref{sec:notation} introduces our notation.
Section~\ref{sec:graphPreliminaries} recalls standard 
concepts from graph theory, and can be safely skipped by the expert reader. 
Section~\ref{sec:alphaSO2}, instead, discusses the properties of the set of $2\times 2$
matrices that are multiples of a planar rotation matrix. We denote this set with 
the symbol $\aSOtwo$. The set $\aSOtwo$ is of interest in this paper since the
action of any matrix $\Rtemp \in \aSOtwo$ can be conveniently represented 
as a multiplication between complex numbers, as discussed in Section~\ref{sec:complex}.
Table~\ref{tab:symbols} summarizes the main symbols used in this paper.

\subsection{Notation}
\label{sec:notation}

The cardinality of a set $\calV$ is written as $|\calV|$.
The sets of real and complex numbers are denoted with $\Real{}$ and $\Complex{}$, respectively. 
$\eye_n$ denotes the $n \times n$ identity matrix, $\ones_n$ denotes the 
(column) vector of all ones of dimension $n$, $\zeros_{n \times m}$ 
denotes the $n \times m$ matrix of all zeros (we also use the 
shorthand $\zeros_{n} \doteq \zeros_{n \times 1}$).
For a matrix $M$, $M\ij$ denotes the element of $M$ in row $i$ and column $j$. 
For matrices with a block structure we use $[M]\ij$ to denote the 
$d \times d$ block of $M$ at the block row $i$ and block column $j$. 
In this paper we only deal with matrices that have $2 \times 2$ blocks, i.e.,
 $d=2$, hence the notation $[M]\ij$ is unambiguous. 

\subsection{Graph terminology}
\label{sec:graphPreliminaries}

A \emph{directed graph}~$\calG$ is a pair~$(\calV,\calE)$, where the \emph{vertices} or \emph{nodes}~$\calV$ are a finite set of elements, and~$\calE\subset\calV\times\calV$ is the set 
of \emph{edges}. Each edge is an ordered pair~$e = (i,j)$. We say that~$e$
is  \emph{incident} on nodes~$i$ and~$j$, \emph{leaves} node~$i$, called \emph{tail}, 
and is \emph{directed towards} node~$j$, called \emph{head}.
The number of nodes is denoted with $n \doteq |\calV|$, while the number of 
edges is $m \doteq |\calE|$.

The \emph{incidence matrix}~$\Incidence$ of a directed graph is
a $m \times \nrNodes$ matrix with elements in~${\{-1,0,+1\}}$ 
that exhaustively describes the graph topology.
Each row of~$\Incidence$ corresponds to an edge and has exactly two non-zero elements.
For the row corresponding to edge~$e=(i,j)$, there 
is a $-1$ on the~$i$-th column  and
a $+1$ on the~$j$-th column.

The set of \emph{outgoing neighbors} of node $i$ 
 is  $\calNout_i \doteq \{j:(i,j) \in \calE \}$.
The set of \emph{incoming neighbors} of node $i$ 
 is  $\calNin_i \doteq \{j:(j,i) \in \calE \}$.
The set of \emph{neighbors} of node $i$ is the union of outgoing and incoming 
neighbors $\calN_i \doteq \calNout_i  \cup \calNin_i$.
\subsection{The set $\aSOtwo$}
\label{sec:alphaSO2}

The set $\aSOtwo$ is defined as 
\[
\aSOtwo \doteq \{\alpha R :\, \alpha \in \Real{},\, R\in\SOtwo \},
\] 
where $\SOtwo$ is the set of 2D rotation 
matrices. 
Recall that $\SOtwo$ can be parametrized by an angle $\theta \in (-\pi,+\pi]$, and 
any matrix $R \in \SOtwo$ is in the form:
\beq
\label{eq:SO2}
R = R(\theta) =
\left[
\ba{cc}
\cos(\theta) & -\sin(\theta) \\
\sin(\theta) & \cos(\theta)
\ea
\right].
\eeq
Clearly, $\SOtwo \subset \aSOtwo$.
The set $\aSOtwo$ is closed under standard matrix \emph{multiplication}, i.e., for any $\Rtemp_1,\Rtemp_2 \in \aSOtwo$,
also the product $\Rtemp_1 \Rtemp_2 \in \aSOtwo$. 
In full analogy with $\SOtwo$, it is also trivial to show that the multiplication is commutative, i.e., 
for any $\Rtemp_1,\Rtemp_2 \in \aSOtwo$ it holds that
$
\Rtemp_1 \Rtemp_2 = \Rtemp_2 \Rtemp_1$.
Moreover, for  $\Rtemp = \alpha R$ with $R\in\SOtwo$ it holds that 
$\Rtemp\tran \Rtemp = |\alpha|^2 \eye_2$.
The set $\aSOtwo$ is also closed under matrix \emph{addition}, since for $R_1,R_2\in \SOtwo$, we have that
\bea
\label{eq:addClosure}
\alpha_1 R_1 \!\!&+&\!\! \alpha_2 R_2 
= 
\alpha_1 \matTwo{c_1 & -s_1 \\ s_1 & c_1 } + \alpha_2 \matTwo{c_2 & -s_2 \\ s_2 & c_2 } = \\
&=&
 \matTwo{\alpha_1 c_1 + \alpha_2 c_2 & -(\alpha_1 s_1 + \alpha_2 s_2)
\\ 
\alpha_1 s_1 + \alpha_2 s_2 & \alpha_1 c_1 + \alpha_2 c_2} = 
 \matTwo{a & -b \\ b & a} = \alpha_{3} R_{3} \ , \nonumber 
\eea
where we used the shorthands $c_i$ and $s_i$ for $\cos(\theta_i)$ and $\sin(\theta_i)$, and 
we defined $a \doteq \alpha_1 c_1 + \alpha_2 c_2$ 
and $b \doteq \alpha_1 s_1 + \alpha_2 s_2$.
In~\eqref{eq:addClosure}, the scalar $\alpha_{3} \doteq \pm \sqrt{a^2 + b^2}$ (if nonzero)
normalizes ${\scriptsize \matTwo{a & -b \\ b & a}}$, such that  $R_{3}
\doteq {\scriptsize\matTwo{a/\alpha_{3} & -b/\alpha_{3} \\ b/\alpha_{3} & a/\alpha_{3}}}$ 
is a rotation matrix; if $\alpha_{3}=0$, then $\alpha_1 R_1 + \alpha_2 R_2 = \zeros_{2\times 2}$, 
which also falls in our definition of $\aSOtwo$.
From this reasoning, it is clear that an alternative definition of 
$\aSOtwo$ is 
\beq
\label{eq:defaSOtwo}
\aSOtwo \doteq \left\{{\small \matTwo{a & -b \\ b & a}}:a,b\in \Real{} \right\}.
\eeq
$\aSOtwo$ is tightly coupled with the set of complex numbers $\Complex{}$. 
Indeed, a matrix in the form~\eqref{eq:defaSOtwo} is also known as a \emph{matrix representation} 
of a complex number~\cite{Hazewinkel01book}.  We explore the implications of this 
fact for \PGO in Section~\ref{sec:complex}.


\begin{table}[h!]%
\caption{Symbols used in this paper}\label{tab:symbols}%
\vspace{-1mm}
\newcommand{\tablesection}[1]{%
    \multicolumn{2}{l}{\rule{0pt}{1.1em}\emph{#1}}\\ \hline \rule{0pt}{1.2em}%
}%
\renewcommand{\tabcolsep}{1pt}
\centering%
\begin{tabular}{p{4cm}p{8.0cm}}
\tablesection{Graph}
\hspace{-1mm}$\calG = (\calV,\calE)$  & Directed graph \\
$m$ & Number of edges \\
$n$ & Number of nodes \\
$\calV$ &                 Vertex set; $|\calV|=n$ \\
$\calE$ &                Edge set; $|\calE|=m$ \\
$e=(i,j)\in\calE$  & Edge between nodes $i$ and $j$ \\
$\Incidence \in \Real{n\times m}$    & Incidence matrix of $\calG$ \\
$\Aanc \in \Real{(n-1)\times m}$    & Anchored incidence matrix of $\calG$ \\
$\Laplacian = \Incidence\tran \Incidence$ 
& Laplacian matrix of $\calG$ \\
$\Lanc = \Aanc\tran \Aanc$ 
 & Anchored Laplacian matrix of $\calG$ \\

\\

\tablesection{Real \PGO formulation}
\hspace{-1mm}$\AugIncidence = \Incidence \otimes \eye_2$  & Augmented incidence matrix  \\
$\bar \Aanc = \Aanc \otimes \eye_2$  & Augmented anchored incidence matrix  \\
$\bar \Laplacian = \Laplacian \otimes \eye_2$ 
& Augmented Laplacian matrix \\
$\Wfull \in \Real{4n \times 4n}$ & Real pose graph matrix \\
$\Wanc \in \Real{(4n-2) \times (4n-2)}$ & Real anchored pose graph matrix \\
$p \in \Real{2n}$  & Node positions  \\
$\panc \in \Real{2(n-1)}$  & Anchored node positions  \\
$r \in \Real{2n}$  & Node rotations  \\

\\

\tablesection{Complex \PGO formulation}
\hspace{-1mm}$\Wcpx \in \Complex{(2n-1) \times (2n-1)}$ & Complex anchored pose graph matrix \\
$\pcpx \in \Complex{n-1}$  & Anchored complex node positions  \\
$\rcpx \in \Complex{n}$  & Complex node rotations  \\

\\

\tablesection{Miscellanea}
\hspace{-1mm}$\SOtwo$ & 2D rotation matrices \\
$\aSOtwo$ & Scalar multiple of a 2D rotation matrix \\
$|\calV|$ & Cardinality of the set $\calV$ \\
$\eye_n$   & $n \times n$ identity matrix\\
$\zeros_n$ ($\ones_n$) &  Column vector of zeros (ones) of dimension~$n$\\
$\tr(X)$ & Trace of the matrix $X$\\
\end{tabular}
\vspace{-0cm}
\end{table}



\section{Pose graph optimization in the complex domain}
\label{sec:PGO}


\subsection{Standard PGO}
\label{sec:problemStatement}

\PGO estimates $\nrNodes$ poses
 from $m$ relative pose measurements.
We focus on the planar case, in which 
the $i$-th pose $\pose_i$ is described by the pair $\pose_i \doteq (p_i, R_i)$, where 
 $p_i \!\in\! \Real{2}$ is a position in the plane, and $R_i \!\in\! \SOtwo$ is a planar rotation. 
  The pose measurement between two nodes, say $i$ and $j$, is described by the pair $(\Deltaij,\Rij)$, where 
 $\Deltaij \!\in\! \Real{2}$ and $\Rij \!\in\! \SOtwo$ are the relative position and rotation measurements, respectively.

\begin{figure}[h]
\centering
\includegraphics[scale=0.25]{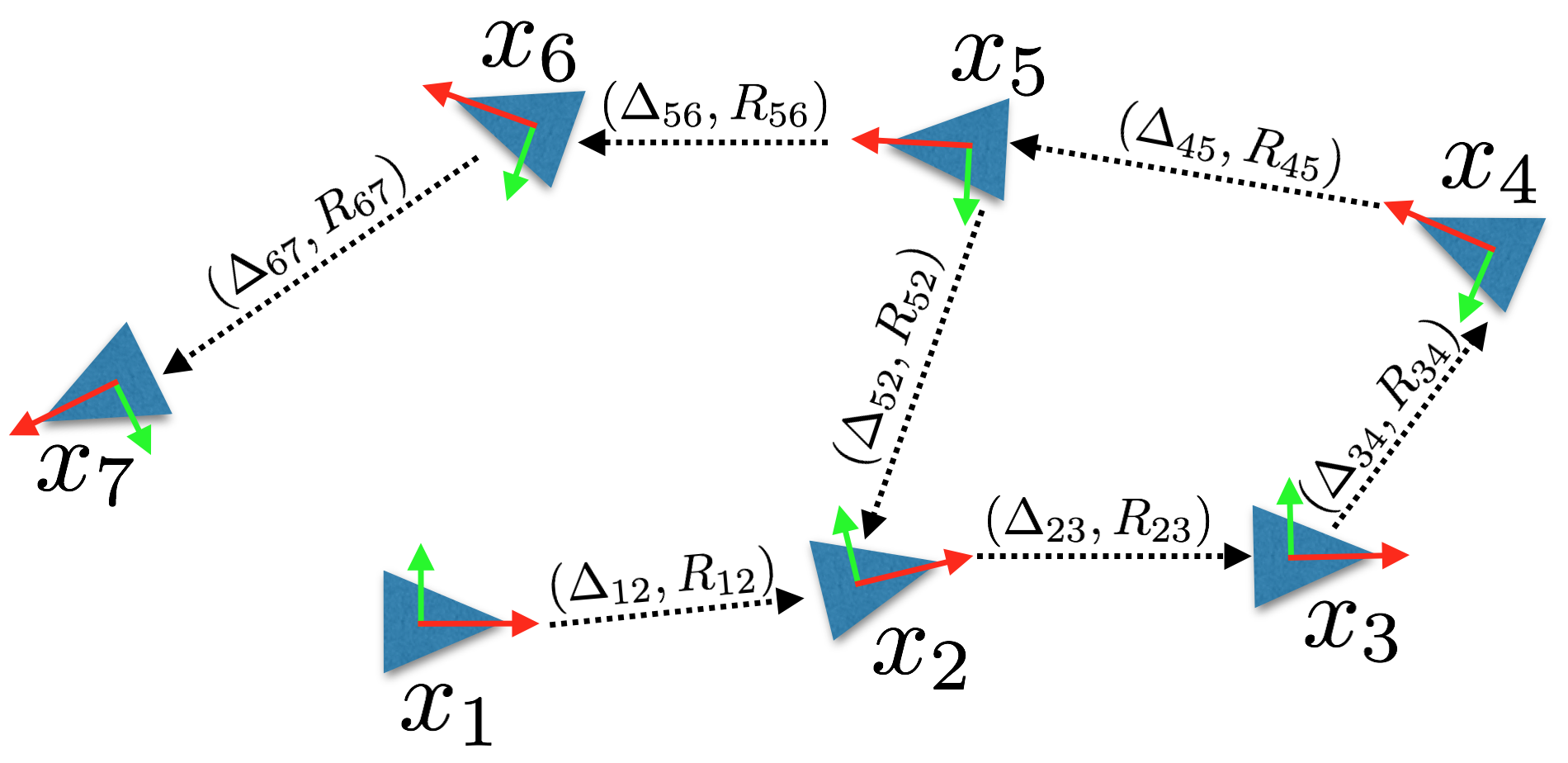} 
\caption{\label{fig:pgo} Schematic representation of \PGOlong:  
the objective is to associate a pose $x_i$ to each node of a directed 
graph, given relative pose measurements $(\Deltaij,\Rij)$ for each edge $(i,j)$ 
in the graph.
}
\end{figure}

 The problem can be visualized as a directed graph $\calG(\calV,\calE)$, where 
  an unknown pose is attached to each node in the set $\calV$, and each edge $(i,j) \in \calE$
   corresponds to a relative pose measurement between nodes $i$ and $j$ (Fig.~\ref{fig:pgo}). 

 In a noiseless case, the measurements satisfy:
\bea
\label{eq:noiselessModel}
\Deltaij = R_i\tran \left( p_j - p_i \right),
\quad&\quad 
\Rij = R_i\tran R_j \;, 
\eea
%
and we can compute the unknown rotations  $\{R_1,\ldots,R_{\nrNodes}\}$
 and positions $\{p_1,\ldots,p_{\nrNodes}\}$  by solving a set of linear equations 
(relations~\eqref{eq:noiselessModel} become linear after rearranging the rotation $R_i$ 
to the left-hand side). 
In absence of noise, the problem admits a unique solution as long as 
one fixes the pose of a node (say $p_1 = \zeros_2$ and $R_1 = \eye_2$) 
and the underling graph is connected. 

In this work we focus on connected graphs, as these are the ones of practical interest
 in \PGO (a graph with $k$ connected components can be split in $k$ subproblems, 
 which can be solved and analyzed independently).

\begin{assumption}[Connected Pose Graph]
\label{ass:connected}
The graph $\calG$ underlying the pose graph optimization problem is 
 connected.
\end{assumption}

In presence of noise, the relations~\eqref{eq:noiselessModel} cannot be met exactly and 
pose graph optimization looks 
for a set of positions $\{p_1,\ldots,p_{\nrNodes}\}$ and rotations 
$\{R_1,\ldots,R_{\nrNodes}\}$ 
that minimize the mismatch with respect to the measurements. 
This mismatch can be quantified by different cost functions. 
We adopt the formulation proposed in \cite{Carlone15icra-verify}:
\beq
\label{eq:PGO0}
\min_{ \{p_i\}, \{R_i\} \in \SOtwo^n } \sum_{(i,j)\in\calE} \| \Deltaij - R_i\tran (\diffp) \|_2^2 + 
\frac{1}{2} \|\Rij - R_i\tran R_j\|_F^2,
\eeq
 where $\|\cdot\|_2$ is the standard Euclidean distance and $\|\cdot\|_F$ is the Frobenius 
 norm. The Frobenius norm $\| R_a - R_b \|$ 
  is a standard measure of distance between 
two rotations $R_a$ and $R_b$, and it is commonly referred to as the \emph{chordal} distance, see, e.g.,~\cite{Hartley13ijcv}.
In~\eqref{eq:PGO0}, we used the short-hand notation $\{p_i\}$ (resp. $\{R_i\}$) to denote the 
set of unknown positions $\{p_1\,\ldots,p_n\}$ (resp. rotations).

Rearranging the terms, problem~\eqref{eq:PGO0} can be rewritten as:
\beq
\label{eq:PGO}
\min_{ \{p_i\}, \{R_i\} \in \SOtwo^n } \sum_{(i,j)\in\calE} \|(\diffp) - R_i \Deltaij \|_2^2 + \frac{1}{2} 
\| R_j - R_i \Rij\|_F^2,
\eeq
where we exploited the fact that the 2-norm is invariant to rotation, i.e., for any vector $v$ and 
any rotation matrix $R$ it holds $\|R v\|_2 = \|v\|_2$.
%
%
Eq.~\eqref{eq:PGO} highlights that the objective is a quadratic function of the unknowns. 

The complexity of the problem stems from the fact that the constraint $R_i \in \SOtwo$ is nonconvex, see, e.g.,~\cite{Saunderson14arxiv}. 
To make this more explicit, we follow the line of~\cite{Carlone15icra-verify}, and use a more convenient representation 
for nodes' rotations.
 Every planar rotation $R_i$ can be written as in~\eqref{eq:SO2},
and is fully defined by the vector
\beq
\label{eq:rparametrization}
r_i = 
\left[
\ba{c}
\cos(\theta_i)  \\
\sin(\theta_i) 
\ea
\right].
\eeq
Using this parametrization and with simple matrix manipulation, \eq~\eqref{eq:PGO} becomes 
(\emph{cf.} with \eq~(11) in \cite{Carlone15icra-verify}):
\bea
\label{eq:PGOr}
\min_{\{p_i\}, \{r_i\}} & \sum_{(i,j)\in\calE} \| (\diffp) - D_{ij}r_i \|_2^2 + \| r_j - \Rij r_i\|_2^2\\
\mbox{s.t.:}    & \|r_i\|^2_2 = 1, \;\; i=1,\ldots,n  \nonumber
\eea
where we defined:

\vspace{-0.2cm}

\beq
\label{eq:Deltaij}
D_{ij} = 
\left[
\begin{array}{cc}
\Delta^x_{ij}     \! &  \! - \Delta^y_{ij} \\
\Delta^y_{ij} \! & \! \Delta^x_{ij} 
\end{array}
\right], \quad (\text{with } \Deltaij \doteq [\Delta^x_{ij}  \; \Delta^y_{ij}]\tran) \ ,
\eeq
and where the constraints $\|r_i\|^2_2 = 1$ specify that we look for vectors 
$r_i$ that represent admissible rotations (i.e., such that $\cos(\theta_i)^2 + \sin(\theta_i)^2 = 1$).

Problem~\eqref{eq:PGOr} is a quadratic problem with quadratic equality constraints. 
The latter are nonconvex, hence computing a
local minimum of~\eqref{eq:PGOr} is hard in general.
There are two problem instances, however, for which it is easy to 
compute a global minimizer, which attains zero optimal cost. 
These two cases are recalled in Propositions~\ref{prop:zeroCostTrees}-\ref{prop:zeroCostBalancedGraphs}.

\begin{proposition}[Zero cost in trees]
\label{prop:zeroCostTrees}
An optimal solution for a \PGO problem in the form~\eqref{eq:PGOr} whose underlying graph is a 
tree attains zero cost.
\end{proposition}

The proof is given in Appendix~\ref{sec:proof:zeroCostTrees}.
Roughly speaking, in a tree, we can build an optimal solution by concatenating the 
relative pose measurements, and this solution annihilates the cost function. 
This comes with no surprises, as the \emph{chords} (i.e., the extra edges, added 
to a spanning tree) are indeed the elements that create redundancy and improve the pose estimate.
However, also for graphs with chords, it is possible to attain the zero cost in 
problem~\eqref{eq:PGOr}. 

\begin{definition}[Balanced pose graph]
\label{def:balancedGraphs}
A pose graph is \emph{balanced} if the pose measurements 
compose to  the  identity along each cycle  in the graph\footnote{We use the somehow standard term ``composition'' 
to denote the group operation for $\SEtwo$. For two poses 
$T_1 \doteq (p_1,R_1)$ and $T_2 \doteq (p_2,R_2)$, the 
composition is $T_1 \cdot T_2 = (p_1 + R_1 p_2, R_1 R_2)$~\cite{Chirikjian12book}. 
Similarly, the \emph{identity} element is $(\zeros_2, \eye_2)$.}\footnote{When composing measurements 
along the loop, edge direction is important: 
for two consecutive edges $(i,k)$ and $(k,j)$ along the loop, the composition is 
$T_{ij} = T_{ik} \cdot T_{kj}$, while if the second edge is in the form $(j,k)$, the 
composition becomes $T_{ij} = T_{ik} \cdot T_{jk}\inv$.}.
\end{definition}

In a balanced pose graph, there exists a configuration 
that explains exactly the measurements, as formalized in the 
following proposition.

\begin{proposition}[Zero cost in balanced pose graphs]
\label{prop:zeroCostBalancedGraphs}
An optimal solution for a balanced pose graph optimization problem  attains zero cost.
\end{proposition}

The proof is given in Appendix~\ref{sec:proof:zeroCostBalancedGraphs}.
 The concept of balanced graph describes a noiseless setup, 
while in real problem instances the measurements do not compose to the identity 
along cycles, because of the presence of noise.

%
We note the following fact, which will be useful in Section~\ref{sec:compactFormulation}.

\begin{proposition}[Coefficient matrices in \PGO]
\label{lem:coefficientMatricesInASOtwo}
Matrices $\Dij, \eye_2, -\eye_2, \Rij$ appearing in ~\eqref{eq:PGOr}  belong to $\aSOtwo$.
\end{proposition}
This fact is trivial, since $\Rij,\eye_2 \in \SOtwo \subset \aSOtwo$ 
(the latter also implies $-\eye_2 \in \aSOtwo$).
 Moreover,  the structure of $\Dij$ in~\eqref{eq:Deltaij}
 clearly falls in the definition of matrices in $\aSOtwo$ given in~\eqref{eq:defaSOtwo}. 

\subsection{Matrix formulation and anchoring}
\label{sec:compactFormulation}

In this section we rewrite the cost function~\eqref{eq:PGOr} 
in a more convenient matrix form. 
The original cost is: 
\beq
\label{eq:costOriginal}
f(p,r) \doteq  \sum_{(i,j)\in\calE} \| (\diffp) - \Dij r_i\|_2^2 + \|r_j - \Rij r_i\|_2^2
\eeq  
where we denote with $p \in \Real{2n}$ and $r  \in \Real{2n}$ the vectors 
stacking all nodes positions and rotations, respectively. 
Now, let $\Incidence \in\Real{m \times n}$
denote the \emph{incidence matrix} of the graph underlying the problem: 
if $(i,j)$ is the $k$-th edge, then
$\Incidence_{ki} = -1$, $\Incidence_{kj} = +1$. Let $\AugIncidence = \Incidence\otimes I_2 \in\Real{2m \times 2n}$, and denote
with $\AugIncidence_k\in\Real{2 \times 2n}$ the $k$-th block row of $\AugIncidence$.
 From the structure of $\AugIncidence$, it follows that $\AugIncidence_k p = \diffp$. 
 Also, we define $\barD\in\Real{2m \times 2n}$ as a block matrix
where the $k$-th block row  $\barD_k\in\Real{2 \times 2n}$ corresponding to the $k$-th edge $(i,j)$
is all zeros, except for a $2\times 2$ block $-\Dij$ in the $i$-th block column.
Using the matrices $\AugIncidence$ and $\barD$, the first sum in~\eqref{eq:costOriginal} can be written as:
\beq
\label{eq:useAD}
\sum_{(i,j)\in\calE} \|(\diffp) - \Dij r_i\|_2^2 = \sum_{k=1}^m \| \AugIncidence_k p  + \barD_k r\|_2^2 = 
\| \AugIncidence p + \barD r \|_2^2
\eeq
Similarly,  we define
 $\barU \in\Real{2m \times 2n}$ as a block matrix
where the $k$-th block row   $\barU_k\in\Real{2 \times 2n}$ corresponding to the $k$-th edge $(i,j)$
is all zeros, except for $2\times 2$ blocks in the $i$-th and $j$-th block columns, which 
are equal to $-\Rij$ and $\eye_2$, respectively. Using $\barU$, the second sum in~\eqref{eq:costOriginal} becomes:
\beq
\label{eq:useU}
\sum_{(i,j)\in\calE} \| r_j - \Rij r_i\|_2^2 = \sum_{k=1}^m \| \barU_k r \|_2^2 = 
\| \barU r \|_2^2
\eeq
Combining~\eqref{eq:useAD} and~\eqref{eq:useU}, the cost in~\eqref{eq:costOriginal} becomes:
\bea
\label{eq:costCompact1}
f(p,r) = \left\| \matTwo{\AugIncidence & \barD \\ \zeros & \barU} \vect{p \\ r} \right\|_2^2 &=& 
 \vect{p \\ r}\tran 
 \matTwo{\AugIncidence\tran \AugIncidence & \AugIncidence\tran \barD\\
  \barD\tran \AugIncidence  & \barD\tran \barD + \barU\tran \barU} 
 \vect{p \\ r}   \nonumber \\
 &=& \vect{p \\ r}\tran 
 \matTwo{
 \AugLaplacian & \AugIncidence\tran \barD \\ 
 \barD\tran \AugIncidence &  \bar Q} 
 \vect{p \\ r},
\eea 
where we defined $\bar Q \doteq \barD\tran \barD + \barU\tran \barU$ 
and $\AugLaplacian \doteq \AugIncidence\tran \AugIncidence$, to simplify notation. Note that, since $\AugIncidence \doteq \Incidence \otimes I_2$, 
it is easy to show that $\AugLaplacian = \Laplacian \otimes I_2$, where $\Laplacian \doteq \Incidence\tran \Incidence$ is the Laplacian matrix of the graph underlying 
the problem.
A pose graph optimization instance is thus completely defined by the matrix
\beq
\label{eq:Wdef}
\Wfull \doteq \matTwo{
 \AugLaplacian & \AugIncidence\tran \barD \\ 
 \barD\tran \AugIncidence &  \bar Q} \in \Real{4n \times 4n}
\eeq
From~\eqref{eq:costCompact1}, $\Wfull$ can be easily seen to be symmetric and positive semidefinite.
Other useful properties of $\Wfull$ are stated in the next proposition.
%

\begin{proposition}[Properties of $\Wfull$]
\label{prop:propertiesW}
Matrix $\Wfull$ in~\eqref{eq:Wdef} is positive semidefinite, and
\begin{enumerate}
\item has at least two eigenvalues in zero;
\item is composed by $2\times 2$ blocks $[\Wfull]\ij$, and each block is a multiple 
of a rotation matrix, i.e., $[\Wfull]\ij\in\aSOtwo$, $\forall i,j=1,\ldots,2n$. 
Moreover, the diagonal blocks of $\Wfull$ are nonnegative multiples of the identity matrix, i.e., 
$[\Wfull]\ii = \alpha\ii \eye_2$, $\alpha\ii \geq 0$.
\end{enumerate}
\end{proposition}
A formal proof of Proposition~\ref{prop:propertiesW} is given in~Appendix~\ref{sec:proof:propertiesW}. 
An intuitive explanation of the second claim follows from the fact that (i) $\Wfull$ contains sums and products of the 
matrices in the original formulation~\eqref{eq:PGOr} (which are in $\aSOtwo$ according to 
Lemma~\ref{lem:coefficientMatricesInASOtwo}), and (ii) the set $\aSOtwo$ is closed under 
 matrix sum and product (Section~\ref{sec:alphaSO2}). 

 The presence of two eigenvalues in zero has a very natural geometric interpretation: the cost function 
 encodes inter-nodal measurements, hence it 
 is invariant to global translations of node positions, i.e., $f(p,r) = 
 f(p + p_a, r)$, where $p_a \doteq (\ones_n \otimes \eye_2) 
 a = [a\tran \; \ldots \; a\tran]\tran$ ($n$ copies of $a$), with 
 $a \in \Real{2}$. Algebraically, this translates to the fact that 
 the matrix $(\ones_n \otimes \eye_2) \in \Real{2n \times 2}$ is in the null 
 space of the augmented incidence matrix $\AugIncidence$, 
 which also implies a two dimensional null space for $\Wfull$. 
%

\paragraph{Position anchoring} In this paper we show that the duality properties in pose graph 
optimization are tightly coupled with the spectrum of the matrix $\Wfull$. We are particularly 
interested in the eigenvalues at zero, and from this perspective it is not convenient 
to carry on the two null eigenvalues of $\Wfull$ (claim 1 of Proposition~\ref{prop:propertiesW}), which 
are always present, and are due to an intrinsic observability issue. 

We remove the translation ambiguity by fixing the 
position of an arbitrary node. Without loss of generality, we     
fix the position $p_1$ of the first node to the origin, i.e., $p_1 = 0_2$. 
This process is commonly called \emph{anchoring}.
%
%
Setting $p_1=0$ is equivalent to removing the corresponding columns and rows from $\Wfull$, leading to 
the following ``anchored'' \PGO problem:
\beq
f(r,\panc) = \vect{\zeros_2 \\ \panc \\ r}\tran \Wfull \vect{\zeros_2 \\ \panc \\ r} = 
\vect{\panc \\ r}\tran \Wanc \vect{\panc \\ r}
\eeq
where $\panc$ is the vector $p$ without its first two-elements vector $p_1$, 
and $\Wanc$ is obtained from $\Wfull$ by 
removing the rows and the columns corresponding to $p_1$. The structure of $\Wanc$ is as follows:

\beq
\label{eq:WancDef}
\Wanc=
\matTwo{ \bar \Aanc\tran \bar \Aanc & \bar \Aanc\tran \barD \\ \barD\tran \bar \Aanc & \bar Q}
\doteq 
\matTwo{ \bar \Lanc & \bar {\Sanc} \\ \bar\Sanc\tran & \bar Q}
\eeq
where $\bar \Aanc = \Aanc \otimes \eye_2$, and $\Aanc$ is the \emph{anchored} (or \emph{reduced}) incidence matrix, 
 obtained by removing the first column from $\Incidence$, see, e.g.,~\cite{Carlone14ijrr}.   
On the right-hand-side of~\eqref{eq:WancDef} we defined 
$\bar \Sanc \doteq \bar \Aanc\tran \barD$ and $\bar \Lanc \doteq \bar \Aanc\tran \bar \Aanc$. 

We call $\Wanc$ the \emph{real (anchored) pose graph matrix}. $\Wanc$ is still symmetric 
and positive semidefinite (it is a principal submatrix of a positive semidefinite matrix).
Moreover, since $\Wanc$ is obtained by removing 
 a $2 \times 4n$ block row and a $4n \times 2$ block column from 
 $\Wfull$, it is still composed by $2\times 2$ matrices in $\aSOtwo$, 
 as specified in the following remark. 

\begin{remark}[Properties of $\Wanc$]
\label{rmk:propertiesWanc}
The positive semidefinite matrix $\Wanc$ in~\eqref{eq:WancDef} 
 is composed by $2\times 2$ blocks $[\Wanc]\ij$, that are 
 such that $[\Wanc]\ij\in\aSOtwo$, $\forall i,j=1,\ldots,2n-1$. 
Moreover, the diagonal blocks of $\Wanc$ are nonnegative multiples of the identity matrix, i.e., 
$[\Wanc]\ii = \alpha\ii \eye_2$, $\alpha\geq 0$.
\end{remark}

\noindent
After anchoring, our \PGO problem becomes:
\bea
\label{eq:PGOanc}
f^\star = \min_{\panc,r} & \vect{\panc \\ r}\tran \Wanc \vect{\panc \\ r} \\
\mbox{s.t.:}    & \|r_i\|^2_2 = 1,  &  i=1,\ldots,n \nonumber
\eea
%




\subsection{To complex domain}
\label{sec:complex}

In this section we reformulate problem~\eqref{eq:PGOanc}, in which the decision variables 
are real vectors, into a problem in complex variables. 
The main motivation for this 
choice is that the real representation~\eqref{eq:PGOanc} is somehow redundant: 
as we will show in Proposition~\ref{prop:W_VS_Wtilde}, each eigenvalue 
of $\Wanc$ is repeated twice (multiplicity 2), while the complex representation does not 
have this redundancy, making analysis easier. 
In the rest of this paper, any quantity marked with a tilde ($\tilde \cdot$) lives in the complex domain $\Complex{}$.

\newcommand{\tempVect}{v}
\newcommand{\tempMat}{Z}

Any real vector $\tempVect\in\Real{2}$ can be represented by a complex number $\tilde \tempVect = \eta \e^{\j \varphi}$,
where $\j^2 = -1$ is the \emph{imaginary unit}, $\eta = \|\tempVect\|_2$ and $\varphi$ is the angle that $\tempVect$ forms with the horizontal axis.
We use the operator $(\cdot)^\vee$ to map a 2-vector to the 
corresponding complex number, $\tilde \tempVect = \tempVect^\vee$. 
When convenient, we adopt the notation 
$\tempVect \sim \tilde \tempVect$, meaning that $\tempVect$ and $\tilde \tempVect$ are the vector and the complex representation 
of the same number.

 The action of a real $2\times 2$ matrix $\tempMat$ on a vector $\tempVect\in\Real{2}$ 
cannot be represented, in general, as a scalar multiplication between complex numbers.
However, if $\tempMat \in \aSOtwo$, this is possible. To show this, assume that 
$\tempMat = \alpha R(\theta)$, where
$R(\theta)$ is a counter-clockwise rotation of angle $\theta$.
Then, 
\beq
\label{eq:complexA}
\tempMat \, \tempVect= \alpha R(\theta)\tempVect \sim \tilde z \; \tilde \tempVect,\quad
\mbox{where } \tilde z = \alpha \e^{\j \theta}. 
\eeq
With slight abuse of notation we extend the operator $(\cdot)^\vee$ 
to 
$\aSOtwo$, such that, given $Z = \alpha R(\theta) \in \aSOtwo$,  then $Z^\vee = \alpha \e^{\j \theta} \in \Complex{}$.
%
\toChange{By inspection, one can also verify the following relations between
the sum and product of two matrices $\tempMat_1,\tempMat_2 \in \aSOtwo$ and their 
complex representations $\tempMat_1^\vee,\tempMat_2^\vee \in \Complex{}$:}
\beq
\label{eq:complexMatrices}
\begin{array}{llll}
(\tempMat_1 \; \tempMat_2)^\vee = \tilde \tempMat_1^\vee \; \tempMat_2^\vee
\qquad \qquad
(\tempMat_1 + \tempMat_2)^\vee = \tempMat_1^\vee + \tempMat_2^\vee .
\end{array}
\eeq 
We next discuss how to apply the machinery introduced so far to reformulate problem~\eqref{eq:PGOanc} 
in the complex domain.
The variables in problem~\eqref{eq:PGOanc} are the vectors $\panc \in \Real{2(n-1)}$ and $r \in \Real{2n}$ that are composed by 
2-vectors, i.e.,  $\panc = [\panc_1\tran,\ldots,\panc_{n-1}\tran]\tran$ and $r = [r_1\tran,\ldots,r_n\tran]\tran$, 
where $\panc_i, r_i \in \Real{2}$. Therefore, we define the 
\emph{complex 
positions} and the \emph{complex rotations}:
\beq
\label{eq:complexVectors}
\begin{array}{llllll}
\pcpx &=& [\pcpx_1,\ldots,\pcpx_{n-1}]\tran & \in &  \Complex{n-1}, & \text{where: } \pcpx_i = \panc_i^\vee \\
\rcpx &=& [\rcpx_1,\ldots,\rcpx_n]\tran     & \in & \Complex{n},     & \text{where: } \rcpx_i = r_i^\vee
\end{array}
\eeq 
Using the complex parametrization~\eqref{eq:complexVectors}, the constraints in~\eqref{eq:PGOanc} become:
\beq
\label{eq:constraintsComplex}
|\rcpx_i|^2=1, \quad i=1,\ldots,n.
\eeq
Similarly, we would like to rewrite the objective as a function of $\pcpx$ and $\rcpx$.
This re-parametrization is formalized in the following proposition, whose proof is given in Appendix~\ref{sec:proof:costComplex}.
\begin{proposition}[Cost in the complex domain]
\label{prop:costComplex}
For any pair $(\panc,r)$, the cost function in~\eqref{eq:PGOanc} is such that:
\beq
f(\panc,r) = \vect{\panc \\ r}\tran \Wanc \vect{\panc \\ r} = 
\vect{\pcpx \\ \rcpx}\tran \Wcpx \vect{\pcpx \\ \rcpx} 
\eeq
where the vectors $\pcpx$ and $\rcpx$ are built from $\panc$ and $r$ as in~\eqref{eq:complexVectors}, 
and the matrix $\Wcpx \in \Complex{(2n-1) \times (2n-1)}$ is such that $\Wcpx\ij = [\Wanc]\ij^\vee$, with $i,j=1,\ldots,2n-1$.
\end{proposition}

\begin{remark}[Real diagonal entries for $\Wcpx$]
\label{rmk:realDiagonalWcpx}
According to Remark~\ref{rmk:propertiesWanc}, the diagonal blocks 
 of $\Wanc$ are multiples of the identity matrix, i.e., $[\Wanc]\ii = \alpha\ii \eye_2$.
 Therefore, the diagonal elements of $\Wcpx$ are $\Wcpx\ii = [\Wanc]\ii^\vee = \alpha\ii \in \Real{}$.
\end{remark}

\noindent
Proposition~\ref{prop:costComplex} 
 enables us to rewrite problem~\eqref{eq:PGOanc} as:
\bea
\label{eq:PGOcpx}
f^\star = \min_{\pcpx, \rcpx} & \vect{\pcpx \\ \rcpx}\tran \Wcpx \vect{\pcpx \\ \rcpx} \\
\mbox{s.t.:}    & |\rcpx_i|^2 = 1,  &  i=1,\ldots,n. \nonumber
\eea
We call $\Wcpx$ the \emph{complex (anchored) pose graph matrix}. 
Clearly, the matrix $\Wcpx$ preserves the same block 
structure of $\Wanc$ in~\eqref{eq:WancDef}:
\beq
\label{eq:Wcpx}
\Wcpx \doteq \left[\ba{cc} \Lcpx & \Scpx \\  \Scpx^* & \Qcpx \ea\right]
\eeq
where $\Scpx^*$ is the conjugate transpose of $\Scpx$, and 
$\Lcpx \doteq \Aanc\tran \Aanc$ where $\Aanc$ is the anchored incidence matrix. 
In Section~\ref{sec:duality} we apply Lagrangian duality to the problem~\eqref{eq:PGOcpx}.
Before that, we provide results to 
characterize the spectrum of the matrices $\Wanc$ and $\Wcpx$, drawing connections with the 
recent literature on unit gain graphs,~\cite{Reff:11}.

\subsection{Analysis of the real and complex pose graph matrices} 
\label{sec:complexAnalisys}

In this section we take a closer look at the structure and the properties 
of the real and the complex pose graph matrices $\Wanc$ and $\Wcpx$. 
%
In analogy with~\eqref{eq:costCompact1} and~\eqref{eq:WancDef}, 
we write $\Wcpx$ as
\beq
\label{eq:WcpxFactorized}
\Wcpx = \left[\ba{cc} \Aanc\tran \Aanc & 
\Aanc\tran \tilde D \\  (\Aanc\tran \tilde D)^* & 
\IncidenceUGGcpx^* \IncidenceUGGcpx + \tilde D^* \tilde D \ea\right]
 = 
 \left[\ba{cc} \Aanc & 
\tilde D \\  \zeros & 
\IncidenceUGGcpx \ea\right]^*
 \left[\ba{cc} \Aanc & 
\tilde D \\  \zeros & 
\IncidenceUGGcpx \ea\right]
\eeq
where $\IncidenceUGGcpx \in \Complex{m \times n}$ and $\tilde D \in \Complex{m \times n}$ 
are the ``complex versions'' of $\barU$ and $\barD$ in~\eqref{eq:costCompact1}, 
i.e., they are obtained as $\IncidenceUGGcpx\ij = [\barU]\ij^\vee$ 
and $\tilde D\ij = [\barD]\ij^\vee$, $\forall i,j$.

The factorization~\eqref{eq:WcpxFactorized} is interesting, as it
allows to identify two important matrices that compose $\Wcpx$: 
the first is $\Aanc$, the anchored incidence matrix that we introduced earlier;
the second is $\IncidenceUGGcpx$ which is a generalization of the incidence matrix, 
as specified by Definition~\ref{def:unitGain} and Lemma~\ref{lem:unitGain} in the following.
Fig.~\ref{fig:incidence} reports the matrices $\Aanc$ and $\IncidenceUGGcpx$ for a toy example 
with four poses.


\newcommand{\ug}[1]{\e^{\j \theta_{#1}}}
\begin{figure}[t]
\begin{minipage}{\textwidth}
\begin{tabular}{cc}
\begin{minipage}{6cm}%
\centering
\includegraphics[scale=0.16]{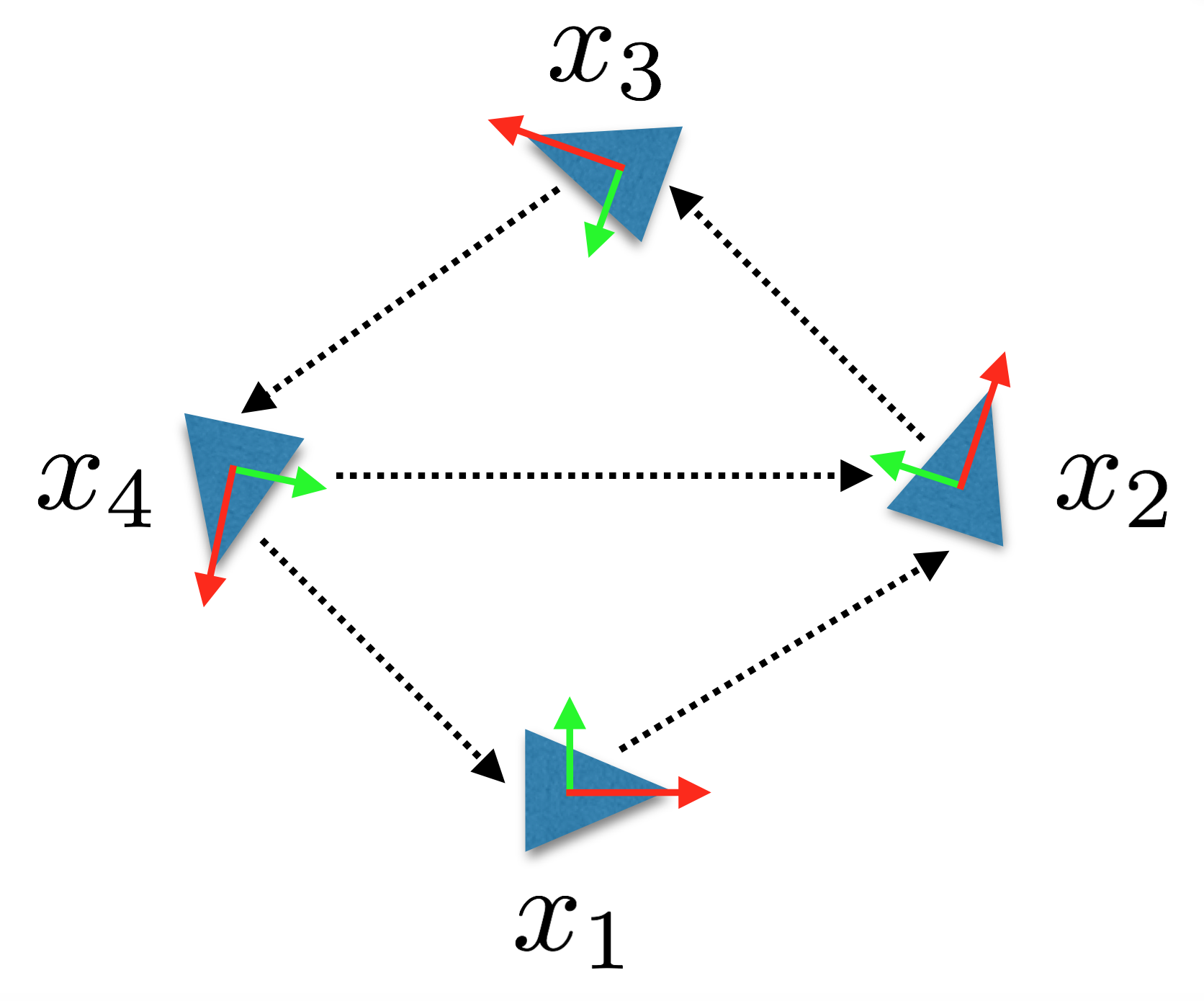} 
\end{minipage}%
&
\begin{minipage}{6cm}%
\small
\centering
Incidence matrix:
\beq
\Incidence = 
\left[
\begin{array}{rrrrr}
-1 & +1 & 0 & 0 \\
0 & -1 & +1 & 0 \\
0 & 0 & -1 & +1 \\
+1 & 0 & 0 & -1 \\
0 & +1 & 0 & -1
\end{array}
\right] 
\begin{array}{r}
(1,2) \\
(2,3) \\
(3,4) \\
(4,1) \\
(4,2) 
\end{array}\nonumber
\eeq 
\end{minipage}%
\end{tabular}%

\begin{tabular}{cc}
\begin{minipage}{5cm}%
\small
\centering
\hspace{0.2cm} Anchored Incidence matrix:
\beq
\Aanc = 
\left[
\begin{array}{rrrr}
+1 & 0 & 0 \\
-1 & +1 & 0 \\
0 & -1 & +1 \\
0 & 0 & -1 \\
+1 & 0 & -1
\end{array}
\right]  \nonumber
\eeq
\end{minipage}%
&
\begin{minipage}{7cm}%
\small
\centering
\hspace{1cm}Complex Incidence matrix:
\beq
\IncidenceUGGcpx = 
\left[
\begin{array}{ccccc}
-\ug{12} & +1 & 0 & 0 \\
0 & -\ug{23} & +1 & 0 \\
0 & 0 & -\ug{34} & +1 \\
+1 & 0 & 0 & -\ug{41}  \\
0 & +1 & 0 & -\ug{42}
\end{array}
\right] \nonumber
\eeq
\end{minipage}%
\end{tabular}%
\end{minipage}%
\caption{\label{fig:incidence} Example of incidence matrix, anchored incidence matrix, and complex
incidence matrix, for the toy \PGO problem on the top left. 
If $R\ij = R(\theta\ij)$ is the relative rotation measurement associated 
to edge $(i,j)$, then the matrix $\IncidenceUGGcpx$ can be seen as the incidence 
matrix of a unit gain graph with gain $\ug{ij}$ associated to each edge $(i,j)$.
}
\end{figure}

\begin{definition}[Unit gain graphs]
\label{def:unitGain}
A \emph{unit gain graph} (see, e.g.,~\cite{Reff:11}) is a graph in which to each orientation 
of an edge $(i,j)$ is assigned a complex number $\tilde z\ij$ (with $|\tilde z\ij|=1$), 
which is the inverse of the complex number $\frac{1}{\tilde z\ij}$ assigned to the opposite orientation $(j,i)$. 
Moreover, a \emph{complex incidence matrix} of a unit gain graph 
is a matrix in which each row 
corresponds to an edge and 
the row corresponding to edge~$e=(i,j)$ has $-\tilde z\ij$ on the~$i$-th column, 
$+1$ on the~$j$-th colum, and zero elsewhere.
\end{definition}

Roughly speaking, a unit gain graph describes a problem in which we 
can ``flip'' the orientation of an edge by inverting the 
corresponding complex weight. 
To understand what this property means in our context, recall the definition~\eqref{eq:useU}, and consider the following chain of equalities:
\beq
\| \barU r \|_2^2 = \sum_{(i,j)\in\calE} \| r_j - \Rij r_i\|_2^2 = \sum_{(i,j)\in\calE} \| r_i - \Rij\tran r_j \|_2^2
\eeq
which, written in the complex domain become:
\bea
\label{eq:unitGainEdge}
\| \IncidenceUGGcpx \rcpx \|_2^2 \!=\! \sum_{(i,j)\in\calE} | \rcpx_j - \e^{\j \theta\ij} \rcpx_i |^2 
\!=\! \sum_{(i,j)\in\calE} | \rcpx_i - \e^{- \j \theta\ij} \rcpx_j |^2 
 \!=\! \sum_{(i,j)\in\calE} | \rcpx_i -\frac{1}{\e^{\j \theta\ij}} \rcpx_j |^2 
\eea
\eq~\eqref{eq:unitGainEdge} essentially says that the term $\| \IncidenceUGGcpx \rcpx \|_2^2$ does not change 
if we flip the orientation of an edge and invert the relative rotation measurement. 
The proof of the following lemma is straightforward from~\eqref{eq:unitGainEdge}.

\begin{lemma}[Properties of $\IncidenceUGGcpx$]
\label{lem:unitGain}
Matrix  $\IncidenceUGGcpx$ is a complex incidence matrix of a unit gain graph with 
weights $R\ij^\vee \!=\! \e^{\j \theta\ji}$ associated to each edge $(i,j)$.
\end{lemma}

Our interest towards unit gain graphs is motivated by the recent results in \cite{Reff:11} on the spectrum 
of the incidence matrix of those graphs. Using these results, 
we can characterize the presence of eigenvalues in zero for 
the matrix $\Wcpx$, as specified in the following 
proposition (proof in Appendix~\ref{sec:proof:szepWtilde}).

\begin{proposition}[Zero eigenvalues in $\Wcpx$]
\label{prop:szepWtilde}
The complex anchored pose graph matrix $\Wcpx$ has a single eigenvalue in zero if and only if 
the pose graph is balanced or is a tree.
\end{proposition}


Besides analyzing the spectrum of $\Wcpx$, it is of interest 
to understand how the complex matrix $\Wcpx$ relates to the 
real matrix $\Wanc$.
The following proposition states that there is 
a tight correspondence between the 
eigenvalues of the real pose graph matrix $\Wanc$ and its 
complex counterpart $\Wcpx$. 

\begin{proposition}[Spectrum of complex graph matrices]
\label{prop:W_VS_Wtilde}
The $2(2n-1)$ eigenvalues of $\Wanc$ are the $2n-1$ eigenvalues of $\Wcpx$, repeated twice.	
\end{proposition}
See Appendix~\ref{sec:proof:W_VS_Wtilde} for a proof.

\section{Lagrangian duality in \PGO}
\label{sec:duality}

\toCheck{
In the previous section we wrote the \PGO problem in complex variables 
as per eq.~\eqref{eq:PGOcpx}. In the following, we refer to this 
problem as the \emph{primal} \PGO problem, that, defining $\xcpx \doteq [\pcpx\tran \; \rcpx\tran]\tran$, 
can be written in compact form as
}
\bea
\label{eq:primal}
\begin{array}{ccccc}
f^\star= & \displaystyle \min_{\xcpx} & \xcpx\tran \tilde W \xcpx && \text{(Primal problem)}\\
		 & \mbox{s.t.:} & |\xcpx_i|^2 = 1, & i=n,\ldots,2n-1, &
\end{array}
\eea
%
%
%

\toCheck{
In this section we derive the Lagrangian dual of~\eqref{eq:primal}, which 
is given in Section~\ref{sec:dual}. Then, in Section~\ref{sec:sdp}, we discuss an SDP relaxation 
of~\eqref{eq:primal}, that can be interpreted as the dual of the dual problem.
Finally, in Section~\ref{sec:analysis} we analyze the properties of the dual problem, 
and discuss how it relates with the primal \PGO problem.}
%

\subsection{The dual problem}
\label{sec:dual}
The Lagrangian of the primal problem~\eqref{eq:primal} is
%
\beas
\label{eq:lagrangian}
\lagrangian(\xcpx,\lam) &=& \xcpx\tran \tilde W \xcpx  + \sum_{i=1}^n \lam_i(1 - |\xcpx_{n+i-1}|^2) 
\eeas
where $\lambda_i \in \Real{}$, $i=1,\ldots,n$, are the \emph{Lagrange multipliers} (or \emph{dual variables}).
Recalling the structure of $\Wcpx$ from~\eqref{eq:Wcpx}, 
the Lagrangian becomes: 
\beas
\lagrangian(\xcpx,\lam) \;=\; 
\xcpx\tran
\matTwo{\Lcpx & \Scpx \\ \Scpx^* & \Qcpx(\lam)}
\xcpx + \sum_{i=1}^n \lam_i
\;=\; 
\xcpx\tran
\Wcpx(\lam)
\xcpx + \sum_{i=1}^n \lam_i,
\eeas
where
for notational convenience we defined
\[
\Qcpx(\lam) \doteq \Qcpx - \diag(\lam_1 ,\ldots,\lam_n), 
\qquad 
\Wcpx(\lam) \doteq \matTwo{\Lcpx & \Scpx \\ \Scpx^* & \Qcpx(\lam)}
\]
The {\em dual function} $d:\Real{n}\to\Real{}$ 
is the infimum  
of the Lagrangian with respect to $\xcpx$:
\bea
\label{eq:dualFunction}
d(\lam) =  \inf_{\xcpx} \; \lagrangian(\xcpx,\lam) 
=
\inf_{\xcpx} \;
\xcpx\tran
\Wcpx(\lam)
\xcpx + \sum_{i=1}^n \lam_i,
\eea
For any choice of $\lam$ the dual function provides a lower bound on the optimal value of 
the primal problem~\cite[Section 5.1.3]{Boyd04book}. Therefore,
the \emph{Lagrangian dual problem} looks for a \emph{maximum} of the dual function over $\lam$: 
\bea
\label{eq:dual0}
d^\star \doteq \max_{\lam} \; d(\lam)  = 
\max_{\lam} \; 
\inf_{\xcpx} \;
\xcpx\tran
\Wcpx(\lam)
\xcpx + \sum_{i=1}^n \lam_i,
\eea
The infimum over $\xcpx$ of $\lagrangian(\xcpx,\lam)$ 
drifts to $-\infty$ unless $\Wcpx(\lam) \succeq 0$. Therefore we 
can safely restrict the maximization to vectors $\lam$ that are such that $\Wcpx(\lam) \succeq 0$; 
 these are called \emph{dual-feasible}. Moreover, 
at any dual-feasible $\lam$, the $\xcpx$ minimizing the 
Lagrangian are those that make $\xcpx\tran \Wcpx(\lam) \xcpx = 0$.
Therefore,~\eqref{eq:dual0} reduces to the following {\em dual problem} 
\bea
\label{eq:dual}
\begin{array}{cccc}
d^\star = & \displaystyle \max_{\lam} &  \sum_i \lam_i, & \qquad \text{(Dual problem)} \\
 & \mbox{s.t.:} & \Wcpx(\lam) \succeq 0. & 
\end{array}
\eea
The importance of the dual problem is twofold. 
 First, it holds that
\beq
\label{eq:weakDuality}
d^\star \leq f^\star 
\eeq
This property is called~\emph{weak duality}, see, e.g.,~\cite[Section 5.2.2]{Boyd04book}.
For particular problems 
 the inequality~\eqref{eq:weakDuality} becomes an equality, and
in such cases we say
that \emph{strong duality} holds.
%
Second, since $d(\lam)$ is concave (minimum of affine functions), 
the dual problem~\eqref{eq:dual} is always convex in $\lam$, 
regardless the convexity properties of the primal problem.
The dual \PGO problem~\eqref{eq:dual} is a semidefinite program (SDP).

For a given $\lam$, we denote by $\calX(\lam)$ the set of $\xcpx$ 
 that attain the optimal value in problem~\eqref{eq:dualFunction}, if any:
 \[
 \calX(\lam) \doteq \{ \xcpx_\lam \in \Complex{\dimWcpx} \;:\; \lagrangian(\xcpx_\lam,\lam) = \min_{\xcpx} \; \lagrangian(\xcpx,\lam)
  =  \min_{\xcpx} \xcpx\tran \Wcpx(\lam) \xcpx \}
 \]
 %
 %
Since we already observed that for any dual-feasible $\lam$ 
the points $\xcpx$ that minimize the Lagrangian are such that $\xcpx\tran \Wcpx(\lam) \xcpx = 0$, 
it follows that: 
\beq
\label{eq:calX2}
\calX(\lam) = \{\xcpx \in\Complex{\dimWcpx}:  \Wcpx(\lam) \xcpx \!=\! 0  \} = \rm{Kernel}(\Wcpx(\lam)), \quad \mbox{for $\lam$ dual-feasible.}
\eeq
%
\toCheck{The following result ensures that if a vector in $\calX(\lam)$ 
is feasible for the primal problem, then it is also an optimal solution for the \PGO problem.} 
\begin{theorem}
\label{thm:Everett}
Given  $\lam\in\Real{n}$, if an $\xcpx_\lam\in\calX(\lam)$ is primal feasible, then $\xcpx_\lam$ is primal optimal; moreover,
$\lam$ is dual optimal, and the duality gap is zero.
\end{theorem}
A proof of this theorem is given in Appendix~\ref{sec:thm:Everett}.

\subsection{SDP relaxation and the dual of the dual}
\label{sec:sdp}

We have seen that a lower bound $d^\star$ on the optimal value $f^\star$ of the primal~\eqref{eq:primal}
can be obtained by solving the Lagrangian dual problem~\eqref{eq:dual}.
Here, we outline another, direct, relaxation method to obtain such bound.

Observing  that $\xcpx\tran \Wcpx \xcpx = \tr( \Wcpx \xcpx \xcpx\tran )$, we rewrite~\eqref{eq:primal} equivalently as
\bea
f^\star= & \displaystyle \min_{\Xcpx,\xcpx} & \tr \Wcpx \Xcpx  \label{eq:primal_sdp_x} \\
&  \subt & \tr E_i \Xcpx = 1, \quad  i=n,\ldots,\dimWcpx, \nonumber \\
&  & \Xcpx = \xcpx\xcpx\tran. \nonumber
\eea
where $E_i$ is a matrix that is zero everywhere, except for the $i$-th diagonal element, 
which is one.
The condition  $\Xcpx = \xcpx\xcpx\tran$ is equivalent to (i) $\Xcpx \succeq 0$ and (ii) $\Xcpx$ has rank one.
Thus, (\ref{eq:primal_sdp_x}) is rewritten by eliminating $\xcpx$ as
\bea
f^\star = & \displaystyle \min_{\Xcpx} & \tr \Wcpx \Xcpx  \label{eq:primal_sdp} \\
& \subt & \tr E_i \Xcpx = 1,  \quad i=n,\ldots,\dimWcpx, \nonumber \\
& & \Xcpx \succeq 0 \nonumber \\
& & \mbox{rank}(\Xcpx) =1. \nonumber
\eea
Dropping the rank constraint, which is non-convex, we obtain the following  SDP relaxation 
(see, e.g.,~\cite{Zhang:00}) of the primal problem:
\beq
\begin{array}{ccccc}
s^\star= & \displaystyle \min_{\Xcpx} & \tr \Wcpx \Xcpx  & & \label{eq:primal_sdp_rel0} \\
          & \subt & \tr E_i \Xcpx = 1, & i=n,\ldots,\dimWcpx, &  \\
& & \Xcpx \succeq 0  &
\end{array}
\eeq
which we can also rewrite as
\bea
\begin{array}{ccccc}
s^\star= & \displaystyle \min_{\Xcpx} & \tr \Wcpx \Xcpx & & \hspace{-0.1cm}\text{(SDP relaxation)}\hspace{-1cm}\label{eq:primal_sdp_rel} \\
& \subt & \Xcpx\ii = 1, & i=n,\ldots,\dimWcpx,  \\
& & \Xcpx \succeq 0  
\end{array}
\eea
where $\Xcpx\ii$ denotes the $i$-th diagonal entry in $\Xcpx$.
Obviously, $s^\star \leq f^\star$, since the feasible set of~\eqref{eq:primal_sdp_rel} contains that of~\eqref{eq:primal_sdp}.
One may then ask what is the relation between 
the Lagrangian dual and the SDP relaxation of problem~\eqref{eq:primal_sdp_rel}:
the answer is that the former is the dual of the latter hence,  under constraint qualification, it holds 
that $s^\star = d^\star$, i.e., the SDP relaxation and the Lagrangian dual approach yield the {\em same} lower bound on $f^\star$. 
This is formalized in the following proposition.

\begin{proposition}
\label{prop_dualrel}
The Lagrangian dual of problem~\eqref{eq:primal_sdp_rel} is problem~\eqref{eq:dual}, and vice-versa.
Strong duality holds between these two problems, i.e., $d^\star=s^\star$.
Moreover, if the optimal solution $\Xcpx^\star$ of~\eqref{eq:primal_sdp_rel} has rank one,
then $s^\star = f^\star$, and hence $d^\star = f^\star$.
\end{proposition}
\toCheck{
{\bf Proof.}
The fact that the SDPs~\eqref{eq:primal_sdp_rel} and~\eqref{eq:dual} 
are related by duality can be found in standard textbooks (e.g. \cite[Example 5.13]{Boyd04book});
moreover, since these are convex programs, under constraint qualification, the duality gap is zero, i.e., $d^\star=s^\star$.
To prove that $\mbox{rank}(\Xcpx^\star)=1 \Rightarrow s^\star = d^\star = f^\star$, 
we observe that (i) $\tr \Wcpx \Xcpx^\star \doteq s^\star \leq f^\star$ since~\eqref{eq:primal_sdp_rel} 
is a relaxation of~\eqref{eq:primal_sdp}. However, when $\mbox{rank}(\Xcpx^\star)=1$, 
 $\Xcpx^\star$ is feasible for problem~\eqref{eq:primal_sdp_rel}, hence, by optimality of 
 $f^\star$, it holds (ii) $f^\star \leq f(\Xcpx^\star) = \tr \Wcpx \Xcpx^\star$. 
 Combining (i) and (ii) we prove that, when $\mbox{rank}(\Xcpx^\star)=1$, 
 then $f^\star = s^\star$, which 
  also implies $f^\star = d^\star$.
\qed
}


\toCheck{
To the best of our knowledge this is the first time in which the SDP relaxation has been proposed 
to solve \PGO. For the rotation subproblem, SDP relaxations 
have been proposed in~\cite{Singer11siam,Saunderson14cdc,Fredriksson12accv}.
According to Proposition~\ref{prop_dualrel}, 
one advantage of the SDP relaxation approach is that we can a-posteriori check 
if the duality (or, in this case, the relaxation) gap is zero, from the optimal solution $\Xcpx^\star$. 
Indeed, if one solves~\eqref{eq:primal_sdp_rel}
and finds that the optimal $\Xcpx^\star$ has rank one, then we actually solved~\eqref{eq:primal}, hence 
the relaxation gap is zero.
Moreover, in this case, from spectral decomposition of $\Xcpx^\star$ we can 
get a vector $\xcpx^\star$ such that $\Xcpx^\star = (\xcpx^\star) (\xcpx^\star)^*$, and 
this vector is an optimal solution to the primal problem.
} 

In the following section we derive similar a-posteriori conditions for the dual problem~\eqref{eq:dual}.  
These conditions enable the computation of a primal optimal solution. Moreover, they allow discussing the 
uniqueness of such solution.
Furthermore, we prove that in special cases we can provide \emph{a-priori} conditions 
that guarantee that the duality gap is zero.


%

\subsection{\toCheck{Analysis of the dual problem}} 
\label{sec:analysis}

In this section we provide conditions under which the duality gap is zero. 
These conditions depend on the spectrum of $\Wcpx(\lam^\star)$, which 
arises from the solution of~\eqref{eq:dual}. We 
refer to $\Wcpx(\lam^\star)$ as the \emph{penalized pose graph matrix}.
A first proposition establishes that~\eqref{eq:dual}
attains an optimal solution.

\begin{proposition}
\label{prop:dualattained}
	The optimal value $d^\star$ in \eqref{eq:dual} is attained at a finite
	$\lam^\star$. Moreover, the penalized pose graph matrix $\Wcpx(\lam^\star)$ 
	has an eigenvalue in 0.
\end{proposition}

\noindent
{\bf Proof.}
Since $\Wcpx(\lam) \succeq 0$ implies that the diagonal entries are nonnegative,
the feasible set of~\eqref{eq:dual}  is contained in the set
$\{\lam:\; \Wcpx_{ii} - \lam_i \geq 0, \; i=1,\ldots,\dimWcpx\}$ 
(recall that $\Wcpx_{ii}$ are reals according to Remark~\ref{rmk:realDiagonalWcpx}).
On the other hand, $\lam_l = \zeros_{\dimWcpx}$ is feasible 
 and all points
in the set $\{\lam:\, \lam_i \geq 0$ yield an objective that is at least 
as good as
the objective value at $\lam_l$. Therefore, the problem is equivalent
to $\max_{\lam}\; \sum_i \lam_i$ subject to the original constraint, plus a box constraint on
$\lam\in\{0 \leq \lam_i\leq \Wcpx_{ii}, \; i=1,\ldots,n \}$. 
Thus we maximize a linear function over a compact set, hence a finite optimal 
solution $\lam^\star$  must be attained.

Now let us prove that $\Wcpx(\lam^\star)$ has an eigenvalue in zero.
Assume by contradiction that  $\Wcpx(\lam^\star)\succ 0$. From the Schur complement rule we know:
%
%
%
\beq
\Wcpx(\lam^\star) \succ 0
\Leftrightarrow
\left\{
\ba{l}
\Lcpx \succ 0 \\
\Qcpx(\lam^\star)  - \Scpx^* \Lcpx\inv \Scpx \succ 0
\ea
\right.
\eeq
\toCheck{The condition $\Lcpx \succ 0$ is always satisfied for a connected graph, since 
$\Lcpx = \Aanc\tran \Aanc$, and the anchored incidence matrix $\Aanc$, 
obtained by removing a node from the original incidence matrix, is always full-rank
for connected graphs~\cite[Section 19.3]{Schrijver98book}.}
Therefore, our assumption $\Wcpx(\lam^\star) \succ 0$ 
implies that
\beq
\Qcpx(\lam^\star)  - \Scpx^* \Lcpx\inv \Scpx = 
\Qcpx  - \Scpx^* \Lcpx\inv \Scpx  - \diag(\lam^\star) \succ 0
\eeq
Now, let
\[
\epsilon = \lam\ped{min}(\Qcpx(\lam^\star)  - \Scpx^* \Lcpx\inv \Scpx)  > 0.
\]
which is positive by the assumption $\Wcpx(\lam^\star) \succ 0$.
Consider $\lam = \lam^\star + \epsilon\one$, then
\[
\Qcpx(\lam)  - \Scpx^* \Lcpx\inv \Scpx 
 = 
 \Qcpx(\lam)  - \Scpx^* \Lcpx\inv \Scpx -\epsilon I \succeq 0,
\]
thus $\lam$ is dual feasible, and $\sum_i \lam_i > \sum_i \lam_i^\star$, which would
contradict optimality of $\lam^\star$. We thus proved that $\Qcpx(\lam^\star)$ must have a zero eigenvalue.
\qed

\vspace{.2cm}



\begin{proposition}[No duality gap]
\label{prop:nogapPoses}
If the zero eigenvalue of the penalized pose graph matrix $\Wcpx(\lam^\star)$ is {\em simple} then 
the duality gap is zero, i.e., $d^\star = f^\star$.
\end{proposition}

%




\noindent
{\bf Proof.} 
\toCheck{
We have already observed in Proposition~\ref{prop_dualrel} that~\eqref{eq:primal_sdp_rel} 
is the dual problem of~\eqref{eq:dual}, therefore, we can interpret $\Xcpx$ 
as a Lagrange multiplier for the constraint $\Wcpx(\lam) \succeq 0$. 
If we consider the optimal solutions $\Xcpx^\star$ and $\lam^\star$ of~\eqref{eq:primal_sdp_rel}
 and~\eqref{eq:dual}, respectively, the \emph{complementary slackness} condition 
ensures that $\tr (\Wcpx(\lam^\star) \Xcpx^\star)=0$ (see~\cite[Example 5.13]{Boyd04book}). 
Let us parametrize $\Xcpx^\star \succeq 0$ as
\[
\Xcpx^\star = \sum_{i=1}^{2n-1} \mu_i  \tilde v_i   \tilde v_i^\ast,
\]
where $0\leq \mu_1\leq \mu_2\leq \cdots\leq \mu_{2n-1}$ are the eigenvalues of $\Xcpx$, and
$ \tilde v_i$ form a unitary set of eigenvectors. Then, the complementary slackness condition becomes
\beas
\tr (\Wcpx(\lam^\star) \Xcpx^\star) &=&
\tr \left(\Wcpx(\lam^\star)  \sum_{i=1}^{2n-1} \mu_i  \tilde v_i   \tilde v_i^\ast \right)  \\
&=& 
\sum_{i=1}^{2n-1} \mu_i \tr \left(  \Wcpx(\lam^\star)   \tilde v_i   \tilde v_i^\ast \right) 	\\
&=& \sum_{i=1}^{2n-1} \mu_i \; \tilde v_i^\ast \; \Wcpx(\lam^\star)   \tilde v_i \;\;=\; 0.
\eeas
Since $\Wcpx(\lam^\star)\succeq 0$, the above quantity is zero at a nonzero $\Xcpx^\star$ 
($\Xcpx^\star$ cannot be zero since it needs to satisfy the constraints 
$\Xcpx_{ii}=1$) if and 
only if $\mu_i=0$ for $i=m+1,\ldots,2n-1$, and $\Wcpx(\lam^\star) \tilde v_i = 0$ for $i=1,\ldots,m$, 
where
$m$ is the multiplicity of 0 as an eigenvalue of $\Wcpx(\lam^\star)$.
Hence $\Xcpx^\star$ 
 has the form
\beq
\label{eq:XcpxInNullSpace}
\Xcpx^\star = \sum_{i=1}^m \mu_i \tilde v_i \tilde v_i^\ast,
\eeq
where $\tilde v_i$, $i=1,\ldots,m$, form a unitary basis of the null-space of  $\Wcpx(\lam^\star)$.
Now, if $m=1$, then the solution $\Xcpx^\star$
to problem~\eqref{eq:primal_sdp_rel} has rank one, but according to Proposition~\ref{prop_dualrel} this 
 implies $d^\star = f^\star$, proving the claim.
 }
\qed

\vspace{.2cm}
In the following we say that $\Wcpx(\lam^\star)$ satisfies the 
\emph{single zero eigenvalue property} (\SZEP) if its zero eigenvalue is simple. 
The following corollary provides a more explicit relation between the solution of the
primal and the dual problem when  $\Wcpx(\lam^\star)$ satisfies the \SZEP.

\begin{corollary}[\SZEP $\Rightarrow \xcpx^\star \in \calX(\lam^\star)$]
\label{cor:primalFeasilbeInCalX}
If the zero eigenvalue of $\Wcpx(\lam^\star)$ is {\em simple}, then 
the set $\calX(\lam^\star)$ contains a primal optimal solution. 
Moreover, the primal optimal solution is unique, up to an 
arbitrary rotation.
\end{corollary}

\noindent
{\bf Proof.} 
Let $\xcpx^\star$ be a primal optimal solution, and let $f^\star = (\xcpx^\star)^* \Wcpx (\xcpx^\star)$ be the corresponding optimal value. From Proposition~\ref{prop:nogapPoses} we know that the \SZEP implies that the 
duality gap is zero, i.e., $d^\star = f^\star$, hence
\beq
\label{eq:fstarEqdstar0}
\sum_{i=1}^n \lam_i^\star = (\xcpx^\star)^* \Wcpx (\xcpx^\star).
\eeq 
Since $\xcpx^\star$ is a solution of the primal, it must be feasible, hence $|\xcpx^\star_i|^2=1$, $i=n,\ldots,\dimWcpx$.
Therefore, the following equalities holds:
\beq
\label{eq:sumLamEq}
\sum_{i=1}^n \lam_i^\star = \sum_{i=1}^n \lam_i^\star |\xcpx^\star_{n+i-1}|^2 = (\xcpx^\star)^* \matTwo{\zeros &  \zeros\\ \zeros & \mbox{diag}(\lam^\star)} (\xcpx^\star)
\eeq 
Plugging~\eqref{eq:sumLamEq} back into~\eqref{eq:fstarEqdstar0}:
\beq
\label{eq:fstarEqdstar}
(\xcpx^\star)^* \left[ \Wcpx - \matTwo{\zeros &  \zeros\\ \zeros & \mbox{diag}(\lam^\star)} \right] (\xcpx^\star) = 0
\Leftrightarrow
(\xcpx^\star)^* \Wcpx(\lam^\star) (\xcpx^\star) = 0
\eeq 
which proves that $\xcpx^\star$ belongs to the null space of $\Wcpx(\lam^\star)$, which coincides with our 
definition of $\calX(\lam^\star)$ in~\eqref{eq:calX2}, proving the first claim.

Let us prove the second claim. 
From the first claim we know that the \SZEP implies that 
any primal optimal solution is in $\calX(\lam^\star)$. 
Moreover, when $\Wcpx(\lam^\star)$ has a single eigenvalue in zero, 
then $\calX(\lam^\star) = \rm{Kernel}(\Wcpx(\lam^\star))$ is 1-dimensional and 
 can be written as 
$\calX(\lam^\star) = \{ \tilde \gamma \, \xcpx^\star$ : $\tilde \gamma \in \Complex{}\}$, 
or, using the polar form for $\tilde \gamma$:
\beq
\label{eq:multipleXstar}
\calX(\lam^\star) = \{ \eta \e^{\j \varphi} \; \xcpx^\star : \eta,\varphi \in \Real{}\}
\eeq
%
From~\eqref{eq:multipleXstar} it's easy to see that any $\eta \neq 1$ 
would alter the norm of $\xcpx^\star$, leading to a solution that it's not primal feasible. 
On the other hand, any $\e^{\j \varphi} \xcpx^\star$ belongs to $\calX(\lam^\star)$, and 
it's primal feasible ($|\e^{\j \varphi} \xcpx_i^\star|=|\xcpx_i^\star|$), hence 
by Theorem~\ref{thm:Everett}, any $\e^{\j \varphi} \xcpx^\star$ is primal optimal. 
We conclude the proof by noting that the multiplication by $\e^{\j \varphi}$ corresponds to a 
global rotation of the pose estimate $\xcpx^\star$: this can be easily understood from 
the relation~\eqref{eq:complexA}.
\qed

\vspace{.2cm}
Proposition~\ref{prop:nogapPoses} provides an \emph{a-posteriori} condition on the duality gap, that 
requires solving the dual problem; while Section~\ref{sec:experiments} will show that this condition is 
very useful in practice, it is also interesting to devise a-priori conditions, that can 
be assessed from the pose graph matrix $\Wcpx$, without solving the dual problem.
A first step in this direction is the following proposition.

\begin{proposition}[Strong duality in trees and balanced pose graphs]
\label{prop:nogapTreesBalanced}
Strong duality holds for any balanced pose graph optimization problem, 
and for any pose graph whose underlying graph is a tree. 
\end{proposition}

{\bf Proof.}
Balanced pose graphs and trees have in common the fact that 
they attain $f^\star = 0$ (Propositions~\ref{prop:zeroCostTrees}-\ref{prop:zeroCostBalancedGraphs}).
By weak duality we know that $d^\star \leq 0$. However, 
$\lam = 0_n$ is feasible (as $\Wcpx \succeq 0$) and 
attains $d(\lam) = 0$, hence $\lam = 0_n$ is feasible and dual optimal, proving $d^\star = f^\star$. 
\qed


\section{\toCheck{Algorithms}}
\label{sec:algorithms}

\toCheck{
In this section we exploit the results presented so far to devise an algorithm 
to solve \PGO. The idea is to solve the dual problem, 
and use $\lam^\star$ and $\Wcpx(\lam^\star)$ to compute a solution for the primal \PGO problem.
We split the presentation into two sections: Section~\ref{sec:algoSZEP}
discusses the case in which $\Wcpx(\lam^\star)$ satisfies the \SZEP, while 
Section~\ref{sec:algoNOSZEP} discusses the case in which $\Wcpx(\lam^\star)$ has 
multiple eigenvalues in zero. This distinction is important as in the 
former case (which is the most common in practice) we can compute a provably optimal solution for \PGO, while in the 
latter case our algorithm returns an estimate that is not necessarily optimal.
Finally, in Section~\ref{sec:algo} we summarize our algorithm 
and present the corresponding pseudocode.
}

\subsection{Case 1: $\Wcpx(\lam^\star)$ satisfies the \SZEP}
\label{sec:algoSZEP}

\newcommand{\vcpx}{\tilde v}

\toCheck{
According to Corollary~\ref{cor:primalFeasilbeInCalX},
if $\Wcpx(\lam^\star)$ has a single zero eigenvalue, then 
the optimal solution of the primal problem $\xcpx^\star$ is in $\calX(\lam^\star)$, 
where $\calX(\lam^\star)$ coincides with the null space of $\Wcpx(\lam^\star)$, 
as per~\eqref{eq:calX2}.
Moreover, this null space is 1-dimensional, hence
it can be written explicitly as:
 \beq
\label{eq:calXSZEP}
\calX(\lam^\star) = \rm{Kernel}(\Wcpx(\lam^\star))= \{\tilde v \in \Complex{\dimWcpx}:\;  \tilde v = \gamma \xcpx^\star \}, 
\eeq
which means that any vector in the null space is a scalar multiple of the primal optimal solution $\xcpx^\star$.
This observation suggests a computational approach to compute $\xcpx^\star$. 
We can first compute 
an eigenvector $\tilde v$ corresponding to the single zero eigenvalue of $\Wcpx(\lam^\star)$ 
(this is a vector in the null 
space of $\Wcpx(\lam^\star)$). Then, 
since $\xcpx^\star$ must be primal feasible (i.e., $|\xcpx_{n}| = \ldots = |\xcpx_{\dimWcpx}| = 1$), 
 we compute 
a suitable scalar $\gamma$ that makes $\frac{1}{\gamma} \tilde v$ primal feasible. 
This scalar is clearly $\gamma = |\vcpx_{n}| = \ldots = |\vcpx_{\dimWcpx}|$ 
(we essentially need to normalize the norm of the last $n$ entries of $\vcpx$). 
The existence of a suitable $\gamma$, 
 and hence the fact that $|\vcpx_{n}| = \ldots = |\vcpx_{\dimWcpx}| >0$,
 is guaranteed by Corollary~\ref{cor:primalFeasilbeInCalX}. 
 As a result we get the optimal solution $\xcpx^\star = \frac{1}{\gamma} \tilde v$.
 The pseudocode of our approach is given in Algorithm~\ref{alg:primalViaDual}, 
 and further discussed in Section~\ref{sec:algo}.
}

 \subsection{Case 2: $\Wcpx(\lam^\star)$ does \emph{not} satisfy the \SZEP}
 \label{sec:algoNOSZEP}


\toCheck{
Currently we are not able to compute a guaranteed optimal solution for \PGO, when 
$\Wcpx(\lam^\star)$ has multiple eigenvalues in zero.
 Nevertheless,  
it is interesting to exploit the solution of the dual problem for finding a (possibly suboptimal) estimate, 
which can be used, for instance, as initial guess for an iterative technique. 
}

\newcommand{\acpx}{\tilde z}
\newcommand{\Vcpx}{\tilde V}
\newcommand{\Zcpx}{\tilde Z}

\toCheck{
{\bf Eigenvector method.} 
One idea to compute a suboptimal solution from the dual problem
is to follow the same approach of Section~\ref{sec:algoSZEP}: we compute an eigenvector of 
$\Wcpx(\lam^\star)$, corresponding to one of the zero eigenvalues, and we normalize it 
to make it feasible. In this case, we are not guaranteed that $|\vcpx_{n}| = \ldots = |\vcpx_{\dimWcpx}| >0$ 
(as in the previous section), hence the normalization has to be done component-wise, 
for each of the last $\nrNodes$ entries of $\vcpx$.
In the following, we consider an alternative approach, 
which we have seen to perform better in practice (see experiments in Section~\ref{sec:experiments}).
}

\toCheck{
{\bf Null space method.} 
This approach is based on the insight of
 Theorem~\ref{thm:Everett}: if there is a primal feasible $\xcpx \in \calX(\lam^\star)$, then 
 $\xcpx$ must be primal optimal. Therefore we look for a vector $\xcpx \in \calX(\lam^\star)$ 
 that is ``close'' to the feasible set.
According to~\eqref{eq:calX2}, $\calX(\lam^\star)$ coincides with the 
  null space of $\Wcpx(\lam^\star)$. Let us denote with
  $\Vcpx \in \Complex{(2n-1) \times q}$ a basis of the null space of $\Wcpx(\lam^\star)$, 
  where $q$ is the number of zero eigenvalues of $\Wcpx(\lam^\star)$.\footnote{$\Vcpx$ can be 
  computed from singular value decomposition of $\Wcpx(\lam^\star)$.}
Any vector $\xcpx$ in the null space 
  of $\Wcpx(\lam^\star)$ can be written as $\xcpx = \Vcpx \acpx$, for some vector $\acpx \in \Complex{q}$.
  Therefore we propose to compute a possibly suboptimal estimate $\xcpx = \Vcpx \acpx^\star$, 
  where $\acpx^\star$ solves the following 
  optimization problem:
\bea
\label{eq:feasibleInNullSpace}
 \max_{\acpx}  &  \displaystyle\sum_{i=1}^{2n-1} \mbox{real}(\Vcpx_i \acpx) + \mbox{imag}(\Vcpx_i \acpx) \\
 \subt & |\Vcpx_i \acpx|^2 \leq 1, & i=n,\ldots,2n-1 \nonumber 
\eea
where $\Vcpx_i$ denotes the $i$-th row of $\Vcpx$, 
and $\mbox{real}(\cdot)$ and $\mbox{imag}(\cdot)$ return the real and the imaginary part 
of a complex number, respectively. 
For an intuitive explanation of problem~\eqref{eq:feasibleInNullSpace}, we 
notice that the feasible set of the primal problem~\eqref{eq:primal} is 
described by $|\xcpx_i|^2 = 1$, for $i=n,\!\ldots\!,2n\!-\!1$. In problem~\eqref{eq:feasibleInNullSpace} we 
relax the equality constraints to convex inequality constraints 
$|\xcpx_i|^2 \leq 1$, for $i=n,\ldots,2n-1$; these can be written as $|\Vcpx_i \acpx|^2 \leq 1$, 
recalling that we are searching in the null space of $\Wcpx(\lam^\star)$, which is 
spanned by $\Vcpx \acpx$. Then, the objective function in~\eqref{eq:feasibleInNullSpace}
encourages ``large'' elements $\Vcpx_i \acpx$, hence pushing the inequality $|\Vcpx_i \acpx|^2 \leq 1$ 
to be tight. While other metrics 
can force large entries $\Vcpx_i \acpx$, we preferred the linear metric~\eqref{eq:feasibleInNullSpace} 
to preserve convexity. 
}

\toCheck{
Note that $\xcpx = \Vcpx \zcpx^\star$, in general, is neither optimal nor feasible for 
our \PGO problem~\eqref{eq:primal}, hence 
we need to normalize it to get a feasible estimate.
The experimental section provides empirical evidence that, despite being heuristic in 
nature, this method performs well in practice, outperforming --among the others-- 
the eigenvector method presented earlier in this section.
}


 \subsection{Pseudocode and implementation details}
 \label{sec:algo}

\toCheck{
The pseudocode of our algorithm is given in Algorithm~\ref{alg:primalViaDual}.
The first step is to solve the dual problem, and 
check the a-posteriori condition of Proposition~\ref{prop:nogapPoses}.
If the \SZEP is satisfied, then we can compute the optimal solution by 
scaling the eigenvector of $\Wcpx(\lam^\star)$ corresponding to the 
zero eigenvalue $\mu_1$. This is the case described in Section~\ref{sec:algoSZEP}
 and is the most relevant in practice, since the vast majority of robotics 
 problems falls in this case.
}

\toCheck{
The ``else'' condition corresponds to the case in which $\Wcpx(\lam^\star)$ 
has multiple eigenvalue in zero. 
The  pseudocode implements 
the null space approach of Section~\ref{sec:algoNOSZEP}.
The algorithm computes a basis for the null space of $\Wcpx(\lam^\star)$ 
and solves~\eqref{eq:feasibleInNullSpace} to find a
vector belonging to the null space (i.e, in the form $\xcpx = \Vcpx \acpx$) 
that is close to the feasible set. Since such vector is not guaranteed to be 
primal feasible (and it is not in general), the algorithm normalizes 
the last $n$ entries of
$\xcpx^\star = \Vcpx \acpx^\star$, so to satisfy the unit norm constraints in~\eqref{eq:primal}.
Besides returning the estimate $\xcpx^\star$, the algorithm also provides an optimality 
certificate when  $\Wcpx(\lam^\star)$ has a single eigenvalue in zero.
}

\newcommand{\isOpt}{\tt{isOpt}}

\vspace{0.3cm}
\begin{algorithm}[H]
\SetKwInOut{Input}{input}\SetKwInOut{Output}{output}

 \Input{Complex \PGO matrix $\Wcpx$}
 \Output{Primal solution $\xcpx^\star$ and optimality certificate $\isOpt$}
 \BlankLine
 solve the dual problem~\eqref{eq:dual} and get $\lam^\star$ \;
 \eIf{$\Wcpx(\lam^\star)$ has a single eigenvalue $\mu_1$ at zero}
  {
  compute the eigenvector $\tilde v$ of $\Wcpx(\lam^\star)$ corresponding to $\mu_1$\;
  compute $\xcpx^\star = \frac{1}{\gamma} \tilde v$, where $\gamma = |{\tilde v_j}|$, for any $j \in \{n,\ldots,\dimWcpx\}$ \;
  set $\isOpt = \tt{true}$\;
  }{
  compute a basis $\Vcpx$ for the null space of $\Wcpx(\lam^\star)$ using SVD\;
  compute $\acpx^\star$ by solving the convex problem~\eqref{eq:feasibleInNullSpace}\;
  set $\xcpx^\star = \Vcpx \acpx^\star$ and normalize $|{\xcpx_i}|$ to 1, for all $i = n,\ldots,\dimWcpx$\;
  set $\isOpt = \tt{unknown}$\; 
  }
  return $(\xcpx^\star,\isOpt)$
 \caption{Solving \PGO using Lagrangian duality.\label{alg:primalViaDual}}
\end{algorithm}
\vspace{0.3cm}


\newcommand{\NS}{{\tt NS}\xspace} 
\newcommand{\GN}{{\tt GN}\xspace} 
\newcommand{\Eig}{{\tt Eig}\xspace} 
\newcommand{\SDP}{{\tt SDP}\xspace} 
\newcommand{\EigR}{{\tt EigR}\xspace} 

\section{Numerical Analysis and Discussion}
\label{sec:experiments}

\newcommand{\plc}{P_c}

\toCheck{
The objective of this section is four-fold.
First, we validate our theoretical derivation, 
providing experimental evidence that supports the claims.
Second, we show that the duality gap is zero in a vast amount or practical problems.
Third, we confirm the effectiveness of  
 Algorithm~\ref{alg:primalViaDual} to solve \PGO.
Fourth, we provide toy examples in which the duality gap is greater than zero, 
hoping that this can stimulate further investigation towards \emph{a-priori} 
conditions that ensure zero duality gap. }

\paragraph{Simulation setup.} For each run we generate a 
random graph with $n=10$ nodes, unless specified otherwise. 
We draw the position of each pose by a uniform distribution in 
a $10\rm{m} \times 10\rm{m}$ square. Similarly,  
ground truth node orientations are randomly selected in $\angleDomain$.
Then we create set of edges defining   
a spanning \emph{path} of the graph (these are usually called \emph{odometric edges}); 
moreover, we add further edges to the edge set, by 
connecting random pairs of nodes with probability $\plc=0.1$
 (these are usually called \emph{loop closures}).
 From the randomly selected \emph{true} poses, and for each edge $(i,j)$ 
 in the edge set, 
 we generate the relative pose measurement using the following model: 
\beq
\label{eq:measurementNoise}
\ba{llll}
\Deltaij &=& R_i\tran \left( p_j - p_i \right) + \epsilon_{\Delta}, & \qquad \epsilon_{\Delta} \sim N(\zeros_2,\sigma_{\Delta}^2) \\
\Rij &=& R_i\tran \; R_j \; R(\epsilon_R),                                & \qquad \epsilon_{R} \sim N(\zeros,\sigma_R^2) 
\ea
\eeq
where $\epsilon_{\Delta} \in \Real{2}$ and $\epsilon_{R} \in \Real{}$ are zero-mean Normally distributed random variables, 
with standard deviation $\sigma_{\Delta}$ and $\sigma_R$, respectively, and $R(\epsilon_R)$ is a random planar rotation of an angle $\epsilon_R$.
Unless specified otherwise, all statistics are computed over 100 runs.

\paragraph{Spectrum of $\Wcpx$.} In Proposition~\ref{prop:szepWtilde}, we showed that the complex anchored
 pose graph matrix $\Wcpx$ has at most 
one eigenvalue in zero, and the zero eigenvalue only appears when the pose graph is balanced or is a tree. 
 
\begin{figure}[t]
\begin{minipage}{\textwidth}
\begin{tabular}{cc}
\begin{minipage}{6cm}%
\centering
\hspace{-9mm} \includegraphics[scale=0.4]{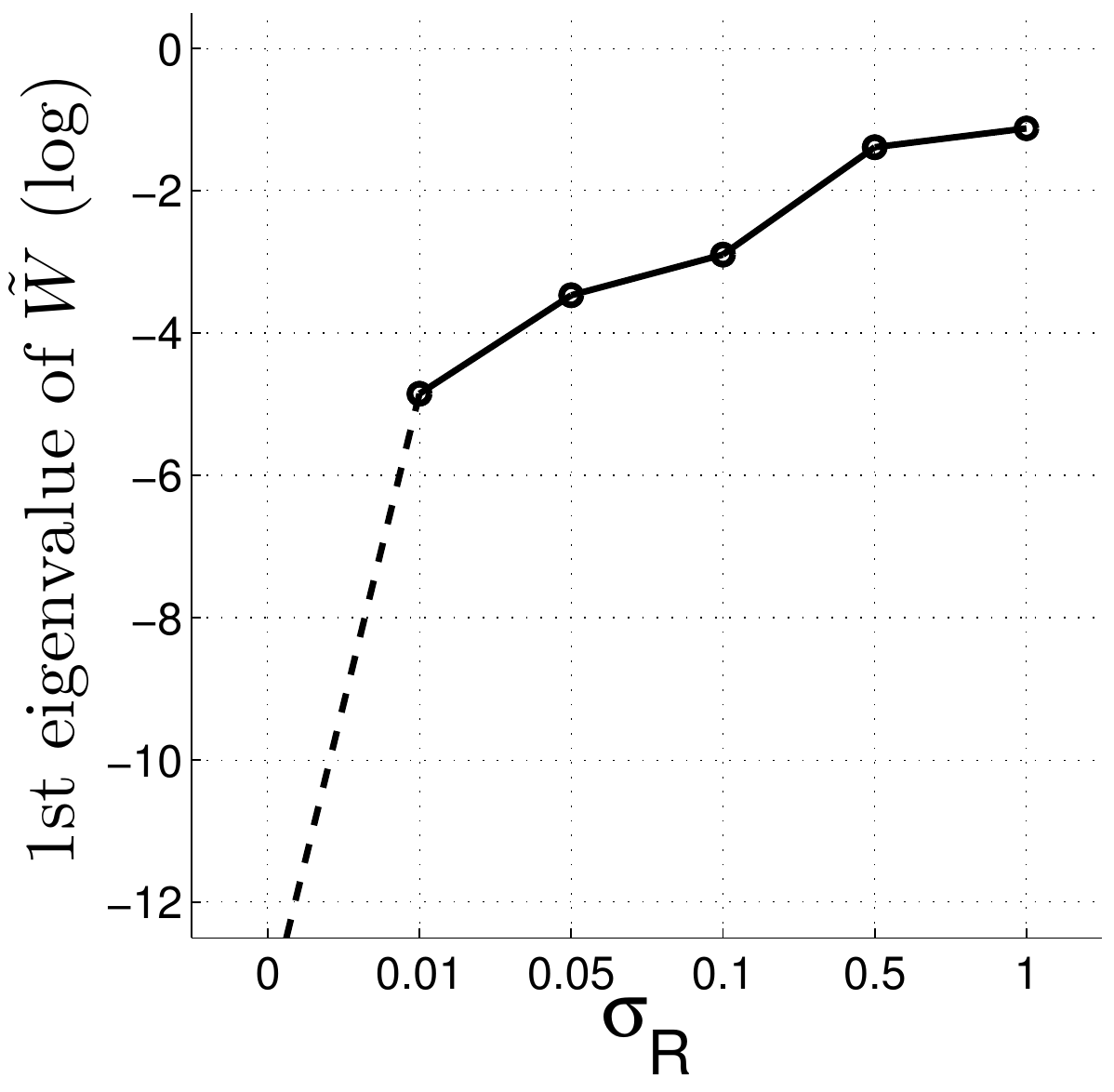} \\
(a)
\end{minipage}%
&
\begin{minipage}{6cm}%
\centering
\hspace{-9mm} \includegraphics[scale=0.4]{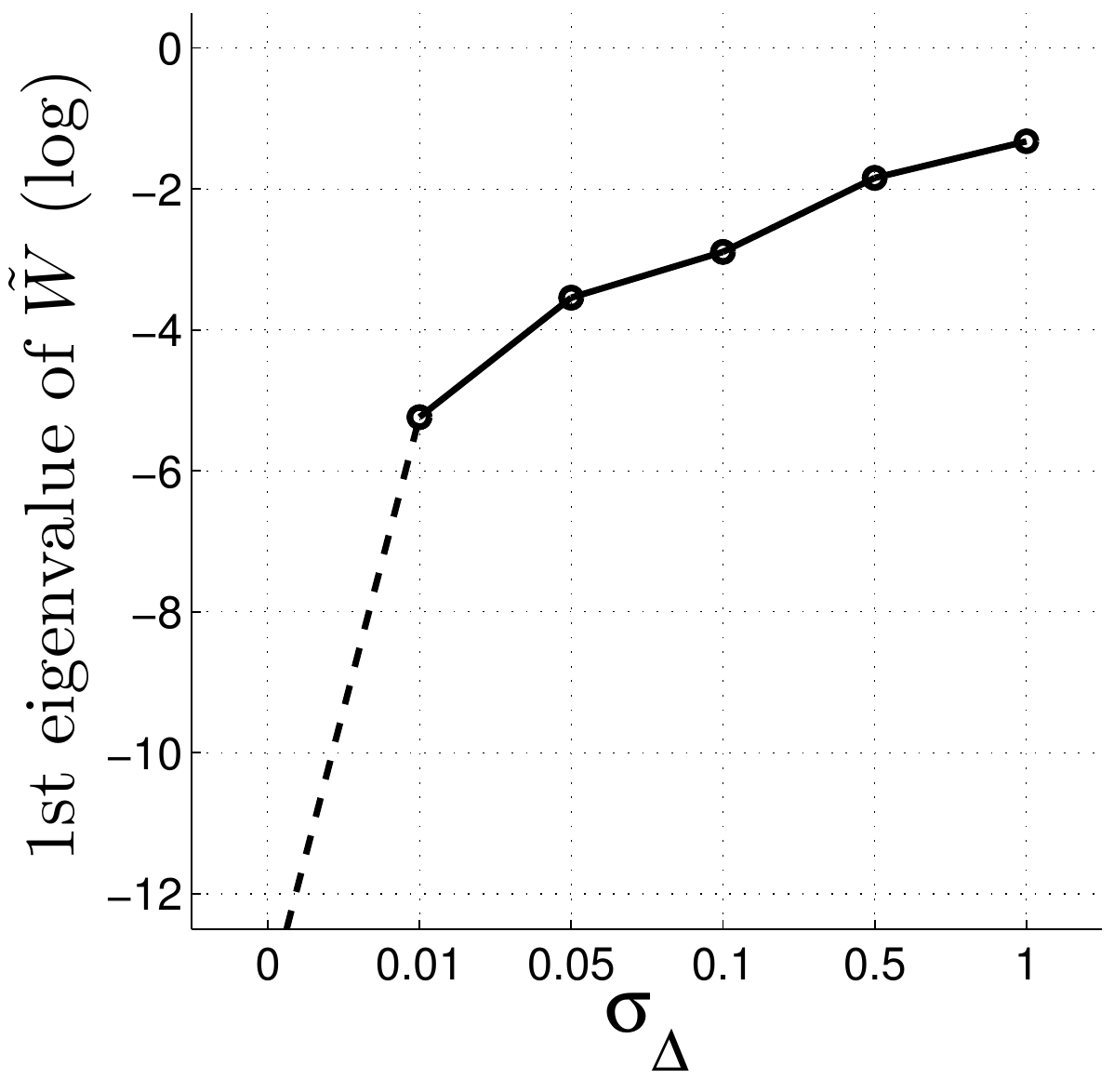} \\
(b)
\end{minipage}%
\end{tabular}%
\end{minipage}%
\caption{Smallest eigenvalue of $\Wcpx$ (in logarithmic scale) for different levels of 
(a) rotation noise (std: $\sigma_R$), and (b) translation noise (std: $\sigma_\Delta$).
The figure show the minimum observed value over 100 Monte Carlo runs, for non-tree graphs. 
The minimum eigenvalue is zero only if the graph is balanced. \label{fig:eig1Wcpx}}
\end{figure}

Fig.~\ref{fig:eig1Wcpx}(a) reports the value of the 
smallest eigenvalue of $\Wcpx$ (in log scale) for different $\sigma_R$, with fixed $\sigma_{\Delta} = 0$m.
When also $\sigma_R$ is zero, the pose graph is balanced, hence the smallest eigenvalue of $\Wcpx$ is 
(numerically) zero. For increasing levels of noise, the smallest eigenvalue increases and stays away from zero.
Similarly, Fig.~\ref{fig:eig1Wcpx}(b) reports the value of the 
smallest observed eigenvalue of $\Wcpx$ (in log scale) for different $\sigma_{\Delta}$, with fixed $\sigma_{R} = 0$rad.

\paragraph{Duality gap is zero in many cases.} 
This section shows that for the levels of measurement noise of practical interest, 
the matrix $\Wcpx(\lam^\star)$  satisfies the Single Eigenvalue Property (\SZEP), 
hence the duality gap is zero (Proposition~\ref{prop:nogapPoses}). 
We consider the same measurement model of Eq.~\eqref{eq:measurementNoise}, 
and we analyze the percentage of tests in which $\Wcpx(\lam^\star)$ satisfies the \SZEP. 

\begin{figure}[t]
\begin{minipage}{\textwidth}
\begin{tabular}{cc}
\begin{minipage}{6cm}%
\centering
\hspace{-4mm} \includegraphics[scale=0.28]{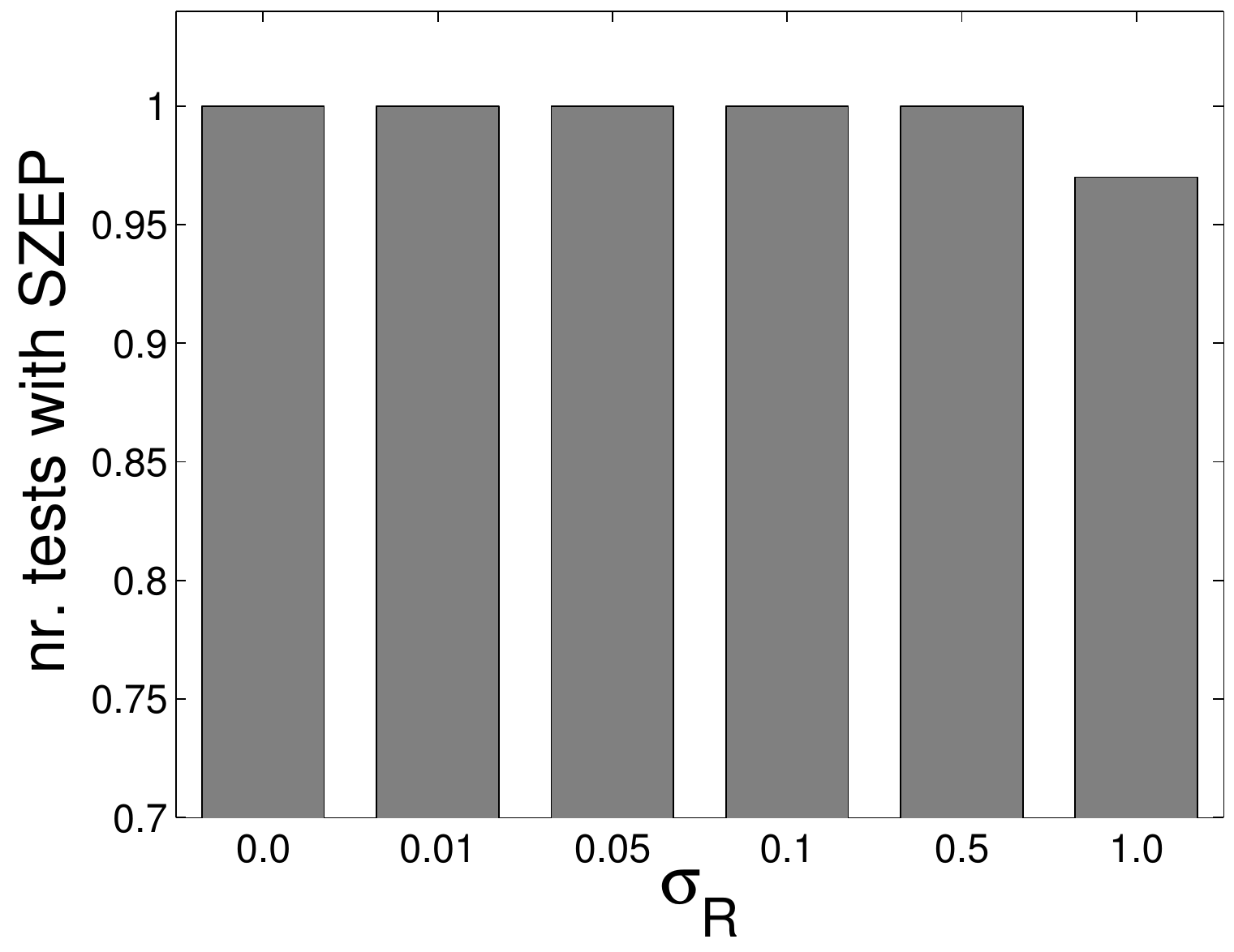} 
\vspace{-0.2cm} \\ (a) {\footnotesize $\sigma_\Delta = 0.1$, $\plc = 0.1$, $n=10$}
\end{minipage}%
&
\begin{minipage}{6cm}%
\centering
\hspace{-4mm} \includegraphics[scale=0.28]{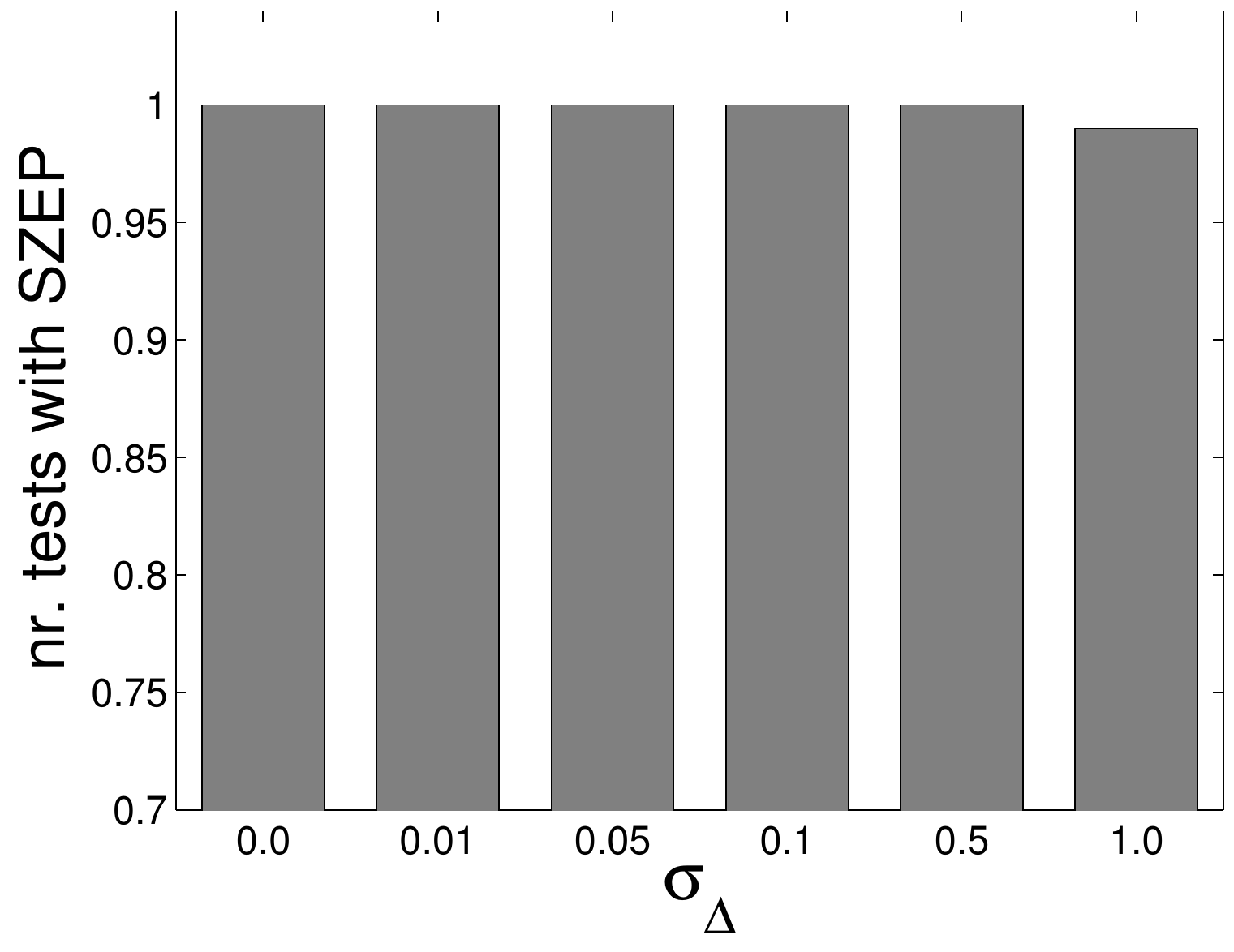}  
\vspace{-0.2cm} \\ (b) {\footnotesize $\sigma_R = 0.1$, $\plc = 0.1$, $n=10$}
\end{minipage}%
\\
\\
\begin{minipage}{6cm}%
\centering
\hspace{-4mm} \includegraphics[scale=0.28]{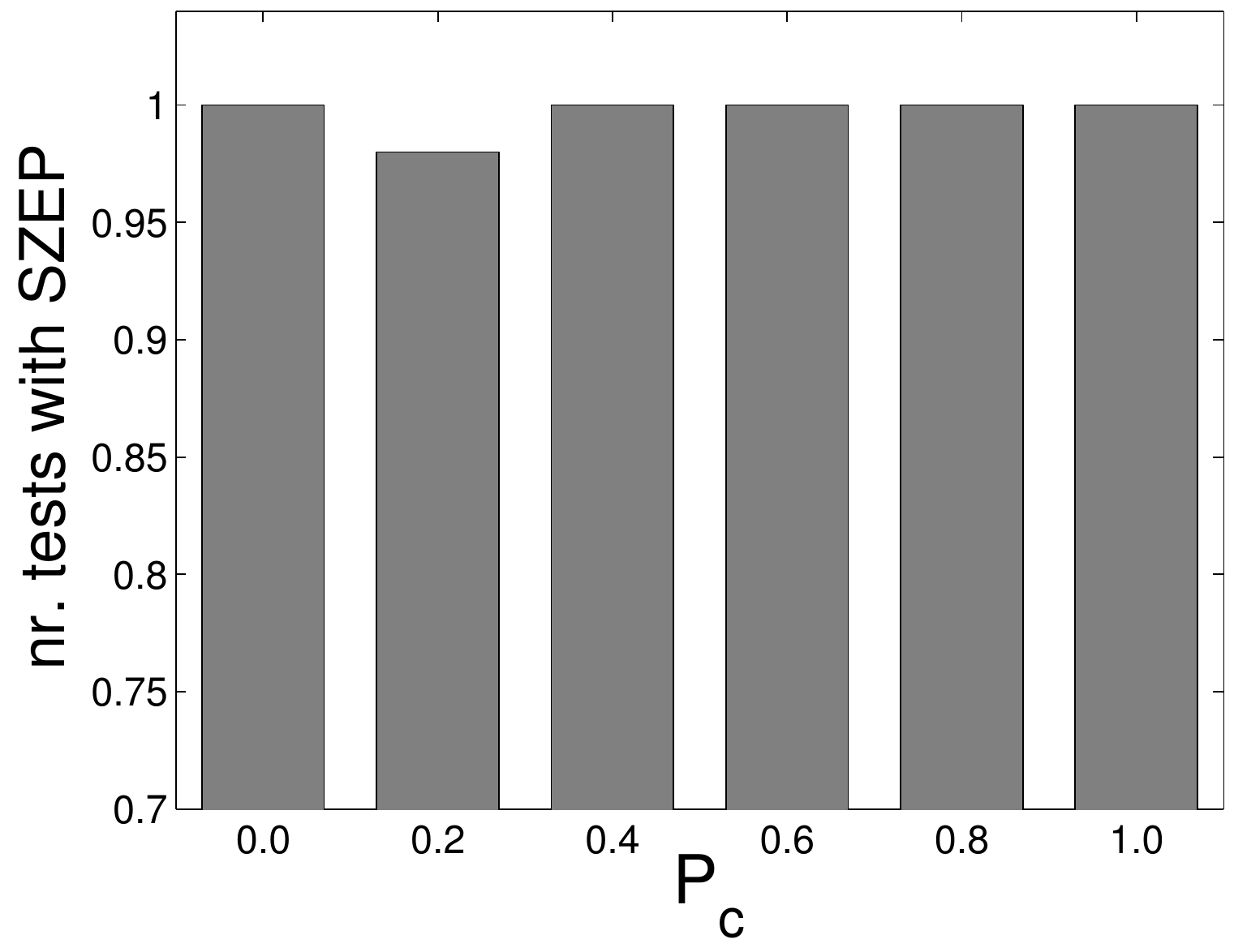}  
\vspace{-0.2cm} \\ (c) {\footnotesize $\sigma_\Delta = \sigma_R = 1$, $n=10$}
\end{minipage}%
&
\begin{minipage}{6cm}%
\centering
\hspace{-4mm} \includegraphics[scale=0.28]{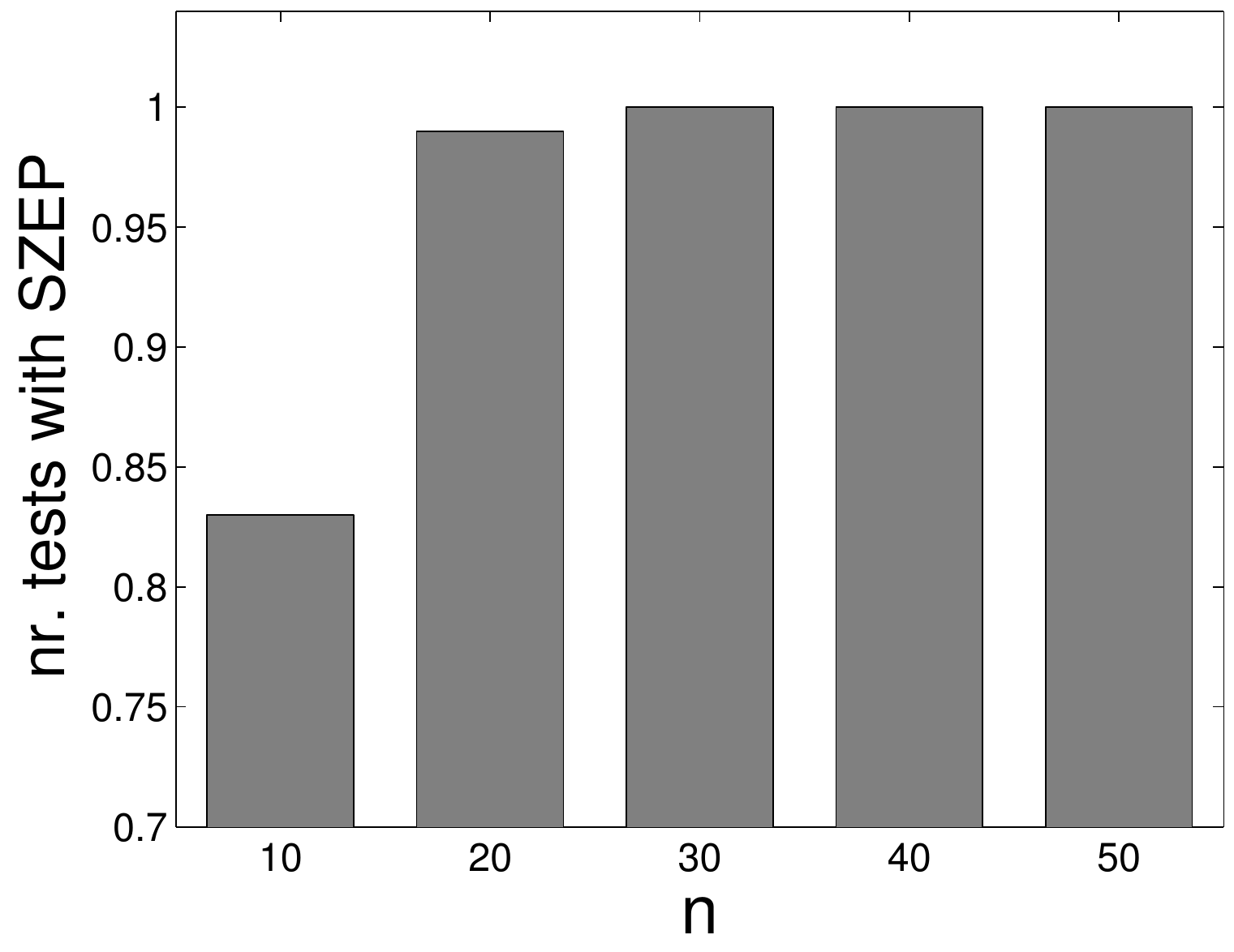}  
\vspace{-0.2cm} \\ (d) {\footnotesize $\sigma_\Delta = \sigma_R = 1$, $\plc = 0.1$}
\end{minipage}%
\end{tabular}%
\end{minipage}%
\caption{Percentage of problems in which $\Wcpx(\lam^\star)$ satisfied the \SZEP property, 
for different (a) rotation measurement noise $\sigma_R$, 
(b) translation measurement noises $\sigma_\Delta$,  
(c) loop closure probability $\plc$, 
 (d) number of nodes $\nrNodes$. \label{fig:percentageSZEP}}
\end{figure}

Fig.~\ref{fig:percentageSZEP}(a) shows the percentage of the experiments in which the penalized 
pose graph matrix $\Wcpx(\lam^\star)$ has a single zero eigenvalue, for different 
values of rotation noise $\sigma_R$, and keeping fixed the translation noise to $\sigma_\Delta = 0.1$m (this is a typical value 
in mobile robotics applications). For $\sigma_R \leq 0.5$rad, $\Wcpx(\lam^\star)$ satisfies the \SZEP in all tests. 
This means that in this range of operation, Algorithm~\ref{alg:primalViaDual} is guaranteed to compute a globally-optimal solution for \PGO. For $\sigma_R = 1$rad, the percentage of successful experiments drops, while still remaining larger than 
$90\%$. Note that $\sigma_R = 1$rad is a very large rotation noise (in robotics, typically $\sigma_R \leq 0.3$rad~\cite{Carlone14tro}), 
and it is not far from the case in which rotation measurements are uninformative (uniformly distributed in $\angleDomain$). 
To push our evaluation further we also tested this extreme case. When rotation noise is uniformly distributed in $\angleDomain$, 
we obtained a percentage of successful tests (single zero eigenvalue) of $69\%$, which confirms that 
 the number of cases in which we can compute a globally optimal solution drops gracefully 
 when increasing the noise levels.

%

Fig.~\ref{fig:percentageSZEP}(b) shows the percentage of the experiments in which $\Wcpx(\lam^\star)$ has a single zero eigenvalue, 
for different values of translation noise $\sigma_\Delta$, and keeping fixed the rotation noise to $\sigma_R = 0.1$rad. 
Also in this case, for practical noise regimes, our approach can compute a global solution in all cases. 
The percentage of successful tests drops to $98\%$ when the translation noise has standard deviation $1\rm{m}$.
We also tested the case of uniform noise on translation measurements. When we draw the measurement noise 
from a uniform distribution in $[-5,5]^2$ (recall that the poses are deployed in a $10 \times 10$ square), 
the percentage of successful experiments is $68\%$.

We also tested the percentage of experiments satisfying the \SZEP for different 
levels of connectivity of the graph, controlled by the parameter $\plc$. We observed $100\%$ successful experiments, independently 
on the choice of $\plc$, for 
$\sigma_R = \sigma_\Delta = 0.1$ and $\sigma_R = \sigma_\Delta = 0.5$. A more interesting case if shown in 
Fig.~\ref{fig:percentageSZEP}(c) and corresponds to the 
case $\sigma_R = \sigma_\Delta = 1$. The \SZEP is always satisfied for $\plc=0$: this is natural as 
$\plc=0$ always produces trees, for which we are guaranteed to satisfy the \SZEP (Proposition~\ref{prop:nogapTreesBalanced}). 
For $\plc = 0.2$ the \SZEP fails in few runs. Finally, increasing the connectivity beyond $\plc = 0.4$ re-establishes 
$100\%$ of successful tests. 
This would suggest that the connectivity level of the graph 
influences the duality gap, and better connected graphs have 
 more changes to have zero duality gap.
 
 Finally, we tested the percentage of experiments satisfying the \SZEP for different 
 number of nodes $n$. We tested the following number of nodes: $n=\{10,20,30,40,50\}$. 
 For $\sigma_R = \sigma_\Delta = 0.1$ and $\sigma_R = \sigma_\Delta = 0.5$ the \SZEP
 was satisfied in $100\%$ of the tests, and we omit the results for brevity.
 The more challenging case $\sigma_R = \sigma_\Delta = 1$ is shown in Fig.~\ref{fig:percentageSZEP}(d). 
The percentage of successful tests increases for larger number of poses. 
We remark that current SDP solvers do not scale well to large problems, hence 
 a Monte Carlo analysis over larger problems becomes prohibitive. 
We refer the reader to~\cite{Carlone15icra-verify} 
for single-run experiments on larger \PGO problems, which confirm that 
the duality gap is zero in problems arising in real-world robotics applications.

\begin{figure}[t]
\centering
\includegraphics[scale=0.33]{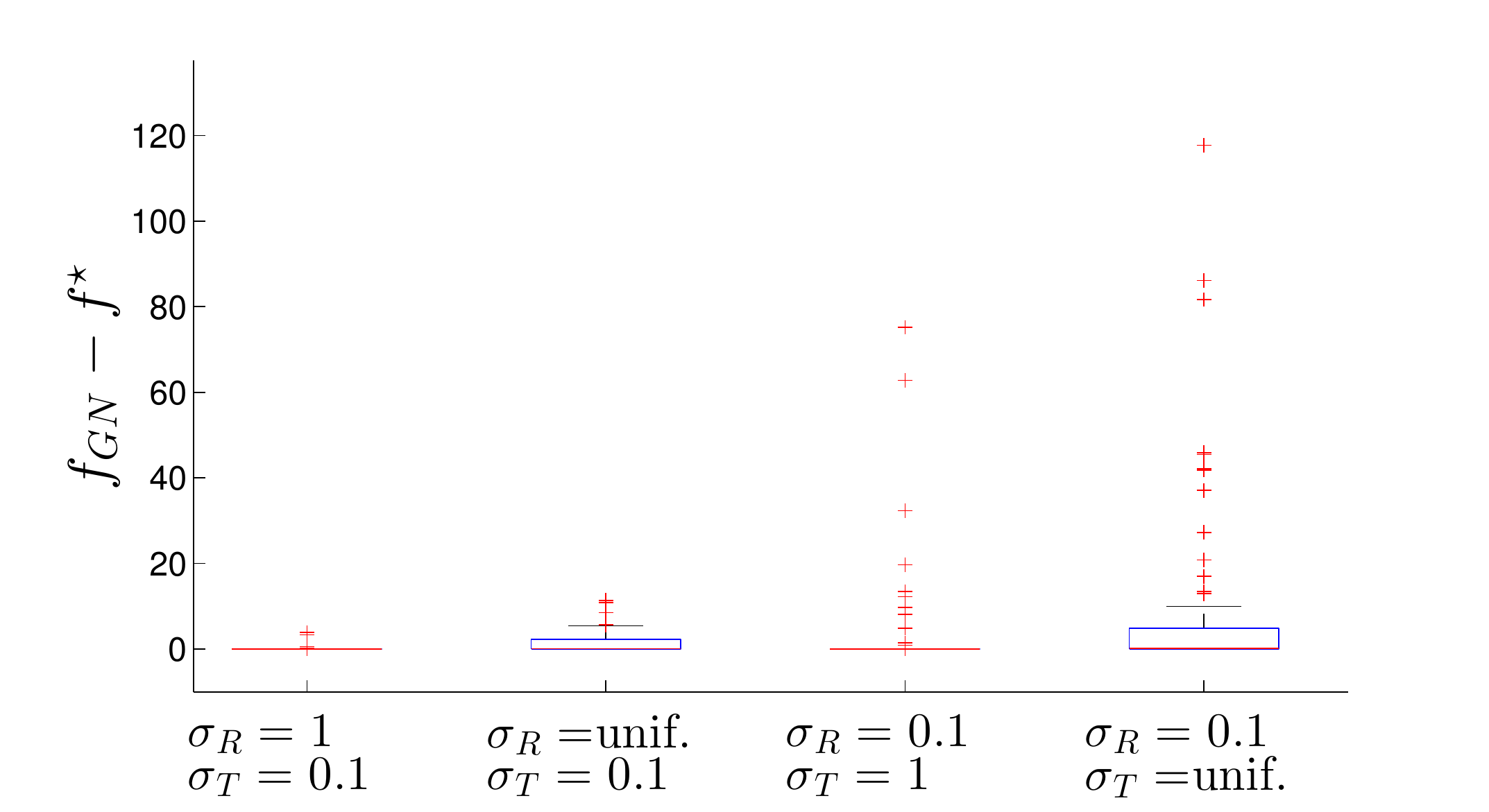} 
\caption{\label{fig:preformancePStarViaDual}
Statistics on tests in which the \SZEP is satisfied: 
the figure reports the gap between the optimal objective $f^\star$ attained by Algorithm~\ref{alg:primalViaDual} 
and the objective $f_{GN}$ attained by a Gauss-Newton method initialized at the true poses. 
We simulate different combinations of noise (see x-axis labels), keeping 
fixed $n=10$ and $\plc = 0.1$. The label ``unif.'' denotes uniform noise for rotations 
(in $\angleDomain$) or translations (in $[-5,+5]$).
}
\end{figure}
 
\toCheck{
\paragraph{Performance of Algorithm~\ref{alg:primalViaDual}}
In this section we show that Algorithm~\ref{alg:primalViaDual} provides an effective solution
for \PGO. When $\Wcpx(\lam^\star)$ satisfies the \SZEP, the algorithm is provably optimal, 
and it enables to solve problems that are already challenging for iterative solvers.
When the $\Wcpx(\lam^\star)$ does \emph{not} satisfy the \SZEP, we show that the proposed 
approach, while not providing performance guarantees, largely outperforms competitors.
}

\toCheck{
\emph{Case 1: $\Wcpx(\lam^\star)$ satisfies the \SZEP.}
When $\Wcpx(\lam^\star)$ satisfies the \SZEP,
 Algorithm~\ref{alg:primalViaDual} is guaranteed to produce a globally optimal solution. 
 However, one may argue that in the regime of operation in which 
the \SZEP holds, \PGO problem instances are sufficiently ``easy'' that commonly used iterative techniques also perform well.
In this paragraph we briefly show that the \SZEP is satisfied in many instances that are hard to solve.  
For this purpose, we focus on the most challenging cases we discussed so far, i.e., problem instances with 
large rotation and translation noise. Then we consider the problems in which  the \SZEP is satisfied and we compare the 
solution of Algorithm~\ref{alg:primalViaDual}, which is proven to attain $f^\star$, versus the solution of a 
Gauss-Newton method initialized at the \emph{true} poses. Ground truth poses are an ideal initial guess
 (which is unfortunately available only in simulation): intuitively, the global minimum of the cost 
 should be close to the ground truth poses (this is one of the motivations for maximum likelihood estimation). 
Fig.~\ref{fig:preformancePStarViaDual} shows the gap between the objective attained by the Gauss-Newton method 
(denoted as $f_{GN}$) and the optimal objective obtained from Algorithm~\ref{alg:primalViaDual}. 
 The figure confirms that our algorithm provides a guaranteed optimal solution in a regime that is 
 already challenging, and in which iterative approaches may fail to converge even from a good initialization.
 }
 

\newcommand{\sizeCol}{2.9cm}
\newcommand{\scaleFig}{0.16}
\newcommand{\mySmallFont}{\tiny}

\begin{figure}[t]
\begin{minipage}{\textwidth}
\hspace{-8mm}
\begin{tabular}{cccc}
\begin{minipage}{\sizeCol}%
\centering
\hspace{-4mm} \includegraphics[scale=\scaleFig]{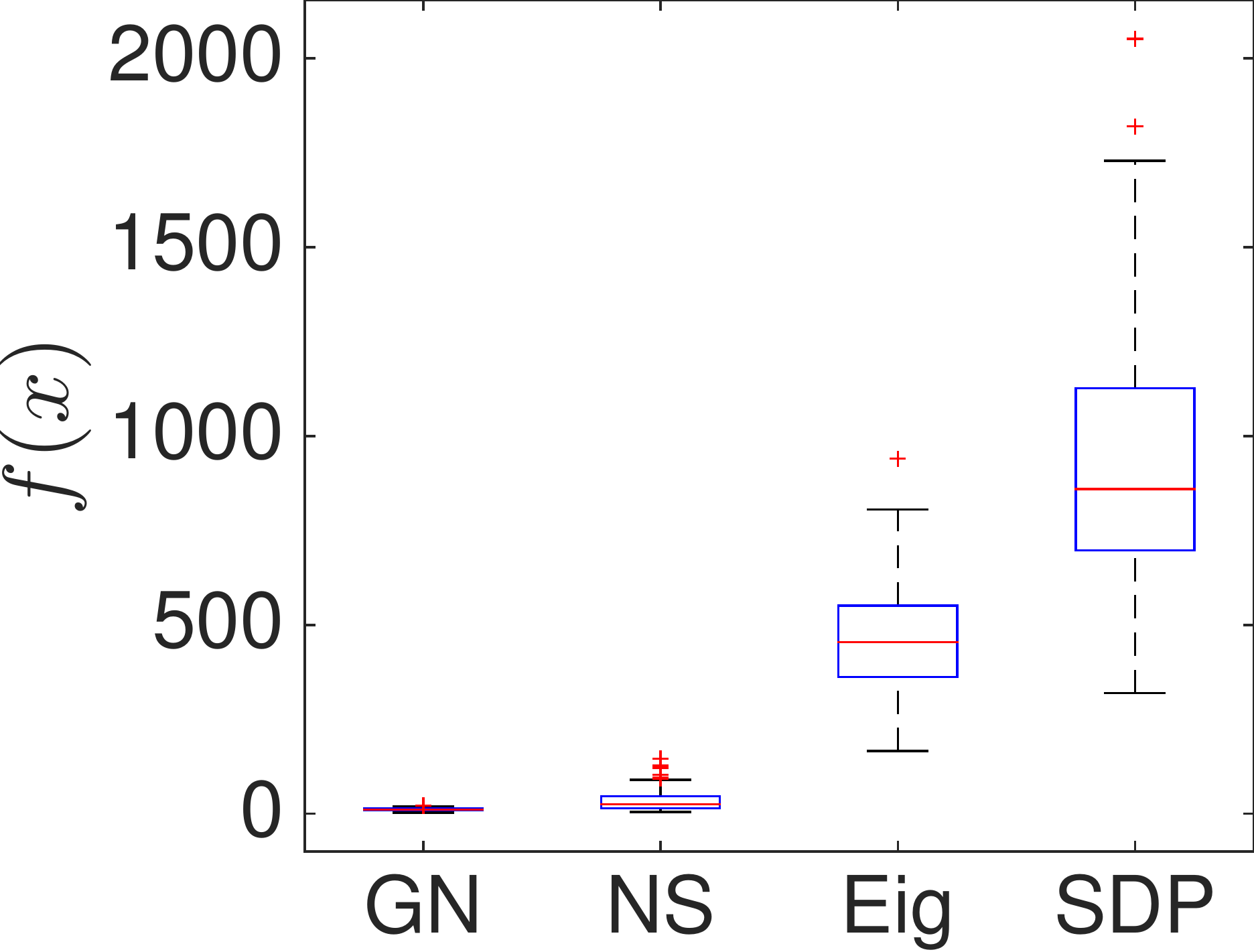} 
\vspace{-0.2cm} \\  {\mySmallFont $\sigma_R = 1$, $\sigma_\Delta = 0.1$} \\ (a1)
\end{minipage}%
&
\begin{minipage}{\sizeCol}%
\centering
\hspace{-4mm} \includegraphics[scale=\scaleFig]{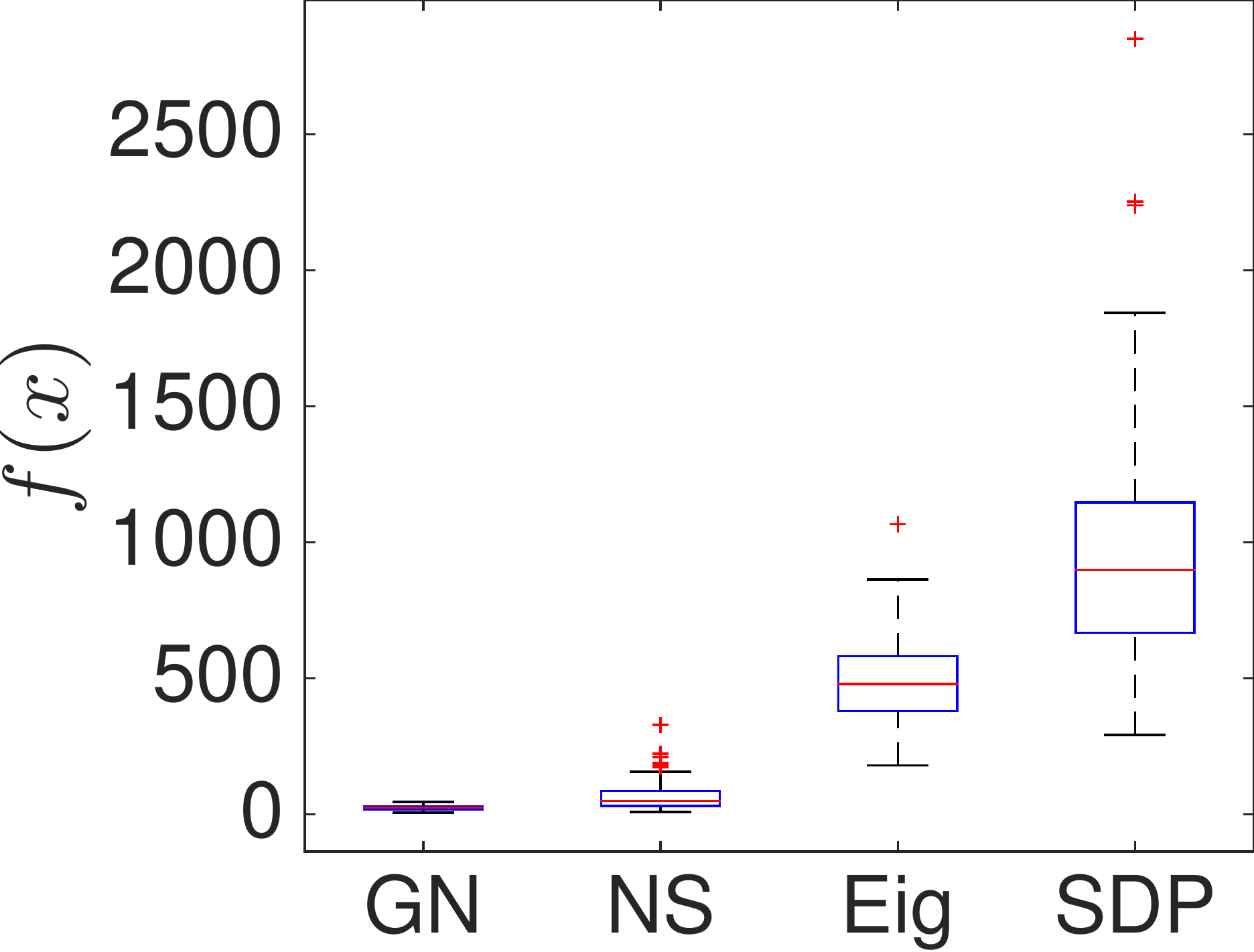}  
\vspace{-0.2cm} \\  {\mySmallFont $\sigma_R = \mbox{unif.}$, $\sigma_\Delta = 0.1$} \\ (a2)
\end{minipage}%
&
\begin{minipage}{\sizeCol}%
\centering
\hspace{-4mm} \includegraphics[scale=\scaleFig]{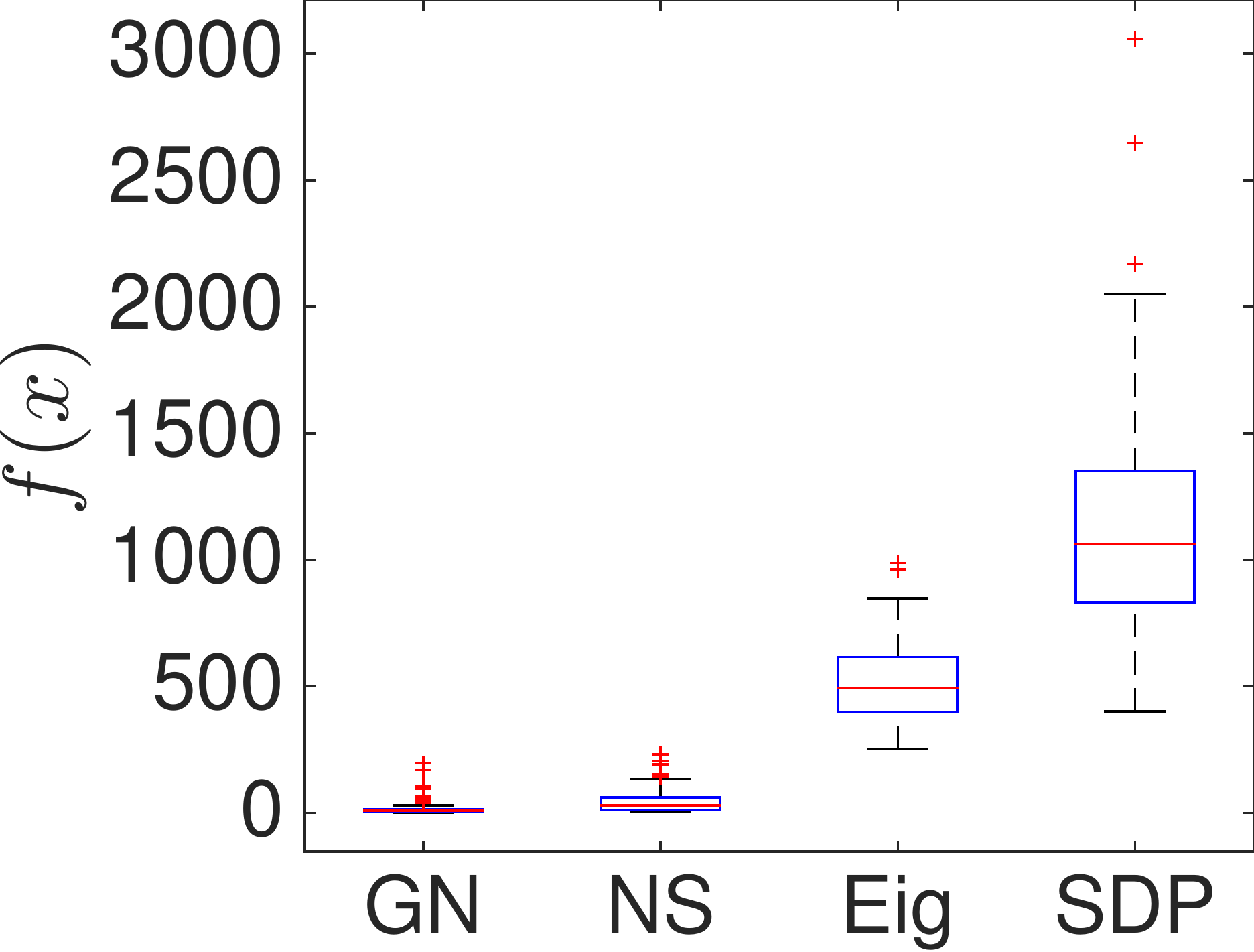} 
\vspace{-0.2cm} \\  {\mySmallFont $\sigma_R = 0.1$, $\sigma_\Delta = 1$} \\ (a3)
\end{minipage}%
&
\begin{minipage}{\sizeCol}%
\centering
\hspace{-4mm} \includegraphics[scale=\scaleFig]{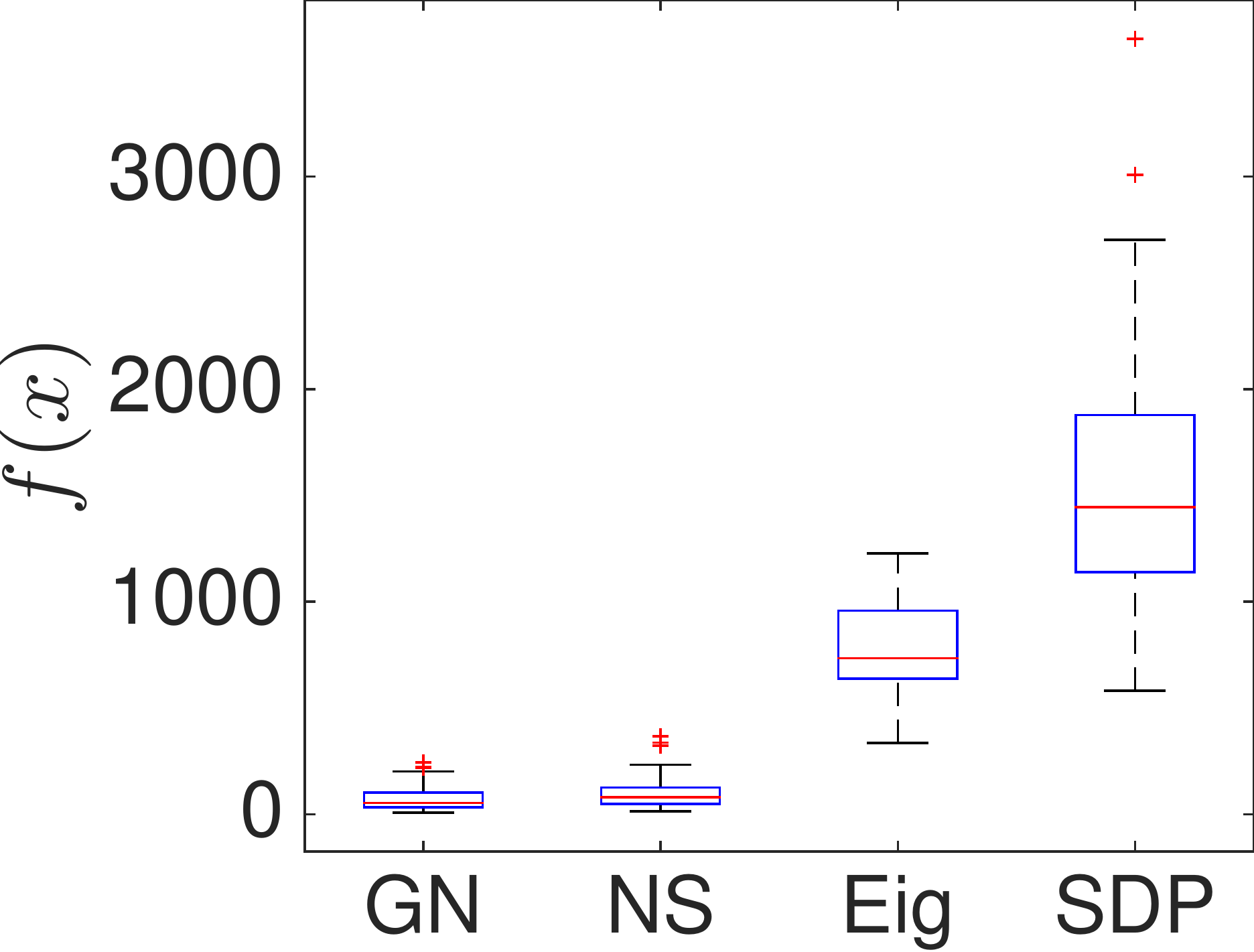} 
\vspace{-0.2cm} \\  {\mySmallFont $\sigma_R = 0.1$, $\sigma_\Delta = \mbox{unif.}$} \\ (a4)
\end{minipage}%
\\
\\
\begin{minipage}{\sizeCol}%
\centering
\hspace{-4mm} \includegraphics[scale=\scaleFig]{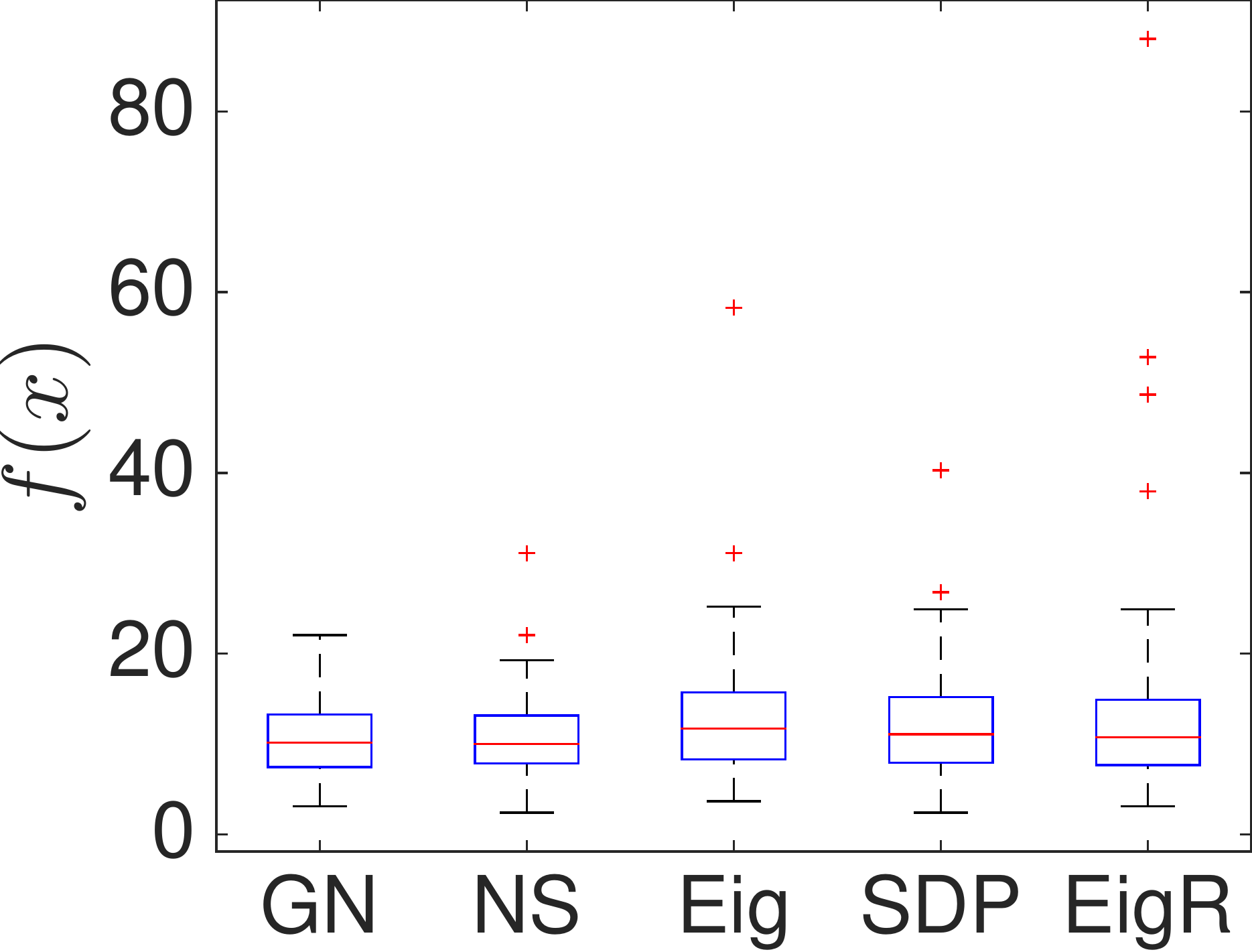} 
\vspace{-0.2cm} \\  {\mySmallFont $\sigma_R = 1$, $\sigma_\Delta = 0.1$} \\ (b1)
\end{minipage}%
&
\begin{minipage}{\sizeCol}%
\centering
\hspace{-4mm} \includegraphics[scale=\scaleFig]{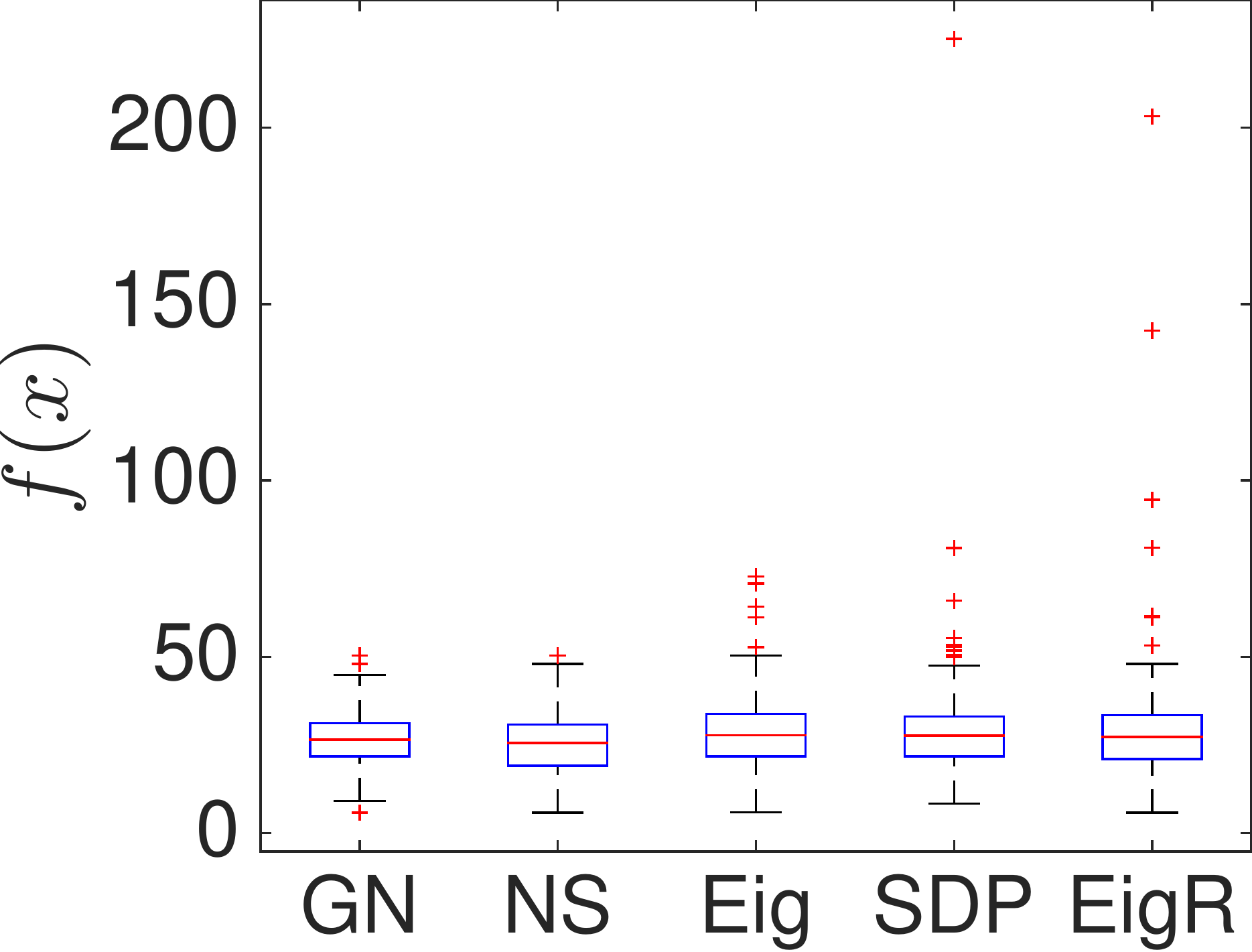}  
\vspace{-0.2cm} \\  {\mySmallFont $\sigma_R = \mbox{unif.}$, $\sigma_\Delta = 0.1$} \\ (b2)
\end{minipage}%
&
\begin{minipage}{\sizeCol}%
\centering
\hspace{-4mm} \includegraphics[scale=\scaleFig]{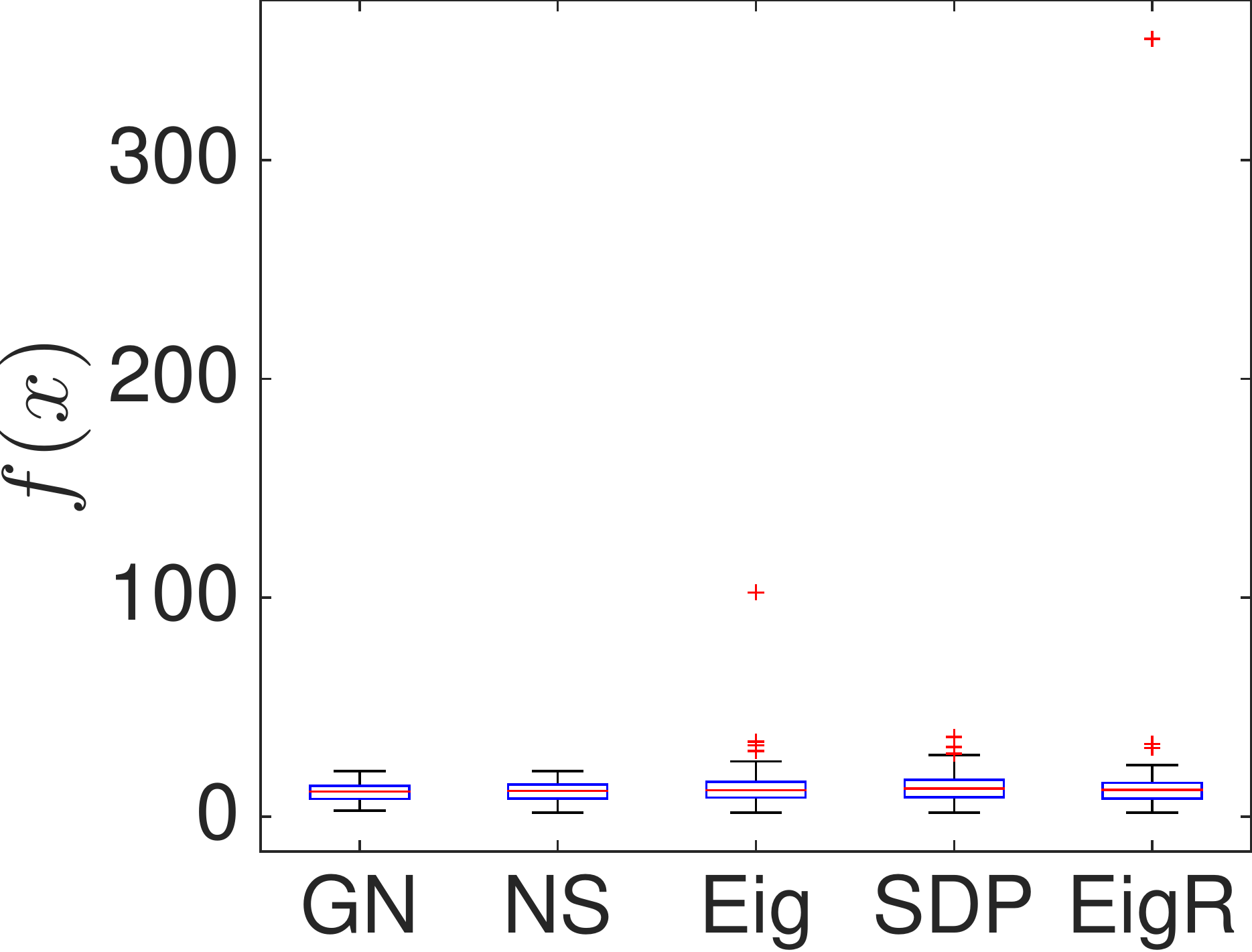} 
\vspace{-0.2cm} \\ {\mySmallFont $\sigma_R = 0.1$, $\sigma_\Delta = 1$} \\ (b3)
\end{minipage}%
&
\begin{minipage}{\sizeCol}%
\centering
\hspace{-4mm} \includegraphics[scale=\scaleFig]{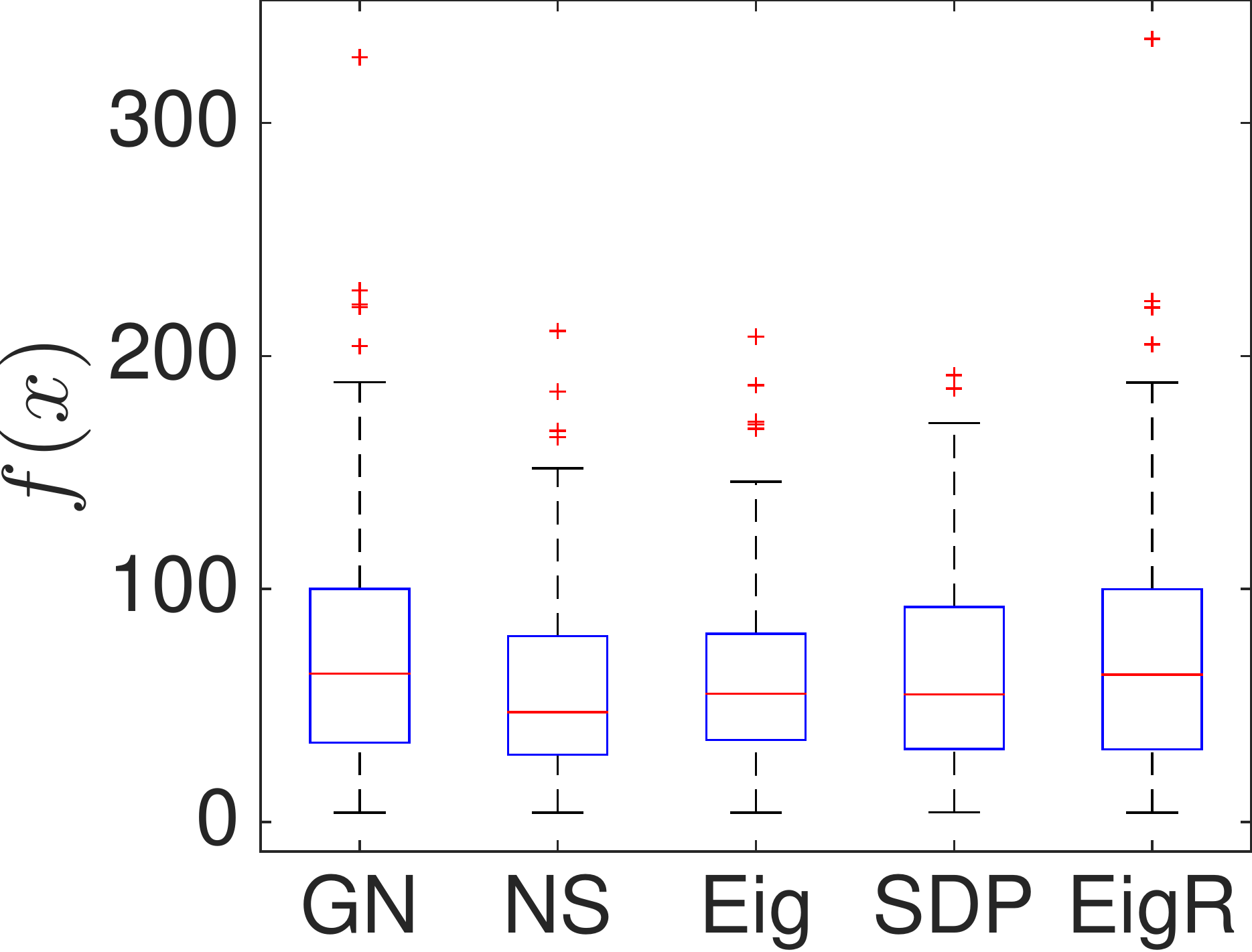} 
\vspace{-0.2cm} \\  {\mySmallFont $\sigma_R = 0.1$, $\sigma_\Delta = \mbox{unif.}$} \\ (b4)
\end{minipage}%
\end{tabular}%
\end{minipage}%
\caption{Statistics on tests in which the \SZEP is \emph{not} satisfied:
(a1)-(a4): Comparison of different \PGO solvers for different levels of noise. 
The compared approaches are:
a Gauss-Newton method initialized at the 
ground truth poses (\GN), the proposed null space approach (\NS), 
the eigenvector method (\Eig), the SDP relaxation (\SDP).
(b1)-(b4): Comparison of the techniques 
\GN, \NS, \Eig, \SDP, refined with a Gauss-Newton method, and 
an alternative approach which solves for rotations first (\EigR).
 \label{fig:subopt}}
\end{figure}







 
\toCheck{
 \emph{Case 2: $\Wcpx(\lam^\star)$ does \emph{not} satisfy the \SZEP.}
 In this case, Algorithm~\ref{alg:primalViaDual} computes an estimate, according to
 the \emph{null space approach} proposed in Section~\ref{sec:algoNOSZEP}; 
 we denote this approach with the label \NS. 
 To evaluate the performance of the proposed approach, we considered 
 100 instances in which the \SZEP was \emph{not} satisfied and we compared our approach
 against the following methods:
 a Gauss-Newton method initialized at the ground truth poses (\GN), 
 the eigenvector method described at the beginning of Section~\ref{sec:algoNOSZEP} (\Eig), and
 the SDP relaxation of Section~\ref{sec:sdp} (\SDP).
 For the \SDP approach, we compute the solution $\Xcpx^\star$ of the relaxed problem~\eqref{eq:primal_sdp_rel}.
 If $\Xcpx^\star$ has rank larger than 1,  we find the closest rank-1 matrix  $\Xcpx_{\text{rank-1}}$ from singular value decomposition~\cite{Eckart36}.
 Then we factorize $\Xcpx_{\text{rank-1}}$ as $\Xcpx_{\text{rank-1}} = \xcpx \xcpx^*$ 
 ($\xcpx$ can be computed via Cholesky factorization of $\Xcpx_{\text{rank-1}}$~\cite{Singer10achm}).
 We report the results of our comparison in the first row of Fig.~\ref{fig:subopt}, where we 
 show for different noise setups (sub-figures (a1) to (a4)), the cost of the estimate produced by the 
 four approaches. The proposed null space approach (\NS) largely outperforms 
 the \Eig  and the \SDP approaches, and has comparable performance with an ``oracle'' \GN approach 
 which knows the ground truth poses. 
}

\toCheck{
 One may also compare the performance of the approaches \NS, \Eig, \SDP after refining the corresponding 
 estimates with a Gauss-Newton method, which tops off residual errors.
 The cost obtained by the different techniques, with the Gauss-Newton refinement, are shown in the second row of 
 Fig.~\ref{fig:subopt}. 
 For this case we also added one more initialization technique in the comparison: 
 we consider an approach that solves for rotations first, using the eigenvalue method in~\cite{Singer10achm}, 
 and then applies the Gauss-Newton method from the rotation guess.\footnote{This was not included 
 in the first row of Fig.~\ref{fig:subopt} as it does not provide a guess for the positions 
 of the nodes.}
 Fig.~\ref{fig:subopt}(b1) to Fig.~\ref{fig:subopt}(b4) show less differences (in average) 
 among the techniques, as in most cases the Gauss-Newton refinement is able to converge starting from all the compared 
 initializations. However, for the techniques \Eig, \SDP, and \EigR we see many red sample points, which denote cases 
 in which the error is larger than the 75th percentile; these are the cases in which the techniques failed to converge and 
 produced a large cost. 
 On the other hand, the proposed \NS approach is less prone to converge to a bad minimum (fewer and 
 lower red samples).
 }

\paragraph{Chain graph counterexample and discussion.}
\begin{figure}[t]
\begin{minipage}{\textwidth}
\begin{tabular}{cc}
\hspace{-10mm}
\begin{minipage}{5cm}%
\centering
\includegraphics[scale=0.35]{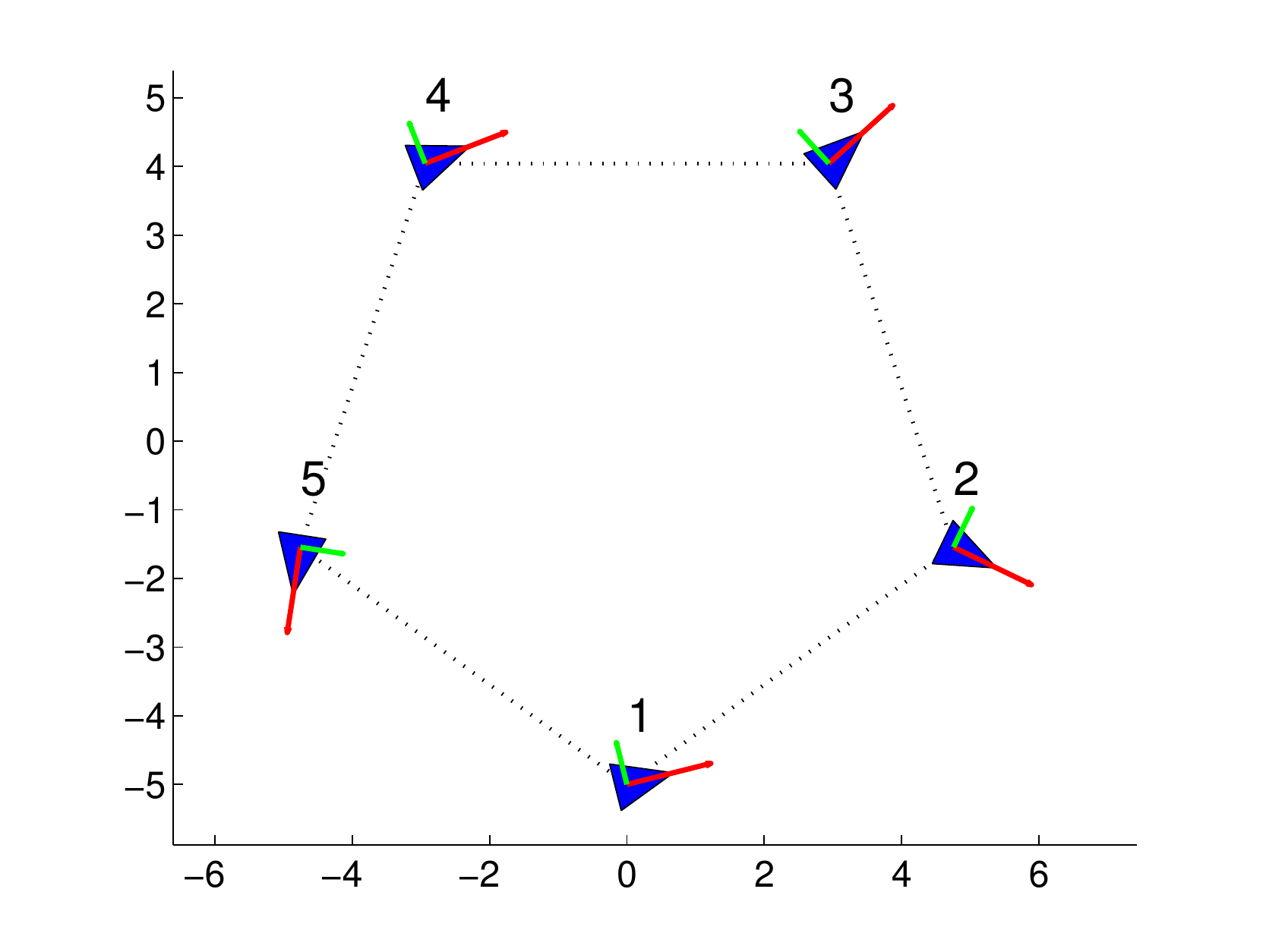}  
\vspace{-1cm}
\end{minipage}%
& 
\hspace{-7mm} \begin{tabular}{cc}
\begin{minipage}{3cm}%
\centering
\includegraphics[scale=0.18]{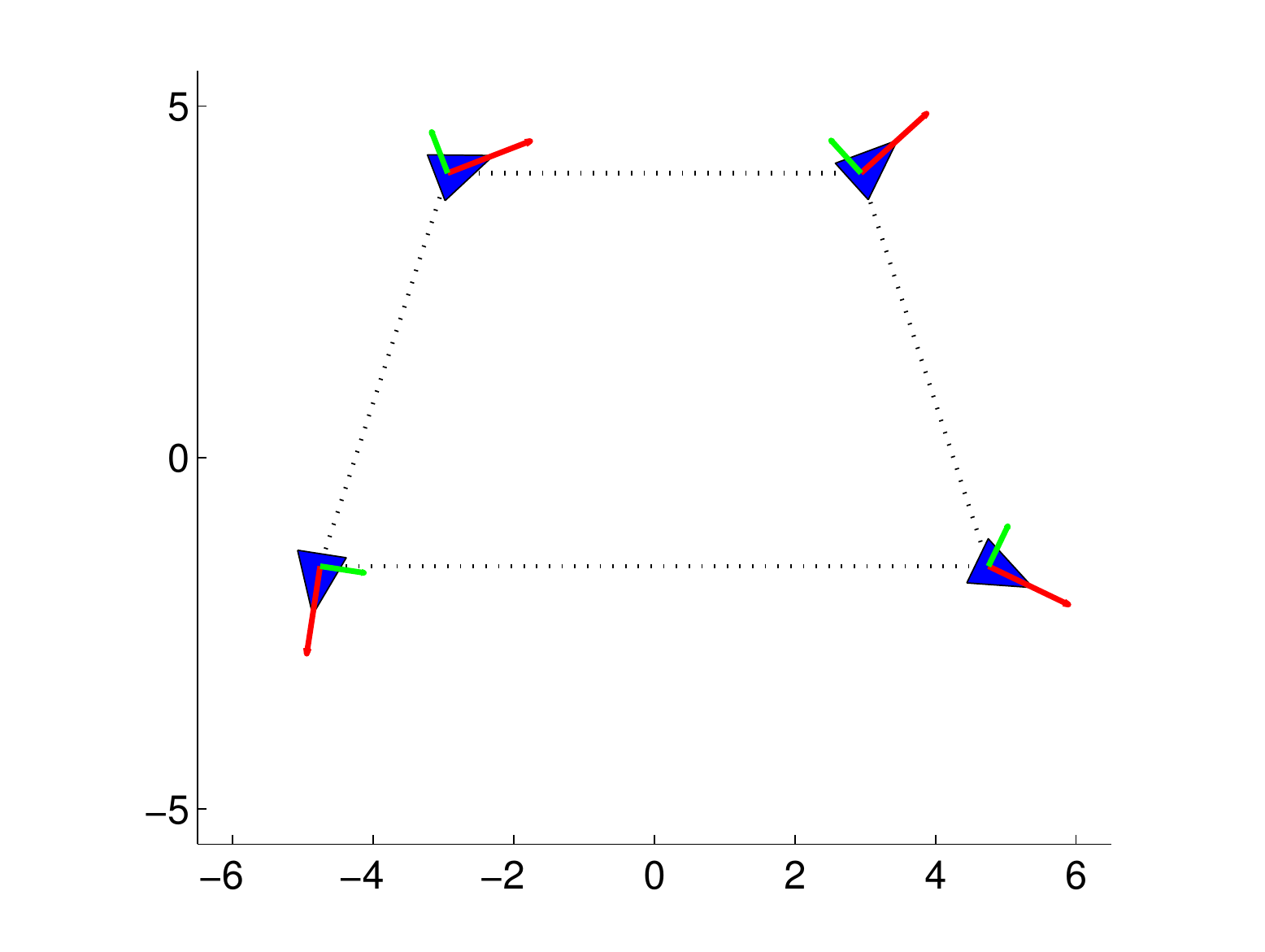} 
\\
\tiny
\begin{tabular}{c|c} %
$\mu_1$ & \red{-1.40e-08} \\ \hline 
$\mu_2$ & 3.33e-03 \\ \hline 
$\mu_3$ & 6.74e-02  \\ \hline 
$\mu_4$ & 4.07e+01
\end{tabular}%
\\
(b)
\end{minipage}%
&
\hspace{-7mm}\begin{minipage}{3cm}%
\centering
\includegraphics[scale=0.18]{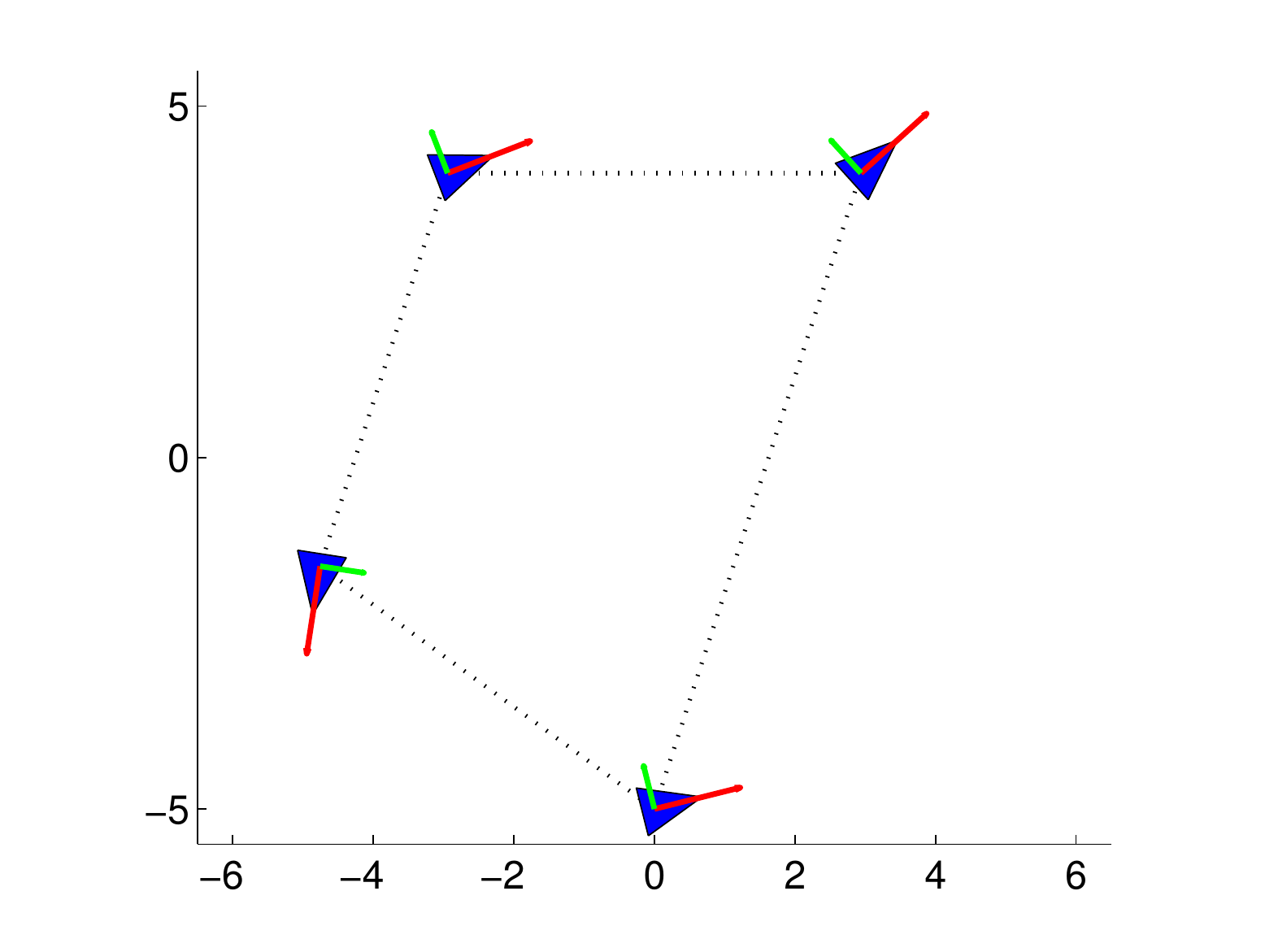} 
\\
\tiny
\begin{tabular}{c|c} %
$\mu_1$ & \red{-5.19e-09}  \\ \hline 
$\mu_2$ & 5.94e-03\\ \hline 
$\mu_3$ & 7.59e-02 \\ \hline 
$\mu_4$ & 4.26e+01
\end{tabular}%
\\
(c)
\end{minipage}%
\end{tabular}
\\
\begin{minipage}{3cm}%
\small
\begin{tabular}{c|c}
$\mu_1$ & \red{-2.87e-09} \\ \hline 
$\mu_2$ & \red{4.90e-09} \\ \hline 
$\mu_3$ & 2.69e-02  \\ \hline 
$\mu_4$ & 1.12e-01
\end{tabular}%
\\
\centering (a)
\end{minipage}%
& 
\begin{tabular}{ccc}
\begin{minipage}{3cm}%
\centering
\includegraphics[scale=0.18, trim = 10mm 0mm 0mm 0mm, clip]{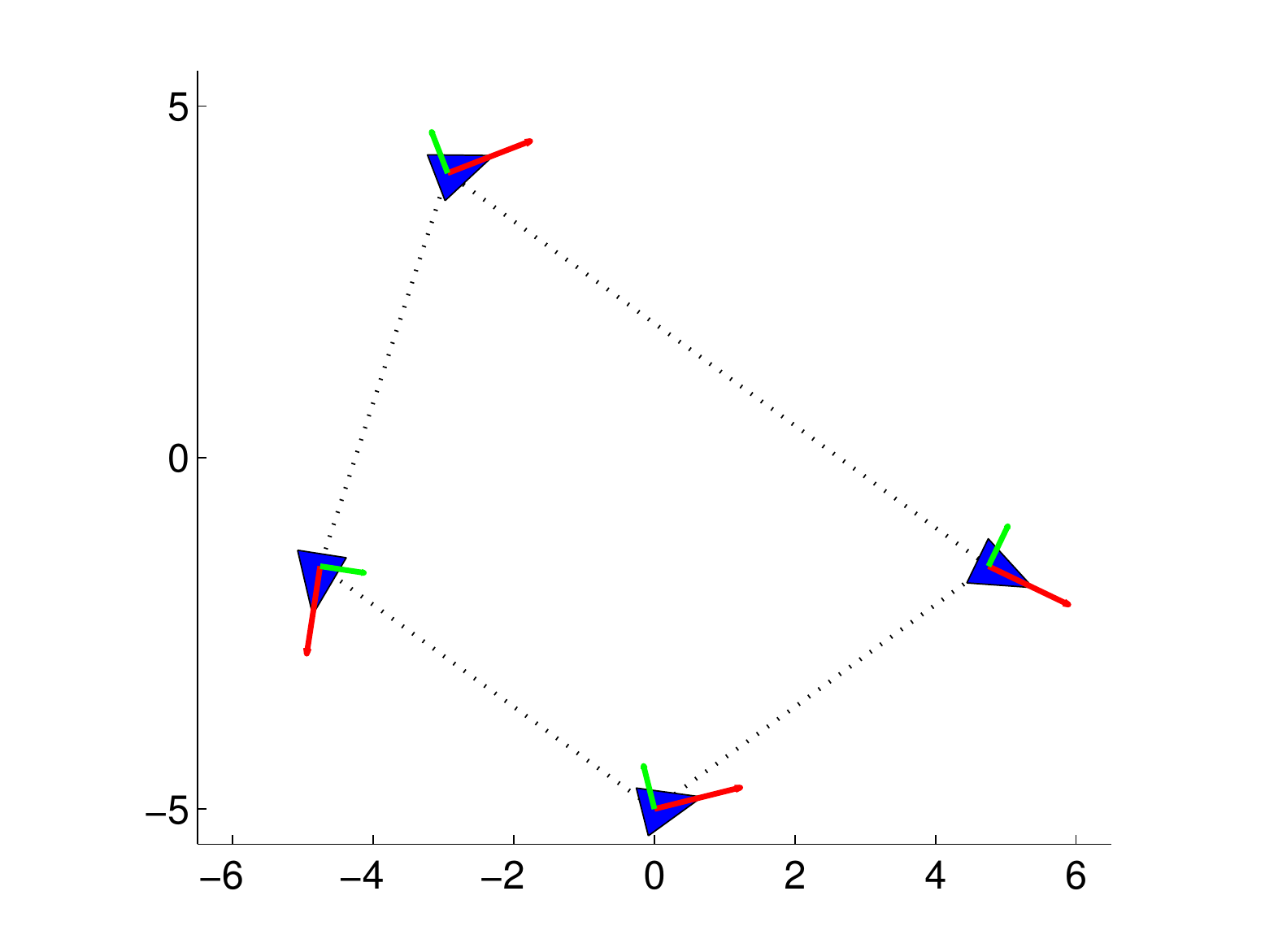} 
\\
\tiny
\begin{tabular}{c|c} %
$\mu_1$ & \red{-1.03e-07} \\ \hline 
$\mu_2$ & \red{8.14e-08} \\ \hline 
$\mu_3$ & 8.82e-02 \\ \hline 
$\mu_4$ & 2.46e+01
\end{tabular}%
\\
(d)
\end{minipage}%
&
\hspace{-10mm}\begin{minipage}{3cm}%
\centering
\includegraphics[scale=0.18, trim = 20mm 0mm 0mm 0mm, clip]{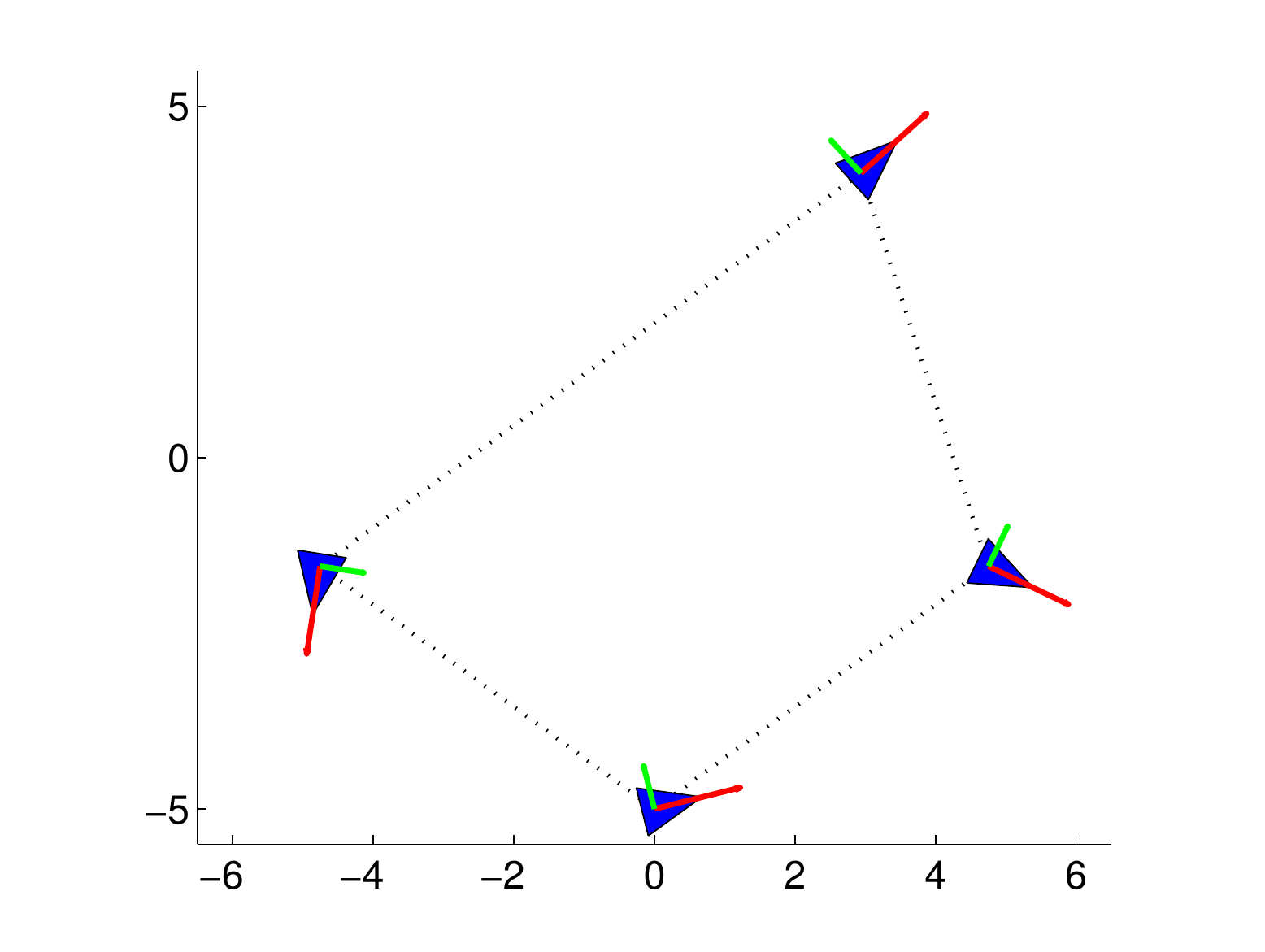} 
\\
\tiny
\begin{tabular}{c|c} %
$\mu_1$ & \red{-4.13e-08} \\ \hline 
$\mu_2$ & 5.29e-03 \\ \hline 
$\mu_3$ & 4.33e-02\\ \hline 
$\mu_4$ &2.40e+01
\end{tabular}%
\\
(e)
\end{minipage}%
&
\hspace{-10mm}\begin{minipage}{3cm}%
\centering
\includegraphics[scale=0.18, trim = 20mm 0mm 0mm 0mm, clip]{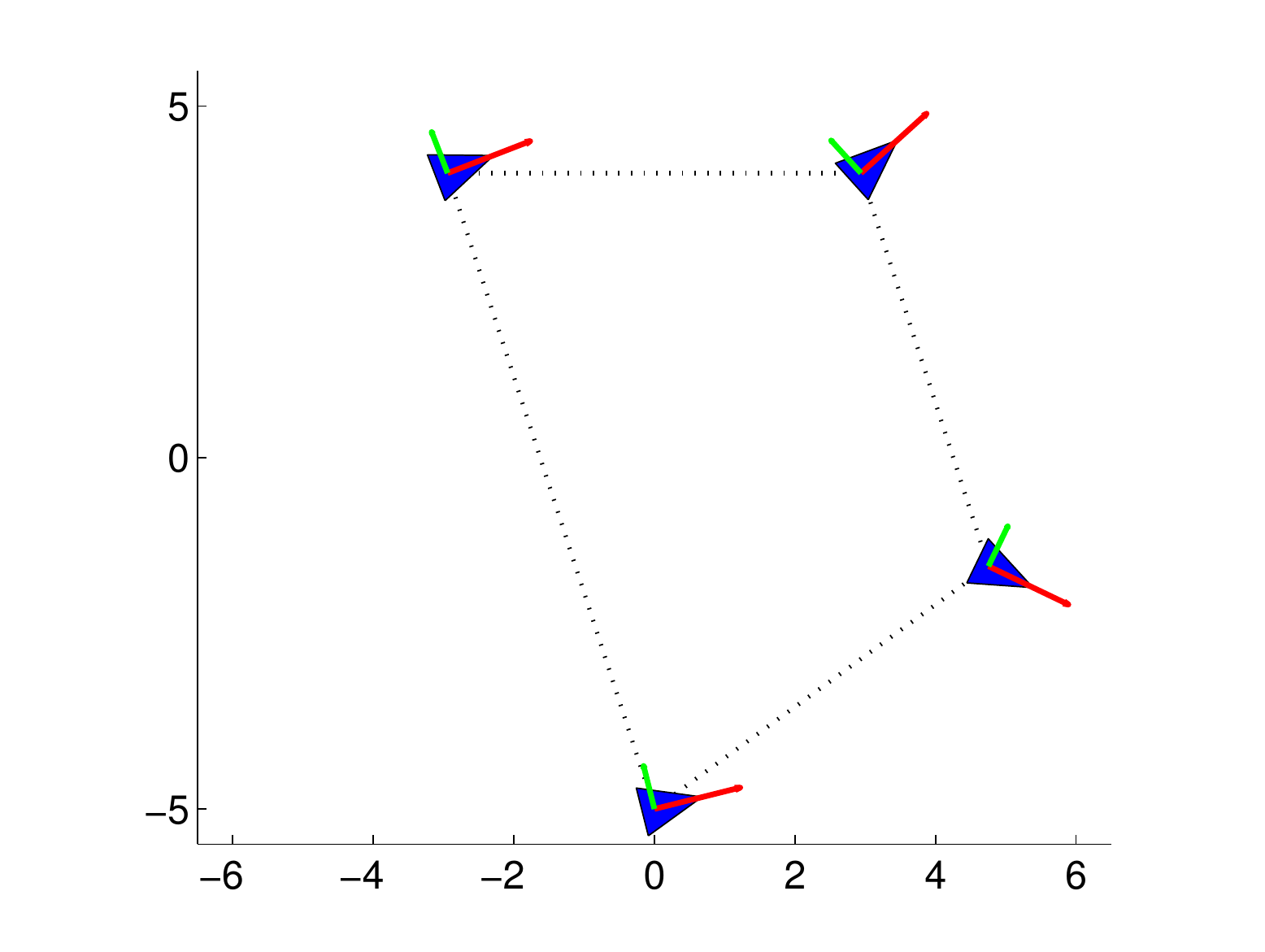} 
\\
\tiny
\begin{tabular}{c|c} %
$\mu_1$ & \red{1.78e-10} \\ \hline 
$\mu_2$ & 5.14e-03 \\ \hline 
$\mu_3$ & 8.43e-02   \\ \hline 
$\mu_4$ & 1.28e+01
\end{tabular}%
\\
(f)
\end{minipage}%
\end{tabular}
\end{tabular}%
\end{minipage}%
\caption{\label{fig:conterexample} 
(a) Toy example of chain pose graph in which the \SZEP fails. In each plot we also report the
 four smallest eigenvalues of the penalized pose graph matrix $\Wcpx(\lam^\star)$ for the corresponding \PGO problem.
 Removing a node from the original graph may change the duality properties of the graph.
In (b), (c), (d), (e), (f) we remove nodes 1, 2, 3, 4, 5, respectively. 
Removing any node, except node 3, leads to a graph that satisfied the \SZEP.
}
\end{figure}

In this section we consider a simple graph topology: the \emph{chain graph}.
A chain graph is a graph with edges $(1,2),(2,3),\ldots,(n-1,n),(n,1)$.
Removing the last edge we obtain a tree (or, more specifically, a \emph{path}), for which the \SZEP 
is always satisfied. 
Therefore the question is: \emph{is the \SZEP always satisfied in \PGO whose underlying graph is a chain?}
The answer, unfortunately, is no.
Fig.~\ref{fig:conterexample}(a) provides an example of a very simple chain graph with 5 nodes 
that fails to meet the \SZEP property. The figure reports the 4 smallest eigenvalues of 
$\Wcpx(\lam^\star)$ ($\mu_1,\ldots,\mu_4$), and the first two are numerically zero.

If the chain graph were balanced, Proposition~\ref{prop:nogapTreesBalanced} says that the \SZEP needs to be satisfied.
Therefore, one may argue that failure to meet the \SZEP depends on the amount of error accumulated along 
the loop in the graph. Surprisingly, also this intuition fails. In Fig.~\ref{fig:conterexample}(b-f) 
we show the pose graphs obtained by removing a single node from the pose graph in Fig.~\ref{fig:conterexample}(a).
When removing a node, say $k$, we introduce a relative measurement between nodes $k-1$ and $k+1$, that is equal to 
the composition of the relative measurements associated to the edges $(k-1,k)$ and $(k,k+1)$ in the original graph.
By constructions, the resulting graphs have the same accumulated errors (along each loop) as the original graph. 
However, interestingly, they do not necessarily share the same duality properties of the original graph. 
The graphs obtained by removing nodes $1,2,4,5$ (shown in figures b,c,e,f, respectively), in fact, satisfy the 
\SZEP. On the other hand, the graph in Fig.~\ref{fig:conterexample}(c) still has 2 eigenvalues in zero.  The data to reproduce these toy examples are reported in Appendix~\ref{sec:app:numericaldata}.

\begin{figure}[t]
\centering
\includegraphics[scale=0.32]{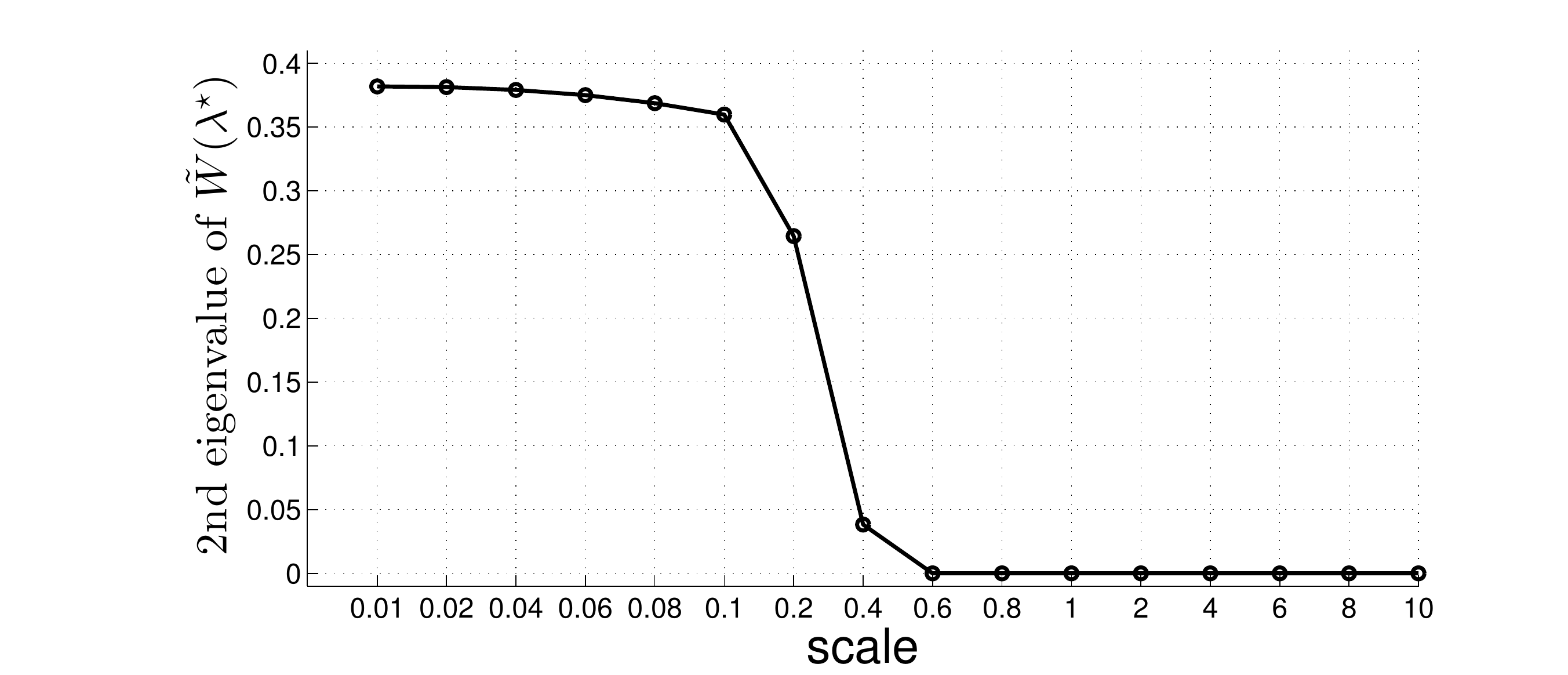} \vspace{-0.3cm}
\caption{\label{fig:scaling}Second eigenvalue of the matrix $\Wcpx(\lam^\star)$ for different 
variations of the toy graph of Fig.~\ref{fig:conterexample}(a). Each variation is 
obtained by scaling the translation measurements of the original graph by the amount 
specified on the x-axis of this figure. When the scale of the measurement is $\leq 0.4$ the 
 second eigenvalue of $\Wcpx(\lam^\star)$ is larger than zero, hence the \SZEP is satisfied. 
}
\end{figure}
 
 We conclude with a test showing that the \SZEP is not only dictated by the underlying rotation subproblem 
but also depends heavily on the translation part of the optimization problem. 
 To show this we consider variations of the \PGO problem in Fig.~\ref{fig:conterexample}(a), 
 in which we ``scale'' all translation measurements by a constant factor. 
 When the scale factor is smaller than one we obtain a \PGO problem in which nodes are closer to each other; 
 for scale $>1$ we obtain larger inter-nodal measurements; the scale equal to 1 coincides with the original problem.
 Fig~\ref{fig:scaling} shows the second eigenvalue of $\Wcpx(\lam^\star)$ for different scaling of the original 
 graphs. Scaling down the measurements in the graph of Fig.~\ref{fig:conterexample}(a) can re-establish the 
 \SZEP. 
 Interestingly, this is in agreement with the convergence analysis of~\cite{Carlone13icra}, which 
 shows that the basin of convergence becomes larger when scaling down the 
 inter-nodal distances.

\section{Conclusion}

We show that the application of Lagrangian duality in \PGO provides an appealing approach 
to compute a globally optimal solution. 
More specifically, we propose four contributions. 
First, we rephrase \PGO as a problem in complex variables. This allows drawing connection with the 
recent literature on \emph{unit gain} graphs, and enables results on the spectrum of the 
pose graph matrix. Second, we formulate the Lagrangian dual problem and we analyze the 
relations between the primal and the dual solutions. Our key result proves that the duality gap is 
connected to the number of eigenvalues of the \emph{penalized pose graph matrix}, which 
arises from the solution of the dual problem. In particular, if this matrix has a 
\emph{single eigenvalue in zero} (\SZEP), then (i) the duality gap is zero, (ii) the primal \PGO problem has a 
unique solution (up to an arbitrary roto-translation), and 
(iii) the primal solution can be computed by \emph{scaling} an eigenvector 
of the penalized pose graph matrix. The third contribution is an algorithm 
that returns a guaranteed optimal solution when the \SZEP is satisfied, and 
 (empirically) provides a very good estimate when the \SZEP fails.
Finally, we report numerical results, that 
  show that (i) the \SZEP holds for noise levels of practical robotics applications, 
 (ii) the proposed algorithm outperforms several existing approaches,
 (iii) the satisfaction of the \SZEP depends on  multiple factors, including graph connectivity, 
 number of poses, and measurement noise.


\section{Appendix}

\subsection{Proof of Proposition~\ref{prop:zeroCostTrees}: Zero Cost in Trees}
\label{sec:proof:zeroCostTrees}

 We prove Proposition~\ref{prop:zeroCostTrees} by inspection, providing a procedure to build
   an estimate that annihilates every summand in~\eqref{eq:PGOr}.
  The procedure is as follows:
\begin{enumerate}
\item Select a root node, say the first node $(p_i,r_i)$, with $i=1$, and set it to the origin, i.e., $p_i = 0_2$, 
 $r_i = [1 \; 0]\tran$ (compare with~\eqref{eq:rparametrization} for $\theta_i = 0$);
\item \label{point2} For each neighbor $j$ of the root $i$, if $j$ is an outgoing neighbor, 
set $r_j = \Rij r_i$, and $p_j = p_i + \Dij r_i$, otherwise set
$r_j = \Rji\tran r_i$, and $p_j = p_i + \Dji r_j$;
\item Repeat point~\ref{point2} for the unknown neighbors of every node that has been computed so far, 
and continue until all poses have been computed.
\end{enumerate}

Let us now show that this procedure produces a set of poses that annihilates the objective in~\eqref{eq:PGOr}.
 According to the procedure, we set the first node to the origin: 
$p_1 = 0_2$,  $r_1 = [1 \; 0]\tran$;
then, before moving to the second step of the procedure, we rearrange the terms in~\eqref{eq:PGOr}: 
we separate the edges into two sets $\calE = \calE_1 \cup \bar{\calE}_1$, 
where $\calE_1$ is the set of edges incident on node $1$ (the root), and 
$\bar{\calE}_1$ are the remaining edges. Then the cost can be written as:
\bea
\label{eq:PGOr-proof1}
f(p,r) = 
\sum_{(i,j)\in \calE_1} \| \diffp - D_{ij}r_i\|_2^2 + \|r_j - \Rij r_i \|_2^2 +  \nonumber \\
 + \sum_{(i,j)\in \bar{\calE}_1} \| \diffp - D_{ij}r_i \|_2^2 + \|r_j - \Rij r_i\|_2^2
\eea
We can further split the set $\calE_1$ into edges that have node $1$ as a tail (i.e., edges 
in the form $(1,j)$) and edges that have node $1$ as head (i.e., $(j,1)$):
\bea
\label{eq:PGOr-proof2}
f(p,r) = 
\sum_{(1,j), j \in \calNout_1} \| p_j - p_1 - D_{1j}r_1\|_2^2 + \| r_j - R_{1j} r_1\|_2^2 + \nonumber \\
+ \sum_{(j,1), j \in \calNin_1} \| p_1 - p_j - D_{j1}r_j \|_2^2 + \| r_1 - R_{j1} r_j \|_2^2 + \nonumber \\
 + \sum_{(i,j)\in \bar{\calE}_1} \| \diffp - D_{ij}r_i \|_2^2 + \|r_j - \Rij r_i\|_2^2
\eea
Now, we set each node $j$ in the first two summands as prescribed in step 2
 of the procedure. By inspection one can verify that this 
choice annihilates the fist two summands and the cost becomes:
\bea
\label{eq:PGOr-proof3}
f(p,r) = \sum_{(i,j)\in \bar{\calE}_1} \| \diffp - D_{ij}r_i\|_2^2 + \|r_j - \Rij r_i\|_2^2
\eea
Now we select a node $k$ that has been computed at the previous step, but has some neighbor 
that is still unknown. As done previously, we split the set $\bar{\calE}_1$ into 
two disjoint subsets: $\bar{\calE}_1 = \calE_k \cup \bar{\calE}_k$, where 
the set  $\calE_k$ contains the edges in $\bar{\calE}_1$ that are incident on $k$, 
and $\bar{\calE}_k$ contains the remaining edges:
%
\bea
\label{eq:PGOr-proof4}
f(p,r) = 
\sum_{ \{ (k,j), j \in \calNout_k \} \cap \bar{\calE}_1 } 
\|p_j - p_k - D_{kj}r_k\|_2^2 + \|r_j - R_{kj} r_k \|_2^2 +  \nonumber\\
 + \sum_{ \{ (j,k), j \in \calNin_k \} \cap \bar{\calE}_1 } 
 \|p_k - p_j - D_{jk}r_j\|_2^2 + \|r_k - R_{jk} r_j \|_2^2 + \nonumber\\
  + \sum_{(i,j)\in \bar{\calE}_k} \| \diffp - D_{ij}r_i\|_2^2 + \|r_j - \Rij r_i\|_2^2
\eea
Again, setting neighbors $j$ as prescribed in  step 2 of the procedure, 
annihilates the first two summands in~\eqref{eq:PGOr-proof4}. 
 Repeating the same reasoning for all nodes that have been computed, but still have unknown 
 neighbors, we can easily show that all terms in~\eqref{eq:PGOr-proof4} become zero 
 (the assumption of graph connectivity ensures that we can reach all nodes), 
 proving the claim. 

 \subsection{Proof of Proposition~\ref{prop:zeroCostBalancedGraphs}: Zero Cost in Balanced Graphs}
\label{sec:proof:zeroCostBalancedGraphs}

Similarly to Appendix~\ref{sec:proof:zeroCostTrees}, we 
 prove Proposition~\ref{prop:zeroCostBalancedGraphs} by showing that in balanced graphs one can always build a
solution that attains zero cost.

For the assumption of connectivity, we can find a spanning tree $\calT$ of the graph, 
and split the terms in the cost function accordingly:
\bea
\label{eq:balanced-proof1}
f(p,r) = 
\sum_{(i,j)\in \calT} \| \diffp - D_{ij}r_i\|_2^2 + \|r_j - \Rij r_i\|_2^2 +  \nonumber \\
 + \sum_{(i,j)\in \bar\calT} \| \diffp - D_{ij}r_i\|_2^2 + \|r_j - \Rij r_i\|_2^2
\eea
where $\bar\calT \doteq \calE \setminus \calT$ are the \emph{chords} of the graph w.r.t. $\calT$.

Then, using the procedure in Appendix~\ref{sec:proof:zeroCostTrees} we construct 
a solution $\{r_i^\star,p_i^\star\}$ that attains zero cost for the measurements in 
the spanning tree $\calT$.
Therefore, our claim only requires to demonstrate that the solution built from the 
spanning tree also annihilates the terms in $\bar\calT$:
\bea
\label{eq:balanced-proof1}
f(p^\star,r^\star) =  \sum_{(i,j)\in \bar\calT} \|p_j^\star - p_i^\star - D_{ij}r_i^\star \|_2^2 + \|r_j^\star - \Rij r_i^\star \|_2^2
\eea

To prove the claim, we consider one of the chords in $\bar\calT$ and we show that 
the cost at $\{r_i^\star,p_i^\star\}$ is zero. 
The cost associated to a chord $(i,j) \in \bar\calT$ is:
\beq
\label{eq:costChord}
\|p_j^\star - p_i^\star - D_{ij}r_i^\star \|_2^2 + \|r_j^\star - \Rij r_i^\star \|_2^2
\eeq
Now consider the unique path $\calP_{ij}$
in the spanning tree $\calT$ that connects $i$ to $j$, and number the nodes 
along this path as $i, i+1, \ldots, j-1, j$.  

Let us start by analyzing the second summand in~\eqref{eq:costChord}, which corresponds to 
the rotation measurements.
According to the procedure in Appendix~\ref{sec:proof:zeroCostTrees} to build the solution 
for $\calT$, we propagate the estimate from the root of the tree. 
 Then it is easy to see that:
 \beq
 \label{eq:pathCompositionRot}
 r^\star_j = R_{j-1j} \cdots R_{i+1i+2} R_{ii+1} r^\star_i
 \eeq
where $R_{ii+1}$ is the rotation associated to the edge $(i,i+1)$, 
 or its transpose if the edge is in the form $(i+1,i)$ (i.e., it is traversed backwards along $\calP_{ij}$).
 Now we notice that the assumption of balanced graph implies that the 
 measurements compose to the identity along every cycle in the graph. 
 Since the chord $(i,j)$ and the path $\calP_{ij}$ form a cycle in the graph, 
 it holds: 
 \beq
 \label{eq:loopCompositionRot}
 R_{j-1j} \cdots R_{i+1i+2} R_{ii+1} = R_{ij}
 \eeq
 Substituting~\eqref{eq:loopCompositionRot} back into~\eqref{eq:pathCompositionRot}
 we get:
 \beq
 r^\star_j = R_{ij} r^\star_i
 \eeq
 which can be easily seen to annihilate the second summand in~\eqref{eq:costChord}. 
 
 Now we only need to demonstrate that also the first summand in~\eqref{eq:costChord} 
 is zero.
  The procedure in Appendix~\ref{sec:proof:zeroCostTrees} leads to the following 
  estimate for the position of node $j$:
 \bea
 \label{eq:rotComposition}
 p^\star_j &=& p^\star_i + D_{ii+1} r^\star_i + D_{i+1i+2} r^\star_{i+1} + \cdots + D_{j-1j} r^\star_{j-1}  \\
 &=& p^\star_i + D_{ii+1} r^\star_i + D_{i+1i+2} R_{ii+1} r^\star_{i} + \cdots + D_{j-1j} R_{j-2j-1} \cdots R_{i+1i+2} R_{ii+1} r^\star_{i}   \nonumber \\
 &=& p^\star_i + \left( D_{ii+1} + D_{i+1i+2} R_{ii+1}  + \cdots + D_{j-1j} R_{j-2j-1} \cdots R_{i+1i+2} R_{ii+1} \right) r^\star_{i}\nonumber 
 \eea
 The assumption of balanced graph implies that position measurements compose 
 to zero along every cycle, hence:
 \bea
 \label{eq:loopCompositionTran}
 \Delta_{ij} &=& \Delta_{ii+1}  + 
  R_{ii+1} \Delta_{i+1i+2}  + R_{i+1i+2} R_{ii+1}  \Delta_{i+2i+3} + \cdots  \nonumber \\
  &+&  R_{j-2j-1} \cdots R_{i+1i+2} R_{ii+1} \Delta_{j-1j}
 \eea
 or equivalently: 
 \bea
 \label{eq:loopCompositionTran2}
 D_{ij} &=& D_{ii+1} + D_{i+1i+2} R_{ii+1}  + \cdots \nonumber \\
 &+& D_{j-1j} R_{j-2j-1} \cdots R_{i+1i+2} R_{ii+1}
 \eea
 Substituting~\eqref{eq:loopCompositionTran2} back into~\eqref{eq:rotComposition} we obtain: 
 \beq
 p^\star_j = p^\star_i + D_{ij} r^\star_{i}\nonumber 
 \eeq
 which annihilates the first summand in~\eqref{eq:costChord}, concluding the proof.

\subsection{Proof of Proposition~\ref{prop:propertiesW}: properties of $\Wfull$}
\label{sec:proof:propertiesW}

 Let us prove that $\Wfull$ has (at least) two eigenvalues in zero. 
 We already observed that the top-left block of $\Wfull$ is
 $\AugLaplacian = \Laplacian \otimes \eye_2$, where $\Laplacian$ is the Laplacian matrix 
 of the graph underlying the \PGO problem. The Laplacian $\Laplacian$ of a connected graph 
 has a single eigenvalue in zero, and the corresponding eigenvector 
 is $\ones_n$ (see, e.g.,~\cite[Sections 1.2-1.3]{Chung96book}), i.e., 
 $\Laplacian \cdot \ones_n = \zeros$. 
 Using this property, it is easy to show that 
 the matrix $N \doteq [\zeros_n\tran \; \ones_n\tran]\tran \otimes \eye_2$ 
 is in the nullspace of $\Wfull$, i.e., $\Wfull N = 0$. Since $N$ has rank 2, this implies 
 that the nullspace of $\Wfull$ has at least dimension 2, which proves the first claim.

 Let us now prove that the matrix $\Wfull$ is composed by  $2\times 2$ blocks $[\Wfull]\ij$, with 
 $[\Wfull]\ij\in\aSOtwo$, $\forall i,j=1,\ldots,2n$, and $[\Wfull]\ii = \alpha\ii \eye_2$ with $\alpha\ii \geq 0$. 
%
We prove this by direct inspection of the blocks of $\Wfull$.
Given the structure of $\Wfull$ in~\eqref{eq:Wdef}, 
%
the claim reduces to proving that the matrices $\AugLaplacian$, $\bar Q$, and $\bar A\tran \barD$ 
are composed by $2\times 2$ blocks in $\aSOtwo$, and the diagonal blocks of $\AugLaplacian$ and $\bar Q$ 
are multiples of the identity matrix.
To this end, we start by observing that $\AugLaplacian = \Laplacian \otimes \eye_2$, hence 
all blocks in $\AugLaplacian$ are multiples of the $2 \times 2$ identity matrix, which also implies that they
 belong to $\aSOtwo$.
%
Consider  next the matrix $\bar Q \doteq \barD\tran \barD + \barU\tran \barU$. 
From the definition of $\barD$ it follows that 
$\barD\tran \barD$  is zero everywhere, except the 
$2 \times 2$ diagonal blocks:
\beq
\label{eq:DtD}
[\barD\tran \barD]_{ii} = \sum_{j \in \calNout_i} \| \Delta_{ij} \|_2^2  \eye_2,
\qquad i=1,\ldots,n.  
\eeq
Similarly, from simple matrix manipulation we obtain the following 
block structure of $\barU\tran \barU$:
\begin{align}
\label{eq:PtP}
[\barU\tran \barU ]_{ii} &= d_i I_2,    \qquad  i=1,\ldots,n;  \nonumber \\
[\barU\tran \barU ]_{ij} &= -\Rij,      \qquad  (i,j)\in\calE; \nonumber\\
[\barU\tran \barU ]_{ij} &= -R\ji\tran, \qquad (j,i)\in\calE; \nonumber\\
[\barU\tran \barU ]_{ij} &= \zeros_{2 \times 2}, \qquad \text{otherwise}.
\end{align}
where $d_i$ is the degree (number of neighbours) of node $i$. 
Combining~\eqref{eq:DtD} and~\eqref{eq:PtP} we get the following 
structure for $\bar Q$: 
\begin{align}
\label{eq:Qblocks}
[\bar Q]_{ii} &= \beta_i I_2,    \qquad  i=1,\ldots,n;  \nonumber\\
[\bar Q]_{ij} &= -\Rij,      \qquad  (i,j)\in\calE; \nonumber\\
[\bar Q]_{ij} &= -R\ji\tran, \qquad (j,i)\in\calE; \nonumber\\
[\bar Q]_{ij} &= \zeros_{2 \times 2}, \qquad \text{otherwise}.
\end{align}
where we defined $\beta_i \doteq d_i + \sum_{j \in \calNout_i} \| \Delta_{ij} \|_2^2$. 
Clearly, $\bar Q$ has blocks in $\aSOtwo$ and the diagonal blocks are nonnegative multiples of $\eye_2$.

Now, it only remains to inspect the structure of $\bar A\tran \barD$. 
 The matrix $\bar A\tran \barD$ has the following structure: 
\begin{align}
\label{eq:AtD}
 [\bar A\tran \barD]_{ii} &= \textstyle \sum_{j \in \calNout_i} \Dij,    \qquad  i=1,\ldots,n;  \nonumber\\
[\bar A\tran \barD]_{ij} &= -\Dji,      \qquad  (j,i)\in\calE; \nonumber\\
[\bar A\tran \barD]_{ij} &= \zeros_{2 \times 2}, \qquad \text{otherwise}.
\end{align}
%
Note that $\sum_{j \in \calNout_i} \Dij$ is the sum of matrices in $\aSOtwo$, 
hence it also belongs to $\aSOtwo$. Therefore, also all blocks of $\bar A\tran \barD$ are in $\aSOtwo$, thus
concluding the proof.

\subsection{Proof of Proposition~\ref{prop:costComplex}: Cost in the Complex Domain}
\label{sec:proof:costComplex}

Let us prove the equivalence between the complex cost and its real counterpart, 
as stated in Proposition~\ref{prop:costComplex}.

We first observe that the dot product between two 2-vectors
 $x_1,x_2 \in \Real{2}$, can be written in terms of their complex representation 
 $\tilde x_1 \doteq x_1^\vee$, and $\tilde x_2 \doteq x_2^\vee$, as follows:
\beq
\label{eq:dotProductComplex}
x_1\tran x_2 = \frac{\tilde x_1^* \tilde x_2 + \tilde x_1 \tilde x_2^*}{2} 
\eeq
%

Moreover, we know that the action of a matrix $Z \in \aSOtwo$ can be written as 
the product of complex numbers, see~\eqref{eq:complexA}.

Combining~\eqref{eq:dotProductComplex} and~\eqref{eq:complexA} we get:
%
\bea
\label{eq:x1Ax2}
x_1\tran Z x_2 &\sim& \frac{\tilde x_1^* \; \tilde z \;  \tilde x_2  + \tilde x_1 \; \tilde z^* \; \tilde x_2^*}{2}
\eea
where $\tilde z = Z^\vee$.
Furthermore, when $Z$ is multiple of the identity matrix, it easy to see that $z = Z^\vee$ is actually a 
real number, and \eq~\eqref{eq:x1Ax2} becomes:
\bea
\label{eq:x1Ax1}
x_1\tran Z x_1 &\sim& 
\tilde x_1^* \; z \; \tilde x_1 
\eea

With the machinery introduced so far, we are ready to rewrite 
the cost $x\tran \Wanc x$ in complex form. 
Since $\Wanc$ is symmetric, the product becomes:
\bea
\label{eq:xWx}
x\tran \Wanc x = 
\sum_{i=1}^{2n-1}
\left[
x_i\tran [\Wanc]\ii x_i + 
\sum_{j=i+1}^{2n-1} 
2\; x_i\tran [\Wanc]\ij x_j 
\right]
\eea
Using the fact that $[\Wanc]\ii$ is a multiple of the identity matrix, $\Wcpx\ii \doteq [\Wanc]\ii^\vee \in \Real{}$, 
and using~\eqref{eq:x1Ax1} we conclude $x_i\tran [\Wanc]\ii x_i = \tilde x_i^* \Wcpx\ii \tilde x_i$. Moreover, defining 
$\Wcpx\ij \doteq [\Wanc]\ij^\vee$ (these will be complex numbers, in general), and using~\eqref{eq:x1Ax2}, 
eq.~\eqref{eq:xWx} becomes:
\bea
\label{eq:xWx-cpx}
x\tran \Wanc x &=&\sum_{i=1}^{2n-1}
\left[
\tilde x_i^* \Wcpx\ii \tilde x_i + 
\sum_{j=i+1}^{2n-1} 
(\tilde x_i^* \Wcpx\ij \tilde x_j + \tilde x_i \Wcpx\ij^* \tilde x_j^*)
\right] \nonumber \\ 
&=& 
\sum_{i=1}^{2n-1}
\left[
\tilde x_i^* \Wcpx\ii \tilde x_i + 
\sum_{j \neq i} 
\tilde x_i^* \Wcpx\ij \tilde x_j
\right] =
\tilde x^* \Wcpx \tilde x
\eea
where we completed the lower triangular part of $\Wcpx$ as $\Wcpx\ji = \Wcpx\ij^*$. 

\subsection{Proof of Proposition~\ref{prop:szepWtilde}: Zero Eigenvalues in $\Wcpx$}
\label{sec:proof:szepWtilde}

Let us denote with $N_0$ the number  of zero eigenvalues of the pose graph matrix $\Wcpx$.
$N_0$ can be written in terms of 
 the dimension of the matrix ($\Wcpx \in \Complex{(\dimWcpx)\times (\dimWcpx)}$)
 and the rank of the matrix:
 \beq
	N_0  = (\dimWcpx) - \mbox{rank}(\Wcpx)
 \eeq
 Now, recalling the
  factorization of $\Wcpx$ given in~\eqref{eq:WcpxFactorized}, 
   we note that:
\beq
\mbox{rank}(\Wcpx)  = \mbox{rank}
\left( 
 \left[\ba{cc} \Aanc & 
\tilde D \\  \zeros & 
\IncidenceUGGcpx \ea\right]
\right) = \mbox{rank}(\Aanc) + \mbox{rank}(\IncidenceUGGcpx)
\eeq
where the second relation follows from the upper triangular 
structure of the matrix. 
Now, we know from~\cite[Section 19.3]{Schrijver98book} 
that the anchored incidence matrix 
$\Aanc$, obtained by removing a row from the 
the incidence matrix of a connected graph, is full rank:
\beq
\mbox{rank}(\Aanc) = n-1
\eeq
Therefore:
\beq
\label{eq:pr1}
N_0  = n - \mbox{rank}(\IncidenceUGGcpx)
\eeq
Now, since we recognized that $\IncidenceUGGcpx$ is the complex incidence matrix
of a unit gain graph (Lemma~\ref{lem:unitGain}), we can use the result of Lemma 2.3 in~\cite{Reff:11}, which says that:
 \beq
 \label{eq:pr2}
 \mbox{rank}(\IncidenceUGGcpx) = n-b,
 \eeq
 where $b$ is the number of connected components in the  graph that are balanced. 
 Since we are working on a connected graph (Assumption~\ref{ass:connected}), $b$ can be either one 
(balanced graph or tree), or zero otherwise. Using~\eqref{eq:pr1} and~\eqref{eq:pr2}, we obtain $N_0 = b$, which implies
 that that $N_0 = 1$ for balanced graphs or trees, or $N_0 = 0$, otherwise.

\subsection{Proof of Proposition~\ref{prop:W_VS_Wtilde}: Spectrum of Complex and Real Pose Graph Matrices}
\label{sec:proof:W_VS_Wtilde}

Recall that any Hermitian matrix has real eigenvalues, and possibly complex eigenvectors.
Let $\mu \in\Real{}$ be an eigenvalue of $\Wcpx$, associated with an eigenvector $\tilde v\in\Complex{2n-1}$, i.e.,
\bea
\Wcpx \tilde v &=& \mu \tilde v \label{eq:eigc1}
\eea
From equation (\ref{eq:eigc1}) we have, for $i=1,\ldots,2n-1$,
\bea
\label{eq:sameEig-i}
\sum_{j=1}^{2n-1} \Wcpx\ij \tilde v_{j} = \mu \tilde v_i 
\Leftrightarrow \sum_{j=1}^{2n-1} [\Wanc]\ij v_j = \mu v_i
\eea
where $v_i$ is such that $v_i^\vee = \tilde v_i$. Since eq.~\eqref{eq:sameEig-i} holds for all $i=1,\ldots,2n-1$, 
it can be written in compact form as:
\beq
\Wanc v = \mu v
\eeq
 hence $v$ is an eigenvector of the real anchored pose graph matrix
$\Wanc$, associated with the eigenvalue $\mu$. This proves that any eigenvalue of $\Wcpx$ is 
also an eigenvalue of $\Wanc$.

To prove that the eigenvalue $\mu$ is actually repeated twice in $\Wanc$, 
consider now equation (\ref{eq:eigc1}) and multiply both members by the 
complex number $\e^{\j \frac{\pi}{2}}$:
\bea
\Wcpx \tilde v \e^{\j \frac{\pi}{2}} &=& \mu \tilde v \e^{\j \frac{\pi}{2}}
\eea
For $i=1,\ldots,2n-1$, we have:
\bea
\label{eq:sameEig2-i} 
\sum_{j=1}^{2n-1} \Wcpx\ij^* \tilde v_{j} \e^{\j \frac{\pi}{2}} = \mu \tilde v_i \e^{\j \frac{\pi}{2}}
\Leftrightarrow 
\sum_{j=1}^{2n-1} [\Wanc]\ij w_j = \mu w_i
\eea
where $w_i$ is such that $w_i^\vee = \tilde v_{j} \e^{\j \frac{\pi}{2}}$.
Since eq.~\eqref{eq:sameEig2-i} holds for all $i=1,\ldots,2n-1$, 
it can be written in compact form as:
\beq
\Wanc w = \mu w
\eeq
hence also $w$ is an eigenvector of
$\Wanc$ associated with the eigenvalue $\mu$. 

Now it only remains to demonstrate that $v$ and $w$ are linearly independent. 
One can readily check that, if $\tilde v_i$ is in the form $\tilde v_i = \eta_i \e^{\j \theta_i}$, then 
%
\beq
\label{eq:vi} 
v_i = 
\eta_i \vect{\cos(\theta_i) \\ \sin(\theta_i)}. 
\eeq
Moreover, observing that $\tilde v_{j} \e^{\j \frac{\pi}{2}} = \eta_i \e^{\j (\theta_i + \pi/2)}$, then
%
\beq
\label{eq:wi}
w_i = 
\eta_i \vect{\cos(\theta_i + \pi/2) \\ \sin(\theta_i +\pi/2)} =  
\eta_i \vect{-\sin(\theta_i) \\ \cos(\theta_i)}
\eeq
From~\eqref{eq:vi} and~\eqref{eq:wi} is it easy to see that 
$v\tran w = 0$, thus $v,w$ are orthogonal, hence independent.
To each eigenvalue $\mu$ of $\Wcpx$ there thus correspond an identical eigenvalue 
of $\Wanc$, of geometric multiplicity at least two. Since $\Wcpx$ has $2n-1$ eigenvalues
and $\Wanc$ has $2(2n-1)$ eigenvalues, we conclude that 
to each eigenvalue $\mu$ of $\Wcpx$ there  correspond exactly two
eigenvalues of $\Wanc$ in $\mu$. The previous proof also shows how the 
set of orthogonal eigenvectors of $\Wanc$ is related to the set of 
eigenvectors of $\Wcpx$.

\subsection{Proof of Theorem~\ref{thm:Everett}: Primal-dual Optimal Pairs}
\label{sec:thm:Everett}

We prove that, given  $\lam\in\Real{n}$, if an $\xcpx_\lam\in\calX(\lam)$ is primal feasible, 
then $\xcpx_\lam$ is primal optimal; moreover,
$\lam$ is dual optimal, and the duality gap is zero.

By weak duality we know that for any $\lam$:
\beq
\label{eq:ineq1}
\calL(x_\lam,\lam) \leq f^\star 
\eeq
However, if $x_\lam$ is primal feasible, 
by optimality of $f^\star$, it must also hold
\beq
\label{eq:ineq2}
f^\star \leq  f(x_\lam)
\eeq
Now we observe that for a feasible $x_\lam$, the terms in the 
Lagrangian associated to the constraints disappear and 
$\calL(x_\lam,\lam) = f(x_\lam)$. Using the latter equality 
and the inequalities~\eqref{eq:ineq1} and~\eqref{eq:ineq2} we get: 
\beq
\label{eq:ineq3}
f^\star \leq  f(x_\lam) = \calL(x_\lam,\lam) \leq f^\star 
\eeq
%
which implies $f(x_\lam)  = f^\star$, i.e., $x_\lam$ is primal optimal.

Further, we have that
\[
d^\star \geq \min_{x} \calL(x,\lam) = \calL(x_\lam,\lam) = f(x_\lam) = f^\star, 
\]
which, combined with weak duality ($d^\star \leq f^\star$), implies that $d^\star = f^\star$
and that $\lam$ attains the dual optimal value.

\subsection{Numerical Data For the Toy Examples in Section~\ref{sec:experiments}}
\label{sec:app:numericaldata}

Ground truth nodes poses, written as $x_i = [p_i\tran, \theta_i]$:

\beq
\begin{array}{ccccccc}
 x_1 &=& [ &   0.0000 &  -5.0000  &  0.2451 & ]\\
 x_2 &=& [ &  4.7553  & -1.5451 &  -0.4496 & ]\\
 x_3 &=& [ &  2.9389  &  4.0451 &   0.7361 & ]\\
  x_4 &=& [ &-2.9389  &  4.0451  &  0.3699 &]\\
 x_5 &=& [ & -4.7553  & -1.5451  & -1.7225 &]
 \end{array}
\eeq

\noindent
Relative measurements, for each edge $(i,j)$, written as $(i,j) : [\Delta\ij\tran, \theta\ij]$:
\beq
\begin{array}{ccccccc}
 (1,2) &:& [ &   4.6606 &   1.2177 &   2.8186 & ]\\
 (2,3) &:& [ & -4.4199  &  4.8043  &  0.1519  & ]\\
 (3,4) &:& [ & -4.1169  &  4.9322  &  0.5638  & ]\\
 (4,5) &:& [ & -3.6351  & -5.0908  & -0.5855&]\\
 (5,1) &:& [ & 3.4744   & 5.9425  &  2.5775 &]
 \end{array}
\eeq


\bibliography{refs,mybiblio} 

\end{document}